\documentclass[10pt,twocolumn,letterpaper]{article}
 \usepackage[pagenumbers]{cvpr} 
\usepackage{xspace}
\usepackage{subcaption}
\usepackage{graphicx}
\usepackage[autolanguage]{numprint}
\usepackage[utf8]{inputenc}
\usepackage{microtype}
\usepackage{soul}
\usepackage{float}
\usepackage{textcomp, gensymb}
\usepackage{amssymb}
\usepackage{booktabs}
\usepackage{array}
\usepackage{relsize}
\usepackage[export]{adjustbox}

\usepackage[dvipsnames]{xcolor}

\newcommand\blfootnote[1]{%
\begingroup 
\renewcommand\thefootnote{}\footnote{#1}%
\addtocounter{footnote}{-1}%
\endgroup 
}
\definecolor{cvprblue}{rgb}{0.21,0.49,0.74}
\usepackage[pagebackref,breaklinks,colorlinks,citecolor=cvprblue]{hyperref}

\def\name{GaussianEditor~}

\author{
\textbf{Yiwen Chen$^{\text{\textbf{\textcolor{red}{*}}}1,2}$} \quad 
\textbf{Zilong Chen$^{\text{\textbf{\textcolor{red}{*}}}3,5}$} \quad 
\textbf{Chi Zhang$^2$} \quad
\textbf{Feng Wang$^3$}  \quad
\textbf{Xiaofeng Yang$^2$} \\ \quad
\textbf{Yikai Wang$^3$} \quad
\textbf{Zhongang Cai$^4$}  \quad 
\textbf{Lei Yang$^4$} \quad 
\textbf{Huaping Liu$^3$} \quad 
\textbf{Guosheng Lin$^{\text{\textbf{\textcolor{red}{**}}}1,2}$}
\vspace{0.2cm} \\
$ ^1$S-Lab, Nanyang Technological University \\
$ ^2$School of Computer Science and Engineering, Nanyang Technological University \\
$ ^3$Department of Computer Science and Technology, Tsinghua University  \\
$ ^4$SenseTime Research \quad
$ ^5$ShengShu\\
{\fontsize{10}{10}\selectfont \url{https://buaacyw.github.io/gaussian-editor}}
}

\title{GaussianEditor: Swift and Controllable 3D Editing with Gaussian Splatting}

\begin{document}

\twocolumn[{
\maketitle
\begin{figure}[H]
    \hsize=\textwidth
    \centering
    \vspace{-10mm}
    \includegraphics[width=01.0\textwidth]{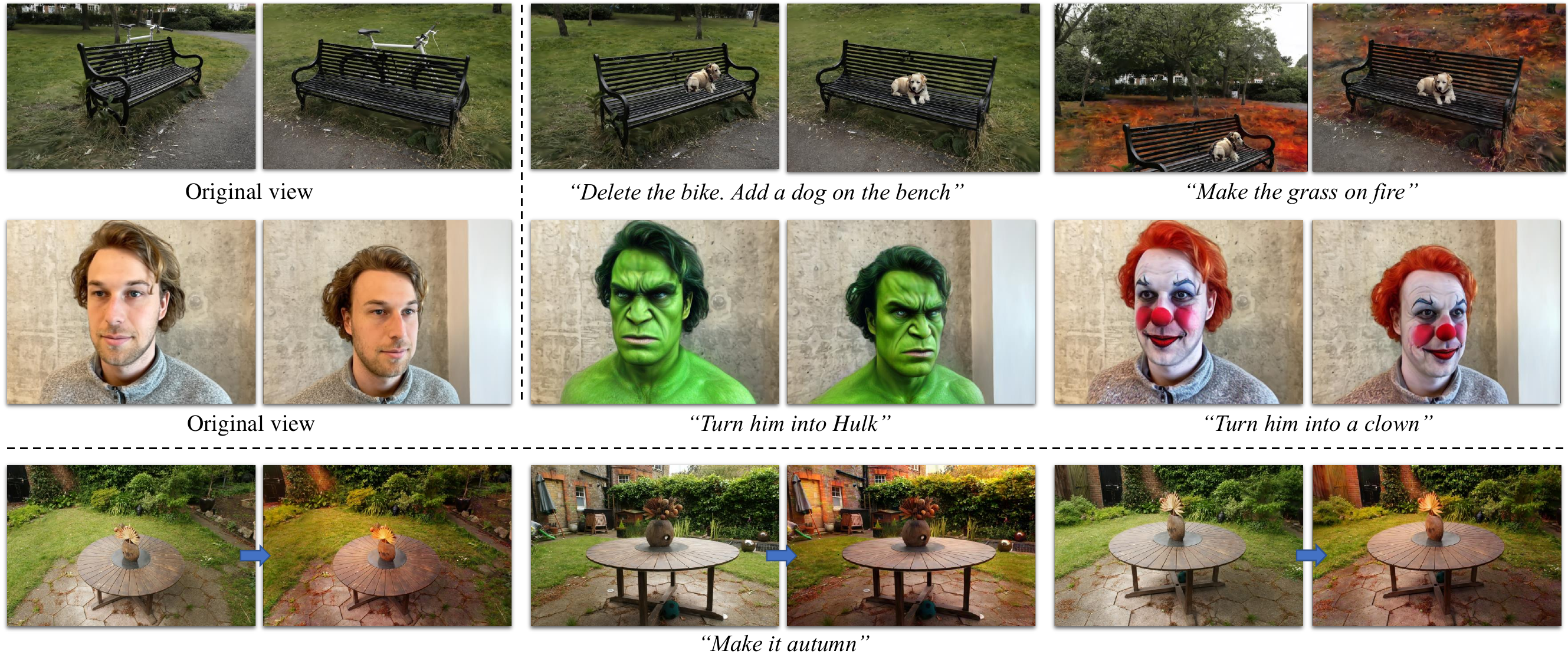}
    \begin{minipage}{1.0\textwidth}
        \vspace{2mm}
        \caption{\textbf{Results of GaussianEditor}. \name offers swift, controllable, and versatile 3D editing. A single editing session only takes 5-10 minutes. Please note our precise editing control, where only the desired parts are modified. Taking the \textit{``Make the grass on fire''} example from the first row of the figure, other objects in the scene such as the bench and tree remain unaffected. }
        \label{fig: teaser}
    \end{minipage}
\end{figure}
}]
\blfootnote{\textcolor{red}{**} Corresponding author.}
\blfootnote{\textcolor{red}{*} The first two authors contributed equally to this work.}

\begin{abstract}
3D editing plays a crucial role in many areas such as gaming and virtual reality. Traditional 3D editing methods, which rely on representations like meshes and point clouds, often fall short in realistically depicting complex scenes.
 On the other hand, methods based on implicit 3D representations, like Neural Radiance Field (NeRF), render complex scenes effectively but suffer from slow processing speeds and limited control over specific scene areas.
In response to these challenges, our paper presents GaussianEditor, an innovative and efficient 3D editing algorithm  based on Gaussian Splatting (GS), a novel 3D representation.
GaussianEditor enhances precision and control in editing through our proposed  Gaussian semantic tracing, which traces the editing target throughout the training process.
Additionally, we propose Hierarchical Gaussian splatting (HGS) to achieve stabilized and fine results under stochastic generative guidance from 2D diffusion models.
We also develop editing strategies for efficient object removal and integration, a challenging task for existing methods.
Our comprehensive experiments demonstrate GaussianEditor's superior control, efficacy, and rapid performance, marking a significant advancement in 3D editing.

\end{abstract}  
\section{Introduction}
\label{sec:intro}
In the evolving field of computer vision, the development of user-friendly 3D representations and editing algorithms is a key objective. Such technologies are vital in various applications, ranging from digital gaming to the growing MetaVerse. Traditional 3D representations like meshes and point clouds have been preferred due to their interactive editing capabilities. However, these methods face challenges in accurately rendering complex 3D scenes.

The recent rise of implicit 3D representations, exemplified by the Neural Radiance Field (NeRF)~\cite{nerf}, represents a paradigm shift in 3D scene rendering. NeRF's capacity for high-fidelity rendering, coupled with its implicit nature that offers significant expansibility, marks a substantial improvement over conventional approaches\cite{barron2021mip,zhang2020nerf++,park2021hypernerf}. This dual advantage has placed a significant focus on the NeRF framework in 3D editing~\cite{wang2022clip, wang2023nerf, haque2023instruct, zhuang2023dreameditor, park2023ed}, establishing it as a foundational approach for a considerable duration. 
However, NeRF's reliance on high-dimensional multilayer perception (MLP) networks for scene data encoding presents limitations. It restricts direct modification of specific scene parts and complicates tasks, like inpainting and scene composition. This complexity extends to the training and rendering processes, hindering practical applications.

In light of these challenges, our research is focused on developing an advanced 3D editing algorithm. This algorithm aims for flexible and rapid editing of 3D scenes, integrating both implicit editing, like text-based editing, and explicit control, such as bounding box usage for specific area modifications. 
To achieve these goals, we choose Gaussian Splatting (GS)~\cite{3dgs} for its real-time rendering and explicit point cloud-like representations.

However, editing Gaussian Splatting (GS)~\cite{3dgs} faces distinct challenges. A primary issue is the absence of efficient methods to accurately identify target Gaussians, which is crucial for precise controllable editing. Moreover, it has been observed in~\cite{chen2023text, tang2023dreamgaussian, yi2023gaussiandreamer} that optimizing Gaussian Splatting (GS) using highly random generative guidance like Score Distillation Sampling~\cite{dreamfusion} poses significant challenges. One possible explanation is that, unlike implicit representations buffered by neural networks, GS is directly affected by the randomness in loss. Such direct exposure results in unstable updates, as the properties of Gaussians are directly changed during training. Besides, each training step of GS may involve updates to a vast number of Gaussian points. This process occurs without the moderating influence of neural network-style buffering mechanisms. As a result, the excessive fluidity of the 3D GS scene hinders its ability to converge to finely detailed results like implicit representations when trained with generative guidance.

To counter these issues, in this work, we propose \name, a novel, swift, and highly controllable 3D editing algorithm for Gaussian Splatting. \name can fulfill various high-quality editing needs within minutes. A key feature of our method is the introduction of Gaussian semantic tracing, which enables precise control over Gaussian Splatting (GS). Gaussian semantic tracing consistently identifies the Gaussians requiring editing at every moment during training. This contrasts with traditional 3D editing methods that often depend on static 2D or 3D masks. Such masks become less effective as the geometries and appearances of 3D models evolve during training. Gaussian semantic tracing is achieved by unprojecting 2D segmentation masks into 3D Gaussians and assigning each Gaussian a semantic tag. As the Gaussians evolve during training, these semantic tags enable the tracking of the specific Gaussians targeted for editing. Our Gaussian tracing algorithm ensures that only the targeted areas are modified, enabling precise and controllable editing. 

Additionally, to tackle the significant challenge of Gaussian Splatting (GS) struggling to fit fine results under highly random generative guidance, we propose a novel GS representation: hierarchical Gaussian splatting (HGS). In HGS, Gaussians are organized into generations based on their sequence in multiple densification processes during training. Gaussians formed in earlier densification stages are deemed older generations and are subject to stricter constraints, aimed at preserving their original state and thus reducing their mobility. Conversely, those formed in later stages are considered younger generations and are subjected to fewer or no constraints, allowing for more adaptability. HGS's design effectively moderates the fluidity of GS by imposing restrictions on older generations while preserving the flexibility of newer generations. This approach enables continuous optimization towards better outcomes, thereby simulating the buffering function achieved in implicit representations through neural networks. Our experiments also demonstrate that HGS is more adept at adapting to highly random generative guidance.

Finally, we have specifically designed a 3D inpainting algorithm for Gaussian Splatting (GS). As demonstrated in Fig.~\ref{fig: teaser}, we have successfully removed specific objects from scenes and seamlessly integrated new objects into designated areas. For object removal, we developed a specialized local repair algorithm that efficiently eliminates artifacts at the intersection of the object and the scene. For adding objects, we first request users to provide a prompt and a 2D inpainting mask for a particular view of the GS. Subsequently, we employ a 2D inpainting method to generate a single-view image of the object to be added. This image is then transformed into a coarse 3D mesh using image-to-3D conversion techniques. The 3D mesh is subsequently converted into the HGS representation and refined. Finally, this refined representation is concatenated into the original GS. The entire inpainting process described above is completed within 5 minutes.

\name offers swift, controllable, and versatile 3D editing. A single editing session typically only takes 5-10 minutes, significantly faster than previous editing processes. Our contributions can be summarized in four aspects:
\begin{enumerate}
    \item We have introduced Gaussian semantic tracing, enabling more detailed and effective editing control.
    \item We propose Hierarchical Gaussian Splatting (HGS), a novel GS representation capable of converging more stably to refined results under highly random generative guidance.
    \item We have specifically designed a 3D inpainting algorithm for Gaussian Splatting, which allows swift removal and addition objects.
    \item Extensive experiments demonstrate that our method surpasses previous 3D editing methods in terms of effectiveness, speed, and controllability.
\end{enumerate}

\section{Related Works}
\label{sec:related}

\subsection{3D Representations}
Various 3D representations have been proposed to address diverse 3D tasks. The groundbreaking work, Neural Radiance Fields (NeRF)~\cite{nerf}, employs volumetric rendering and has gained popularity for enabling 3D optimization with only 2D supervision. However, optimizing NeRF can be time-consuming, despite its wide usage in 3D reconstruction~\cite{li2023neuralangelo,chen2022mobilenerf, barron2022mipnerf360, hedman2021snerg} and generation~\cite{poole2022dreamfusion,lin2023magic3d} tasks.

While efforts have been made to accelerate NeRF training~\cite{mueller2022instant,yu_and_fridovichkeil2021plenoxels}, these approaches primarily focus on the reconstruction setting, leaving the generation setting less optimized. The common technique of spatial pruning does not effectively speed up the generation setting.

Recently, 3D Gaussian splatting~\cite{3dgs} has emerged as an alternative 3D representation to NeRF, showcasing impressive quality and speed in 3D and 4D reconstruction tasks~\cite{3dgs, luiten2023dynamic, yang2023deformable3dgs, wu20234d, yang2023real}. It has also attracted considerable research interest in the field of generation~\cite{chen2023text, tang2023dreamgaussian, yi2023gaussiandreamer}. Its efficient differentiable rendering implementation and model design facilitate fast training without the need for spatial pruning.

In this work, we pioneer the adaptation of 3D Gaussian splatting to 3D editing tasks, aiming to achieve swift and controllable 3D editing, harnessing the advantages of this representation for the first time in this context.

\subsection{3D Editing}
Editing neural fields is inherently challenging due to the intricate interplay between their shape and appearance. 
EditNeRF~\cite{liu2021editing} stands as a pioneering work in this domain, as they edit both the shape and color of neural fields by conditioning them on latent codes. 
Additionally, some works~\cite{wang2022clip, wang2023nerf, gao2023textdeformer, bao2023sine} leverage CLIP models to facilitate editing through the use of text prompts or reference images. 

Another line of research focuses on predefined template models or skeletons to support actions like re-posing or re-rendering within specific categories~\cite{peng2021neural, noguchi2021neural}. 
Geometry-based methods~\cite{yuan2022nerf, yang2022neumesh, xu2022deforming, li2022climatenerf} translate neural fields into meshes and synchronize mesh deformation with implicit fields. 
Additionally, 3D editing techniques involve combining 2D image manipulation, such as inpainting, with neural fields training~\cite{liu2022nerf, kobayashi2022decomposing}. 

Concurrent works~\cite{park2023ed,zhuang2023dreameditor} leverage static 2D and 3D masks to constrain the edit area of NeRF. 
However, these approaches have their limitations because the training of 3D models is a dynamic process, and static masks cannot effectively constrain it. 
In contrast, our research employs Gaussian semantic tracing to track the target Gaussian throughout the entire training process.

\section{Preliminary}
\label{pre:pre}

\subsection{3D Gaussian Splatting}
\label{pre: gs}
GS (Gaussian Splatting)~\cite{3dgs} represents an explicit 3D scene using point clouds, where Gaussians are employed to depict the scene's structure. In this representation, every Gaussian is defined by a center point, denoted as $x$, and a covariance matrix $\Sigma$. The center point $x$ is commonly known as the Gaussian's mean value:

\begin{equation}
\label{formula:gaussian's formula}
    G(x)=e^{-\frac{1}{2}x^T\Sigma^{-1}x}.
\end{equation}
The covariance matrix $\Sigma$ can be decomposed into a rotation matrix $\mathbf{R}$ and a scaling matrix $\mathbf{S}$ for differentiable optimization:
\begin{equation}
\label{formula:covariance decomposition}
    \Sigma = \mathbf{R}\mathbf{S}\mathbf{S}^T\mathbf{R}^T,
\end{equation}
the calculation of gradient flow is detailed in~\cite{3dgs}.

For rendering new viewpoints, the method of splatting, as described in~\cite{yifan2019differentiablesplatting}, is utilized for positioning the Gaussians on the camera planes. This technique, originally presented in~\cite{zwicker2001surfacesplatting}, involves a viewing transformation denoted by $W$ and the Jacobian $J$ of the affine approximation of the projective transformation. Using these, the covariance matrix $\Sigma^{\prime}$ in camera coordinates is determined as follows:

\begin{equation}
    \Sigma^{\prime} = JW\Sigma W^TJ^T.
\end{equation}
To summarize, each Gaussian point in the model is characterized by a set of attributes: its position, denoted as $x \in \mathbb{R}^3$, its color represented by spherical harmonics coefficients $c \in \mathbb{R}^k$ (where $k$ indicates the degrees of freedom), its opacity $\alpha \in \mathbb{R}$, a rotation quaternion $q \in \mathbb{R}^4$, and a scaling factor $s \in \mathbb{R}^3$.
\begin{figure}[h]
    \centering
    \vspace{-3mm}
    \begin{minipage}{0.235\textwidth}
        \centering
        \begin{subfigure}{\linewidth}
            \centering
            \includegraphics[width=1.0\linewidth]{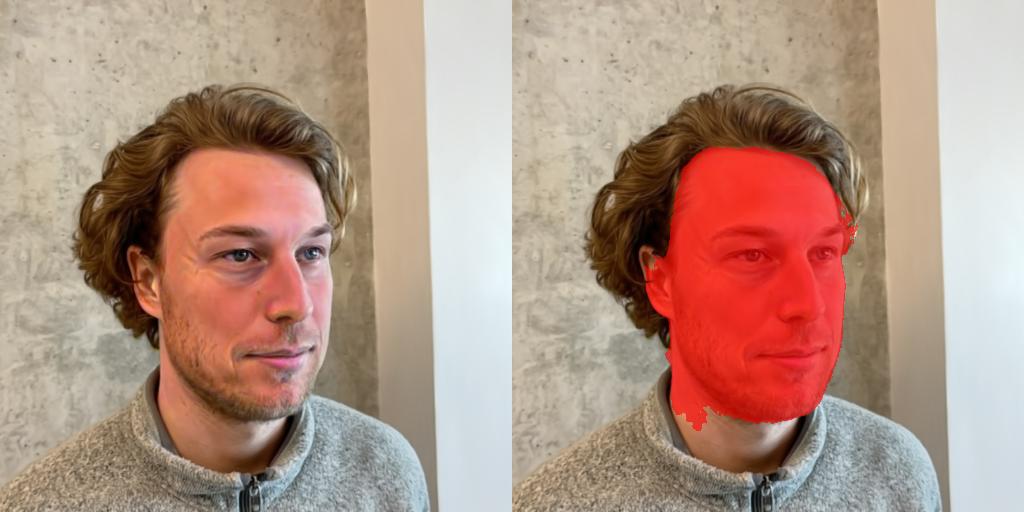}
            \caption*{Before Editing}
        \end{subfigure}
    \end{minipage}
    \hfill
    \begin{minipage}{0.235\textwidth}
        \centering
        \begin{subfigure}{\linewidth}
            \centering
            \includegraphics[width=1.0\linewidth]{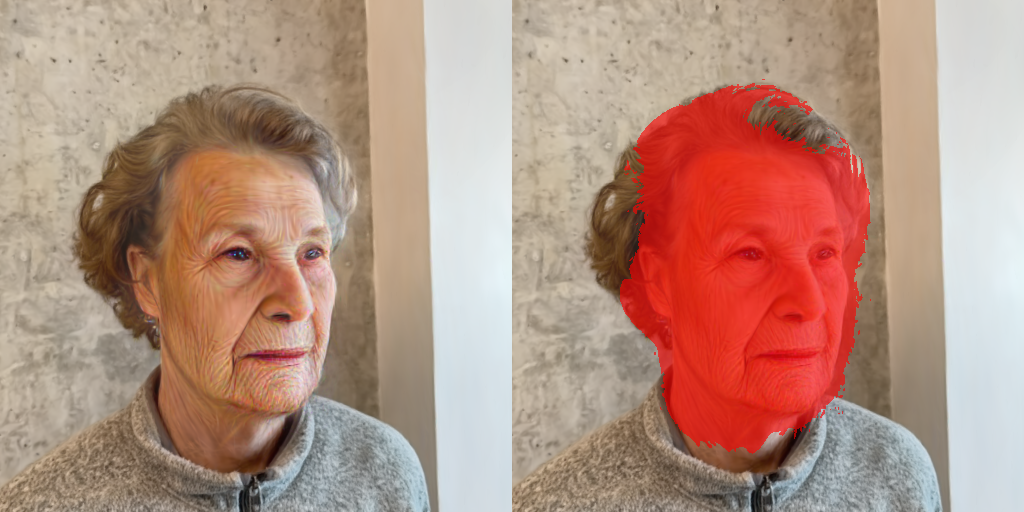}
            \caption*{During Editing}
        \end{subfigure}
    \end{minipage}
    \vspace{-2mm}
    \caption{\textbf{Illustration of Gaussian semantic tracing. Prompt: Turn him into an old lady.} The red mask in the images represents the projection of the Gaussians that will be updated and densified. The dynamic change of the masked area during the training process, as driven by the updating of Gaussians, ensures consistent effectiveness throughout the training duration. Despite starting with potentially inaccurate segmentation masks due to 2D segmentation errors, Gaussian semantic tracing still guarantees high-quality editing results.}
    \label{fig: mask}
    \vspace{-3mm}
\end{figure}
Particularly, for every pixel, the color and opacity of all Gaussians are calculated based on the Gaussian's representation as described in Eq.~\ref{formula:gaussian's formula}. The blending process of $N$ ordered points overlapping a pixel follows a specific formula:
\begin{equation}
\label{formula: splatting&volume rendering}
    C = \sum_{i\in N}c_i \alpha_i \prod_{j=1}^{i-1} (1-\alpha_j).
\end{equation}

where $c_i$ and $\alpha_i$ signify the color and density of a given point respectively. These values are determined by a Gaussian with a covariance matrix $\Sigma$, which is then scaled by optimizable per-point opacity and spherical harmonics (SH) color coefficients.

\subsection{Diffusion-based Editing Guidance}
\label{pre: Editing Guidance}

Recent advancements have seen numerous works elevating 2D diffusion processes to 3D, applying these processes extensively in the realm of 3D editing. Broadly, these works can be categorized into two types. The first type~\cite{dreamfusion, park2023ed,zhuang2023dreameditor,sella2023vox,mikaeili2023sked,cheng2023progressive3d}, exemplified by Dreamfusion's~\cite{dreamfusion} introduction of SDS loss, involves feeding the noised rendering of the current 3D model, along with other conditions, into a 2D diffusion model~\cite{rombach2022high}. The scores generated by the diffusion model then guide the direction of model updates. The second type~\cite{haque2023instruct,shao2023control4d,raj2023dreambooth3d,chen2023it3d} focuses on conducting 2D editing based on given prompts for the multiview rendering of a 3D model. This approach creates a multi-view 2D image dataset, which is then utilized as a training target to provide guidance for the 3D model.

Our work centers on leveraging the exemplary properties of Gaussian Splatting’s explicit representation to enhance 3D editing. Consequently, we do not design specific editing guidance mechanisms but instead directly employ the guidance methods mentioned above. Both types of guidance can be applied in our method. For simplicity, we denote the guidance universally as $D$. Given the parameters of a 3D model, ${\Theta}$, along with the rendered camera pose $p$ and prompt $e$, the editing loss from the 2D diffusion prior can be formulated as follows:
\begin{equation}
 \mathcal{L}_\text{Edit} =D({\Theta}; p, e)
\end{equation}

\section{Method}
\label{sec:method}

We define the task of 3D editing on Gaussian Splatting (GS) as follows: Given a prompt $y$ and a 3D scene represented by 3D Gaussians, denoted by ${\Theta}$, where each ${\Theta}_i = \{x_i, s_i, q_i, \alpha_i, c_i\}$ represents the parameters of the $i$-th Gaussian as detailed in Sec.~\ref{pre: gs}, the objective is to achieve an edited 3D Gaussians, referred to as ${\Theta}_y$, that aligns with or adheres to the specifications of the prompt $y$.

We then introduce our novel framework for performing editing tasks on GS. We first introduce Gaussian semantic tracing in Sec.~\ref{sec:gs seg}, along with a new representation method known as Hierarchical Gaussian Splatting (HGS) in Sec.~\ref{sec: Hierarchical gs}. The GS semantic tracing enables precise segmentation and tracing within GS, facilitating controllable editing operations. Compared to the standard GS, the HGS representation demonstrates greater robustness against randomness in generative guidance and is more adept at accommodating a diverse range of editing scenarios. Additionally, we have specifically designed 3D inpainting for GS, which encompasses object removal and addition (Sec.~\ref{sec: 3D Inpainting}). 

\subsection{Gaussian Semantic Tracing}
\label{sec:gs seg}

Previous works~\cite{park2023ed, zhuang2023dreameditor} in 3D editing usually utilize static 2D or 3D masks to apply loss only within the masked pixels, thus constraining the editing process to only edit the desired area. However, this method has limitations. As 3D representations dynamically change during training, static segmentation masks would become inaccurate or even ineffective. Furthermore, the use of static masks to control gradients in NeRF editing poses a significant limitation, as it confines the editing strictly within the masked area. This restriction prevents the edited content from naturally extending beyond the mask, thus 'locking' the content within a specified spatial boundary. 

Even with the implementation of semantic NeRF ~\cite{zhi2021place}, the gradient control is still only effective at the very beginning of the training since the ongoing updates to NeRF lead to a loss of accuracy in the semantic field.

\begin{figure}[t]
    \centering
    \begin{minipage}{0.23\textwidth}
        \centering
        \begin{subfigure}{\linewidth}
            \centering
            \includegraphics[width=1.0\linewidth]{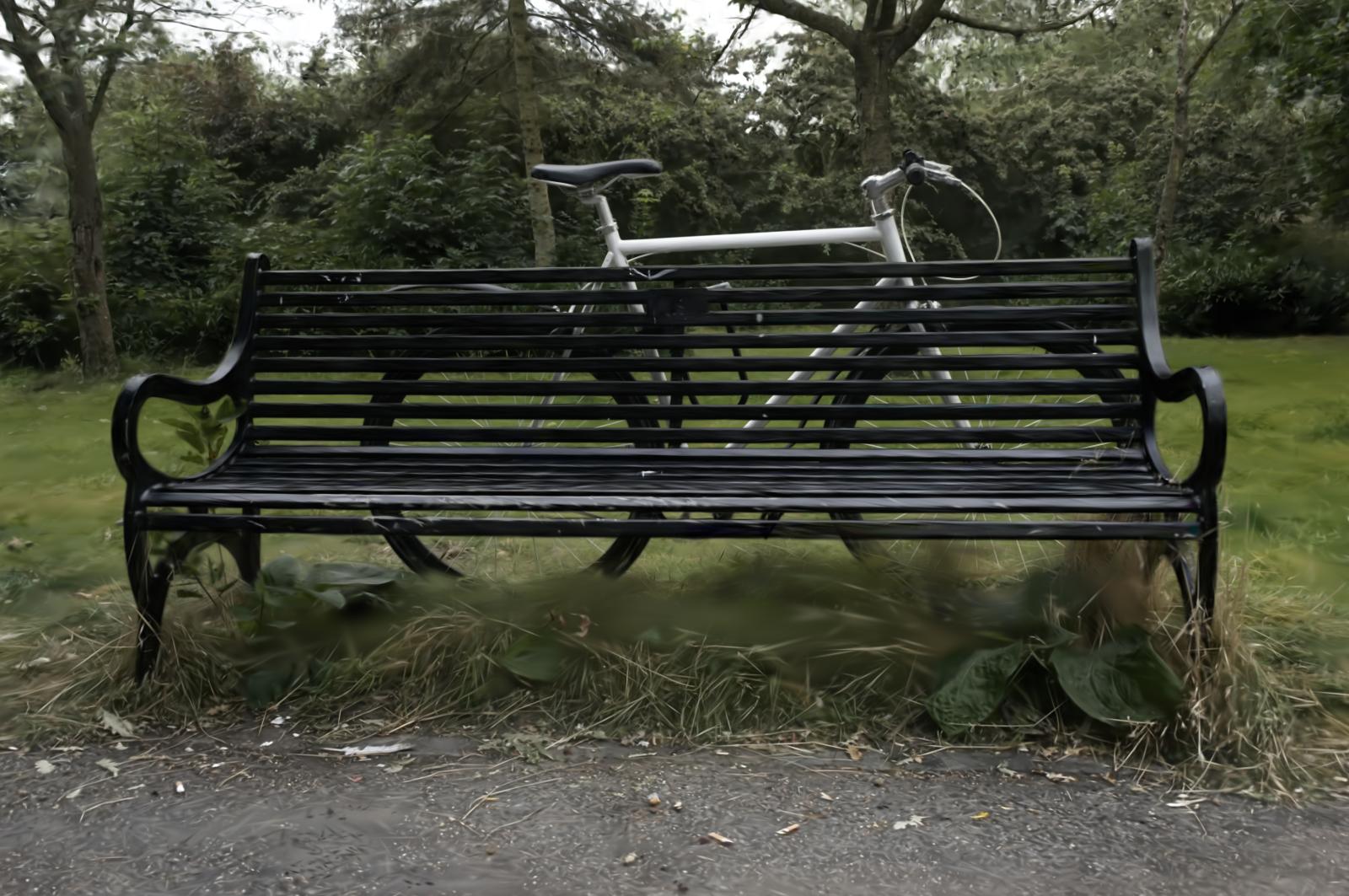}
            \caption*{Original GS}
        \end{subfigure}
    \end{minipage}
    \hfill
    \begin{minipage}{0.23\textwidth}
        \centering
        \begin{subfigure}{\linewidth}
            \centering
            \includegraphics[width=1.0\linewidth]{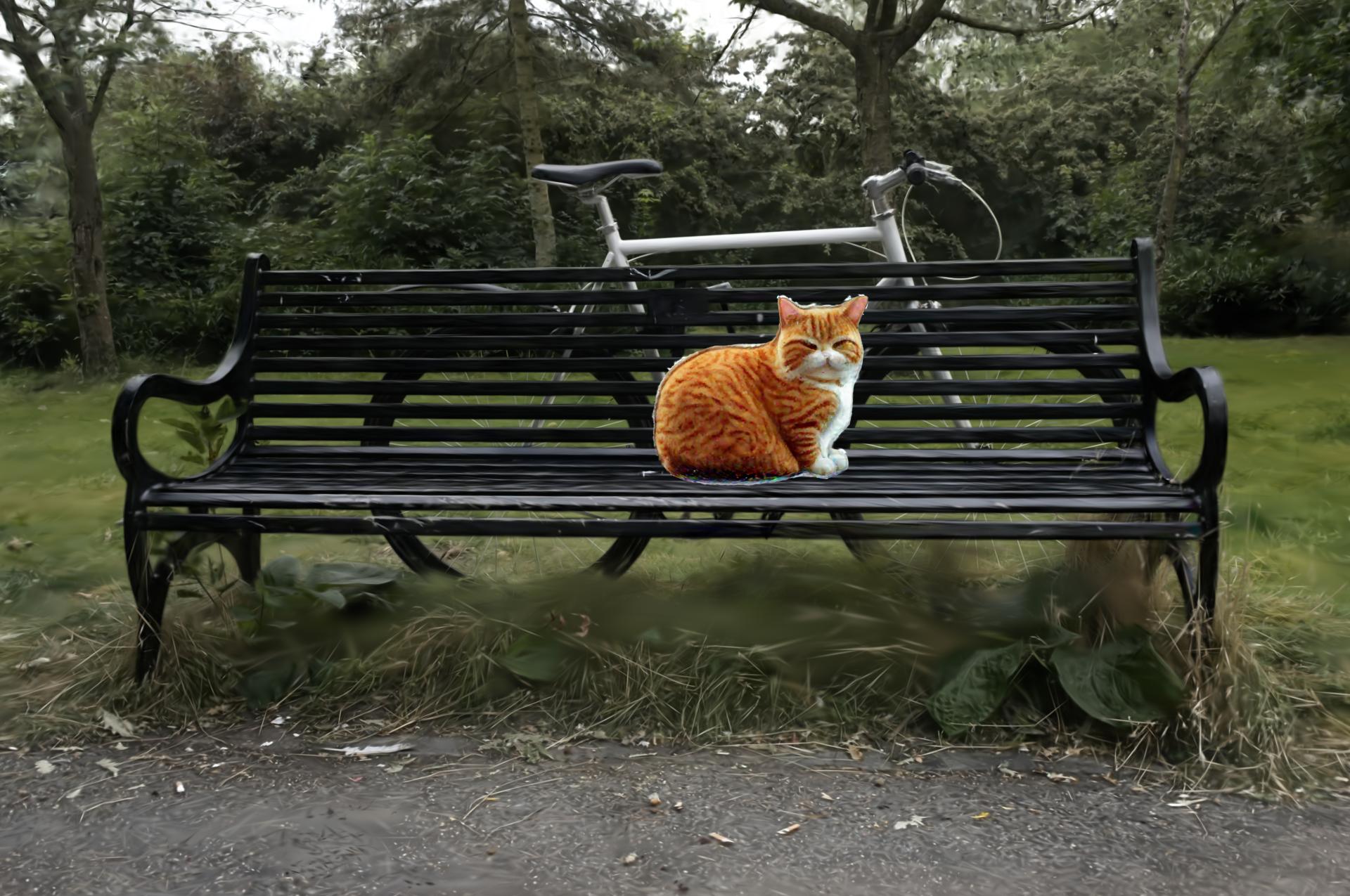}
            \caption*{A Garfield cat on the bench}
        \end{subfigure}
    \end{minipage}

    \begin{minipage}{0.23\textwidth}
        \centering
        \begin{subfigure}{\linewidth}
            \centering
            \includegraphics[width=1.0\linewidth]{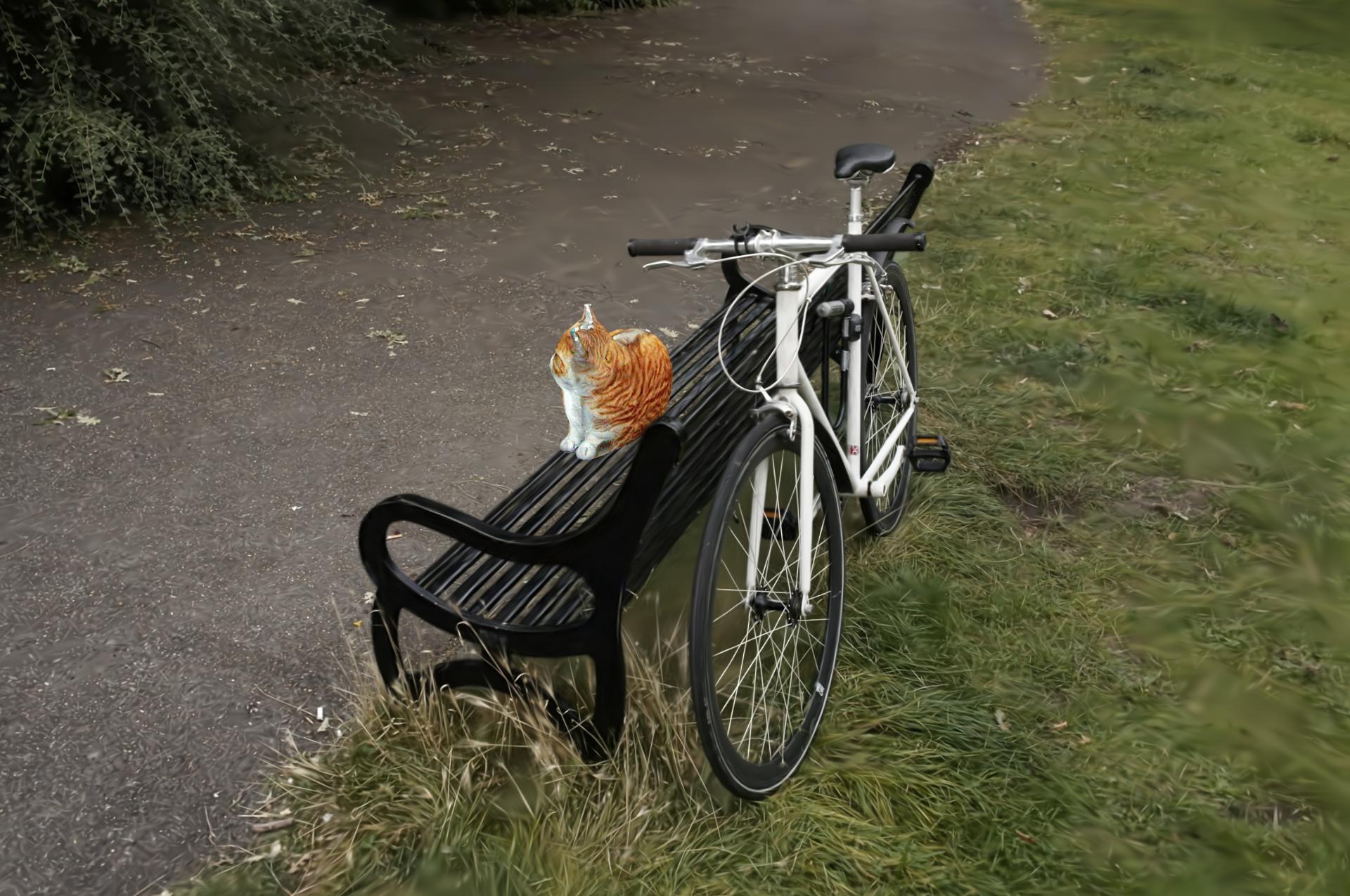}
            \caption*{Novel view 1}
        \end{subfigure}
    \end{minipage}
    \hfill
    \begin{minipage}{0.23\textwidth}
        \centering
        \begin{subfigure}{\linewidth}
            \centering
            \includegraphics[width=1.0\linewidth]{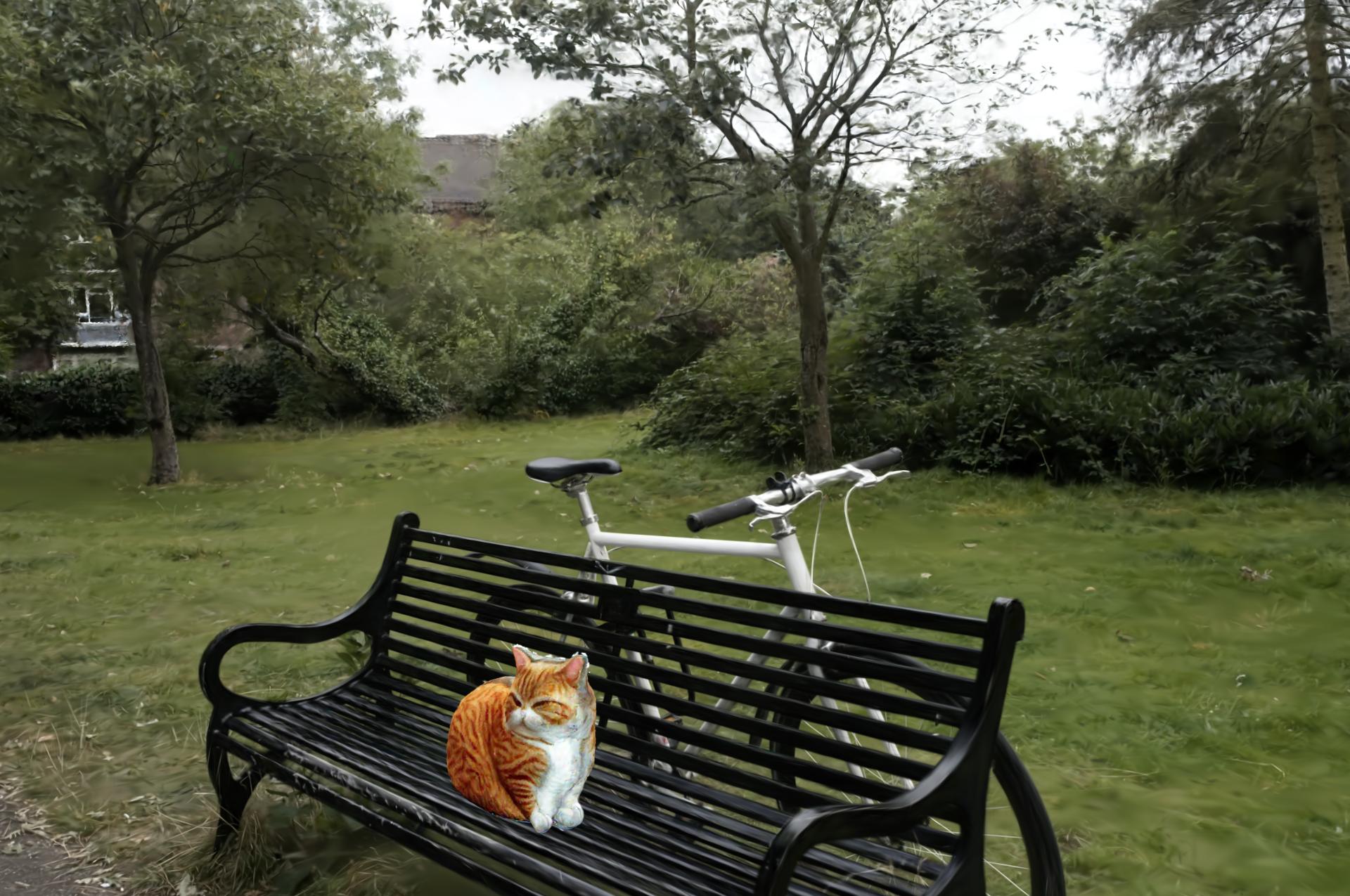}
            \caption*{Novel view 2}
        \end{subfigure}
    \end{minipage}
    \vspace{-3mm}
    \caption{\textbf{3D inpainting for object incorporation.} \name is capable of adding objects at specified locations in a scene, given a 2D inpainting mask and a text prompt from a single view. The whole process takes merely five minutes.}
    \label{3_inpaint_cat}
    \vspace{-5mm}
\end{figure}

To address the aforementioned issue, we have chosen Gaussian Splatting (GS) as our 3D representation due to its explicit nature. This allows us to directly assign semantic labels to each Gaussian point, thereby facilitating semantic tracing in 3D scenes.

Specifically, we enhance the 3D Gaussians ${\Theta}$ by adding a new attribute $m$, where $m_{ij}$ represents the semantic Gaussian mask for the $i$-th Gaussian point and the $j$-th semantic label. With this attribute, we can precisely control the editing process by selectively updating only the target 3D Gaussians. During the densification process, newly densified points inherit the semantic label of their parent point. This ensures that we have an accurate 3D semantic mask at every moment throughout the training process.

As illustrated in Fig.~\ref{fig: mask}, Gaussian semantic tracing enables continuous tracking of each Gaussian's categories during training, adjusting to their evolving properties and numbers. This feature is vital, as it permits selective application of gradients, densification and pruning of Gaussians linked to the specified category. Additionally, it facilitates training solely by rendering the target object, significantly speeding up the process in complex scenes. The semantic Gaussian mask $m$ functions as a dynamic 3D segmentation mask, evolving with the training, allowing content to expand freely in space. This contrasts with NeRF, where content is restricted to a fixed spatial area.

Next, we discuss Gaussian Splatting unprojection, the method we propose to obtain semantic Gaussian mask $m$. For a set of 3D Gaussians ${\Theta}$, we render them from multiple viewpoints to generate a series of renderings $\mathcal{I}$. These renderings are then processed using 2D segmentation techniques~\cite{kirillov2023segment} to obtain 2D segmentation masks $\mathcal{M}$, with each $\mathcal{M}_j$, representing the $j$-th semantic labels.

\begin{figure}[t]
    \centering
    \begin{minipage}{0.155\textwidth}
        \centering
        \begin{subfigure}{\linewidth}
            \centering
            \includegraphics[width=1.0\linewidth]{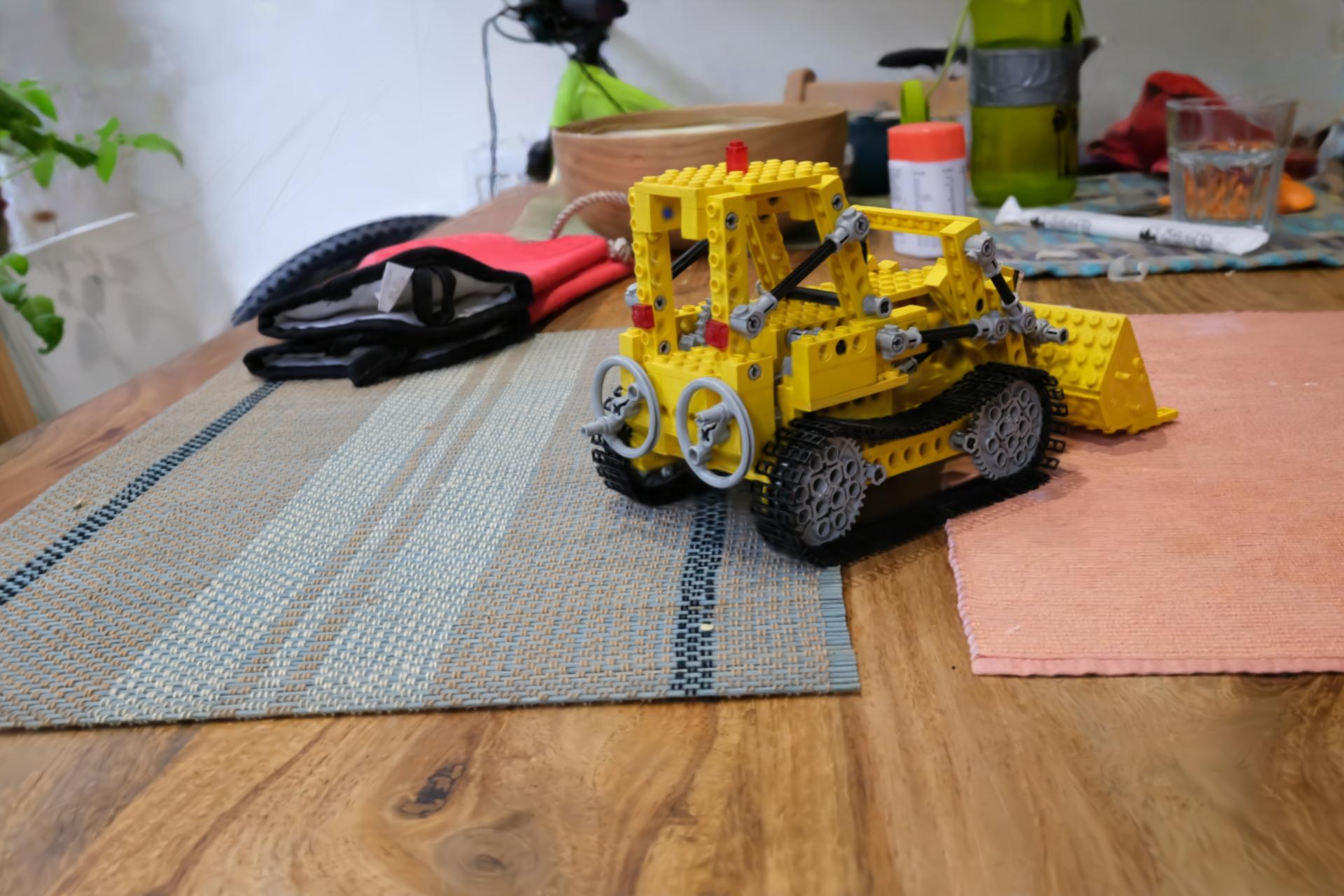}
        \end{subfigure}
    \end{minipage}
    \hfill
    \begin{minipage}{0.155\textwidth}
        \centering
        \begin{subfigure}{\linewidth}
            \centering
            \includegraphics[width=1.0\linewidth]{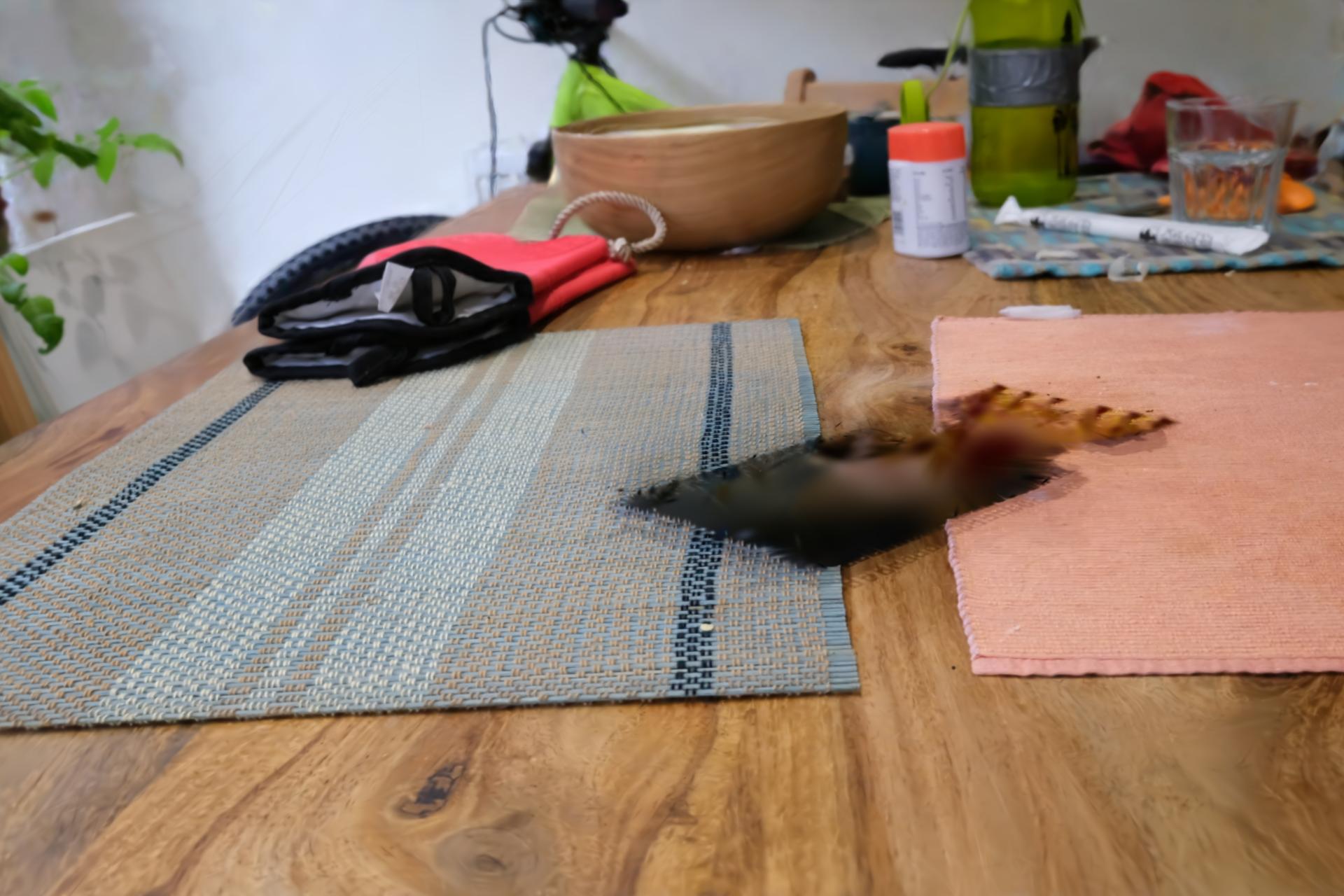}
        \end{subfigure}
    \end{minipage}
    \hfill
    \begin{minipage}{0.155\textwidth}
        \centering
        \begin{subfigure}{\linewidth}
            \centering
            \includegraphics[width=1.0\linewidth]{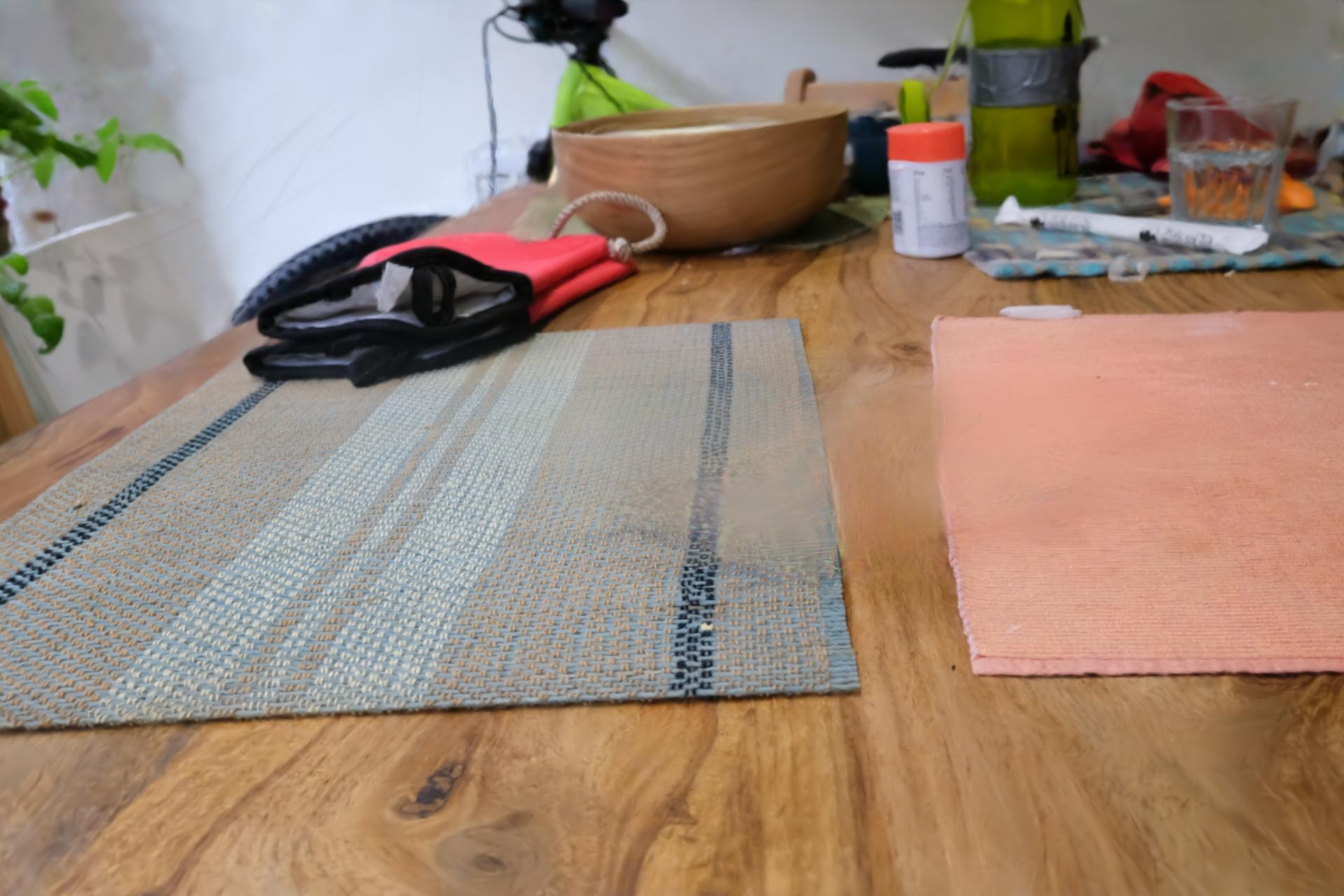}
        \end{subfigure}
    \end{minipage}

    \begin{minipage}{0.155\textwidth}
        \centering
        \begin{subfigure}{\linewidth}
            \centering
            \includegraphics[width=1.0\linewidth]{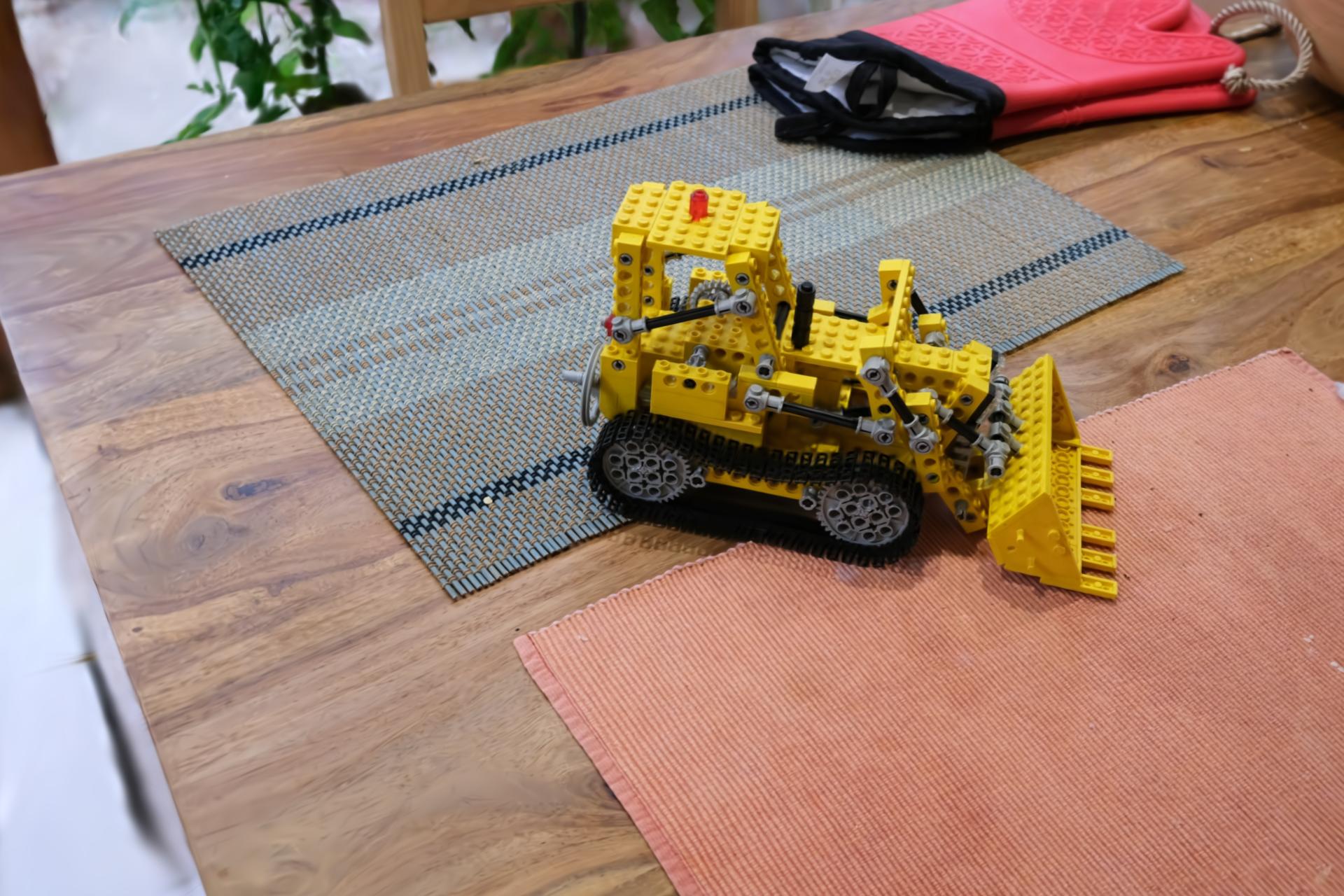}
            \caption*{Original View}
        \end{subfigure}
    \end{minipage}
    \hfill
    \begin{minipage}{0.155\textwidth}
        \centering
        \begin{subfigure}{\linewidth}
            \centering
            \includegraphics[width=1.0\linewidth]{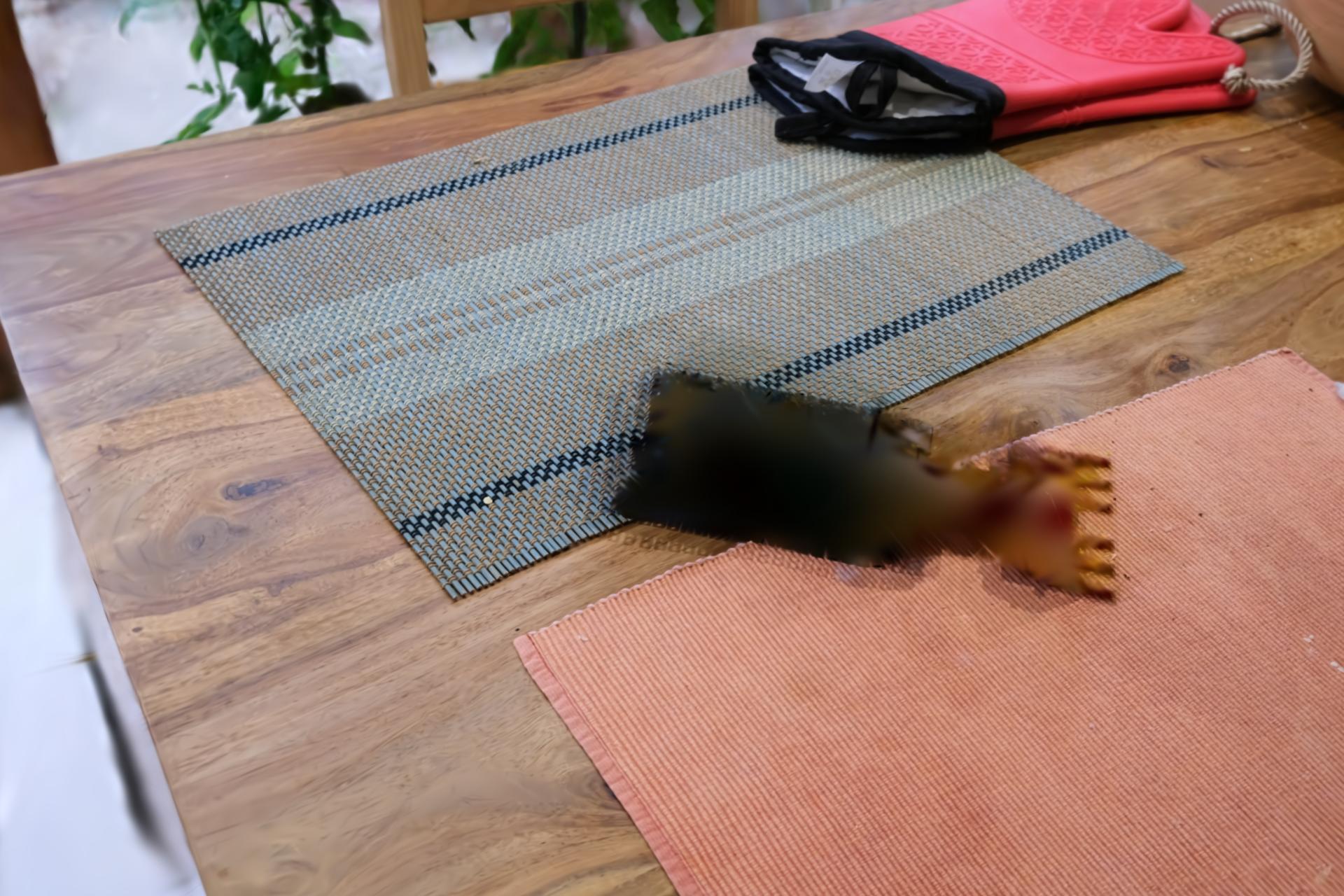}
            \caption*{Removed}
        \end{subfigure}
    \end{minipage}
    \hfill
    \begin{minipage}{0.155\textwidth}
        \centering
        \begin{subfigure}{\linewidth}
            \centering
            \includegraphics[width=1.0\linewidth]{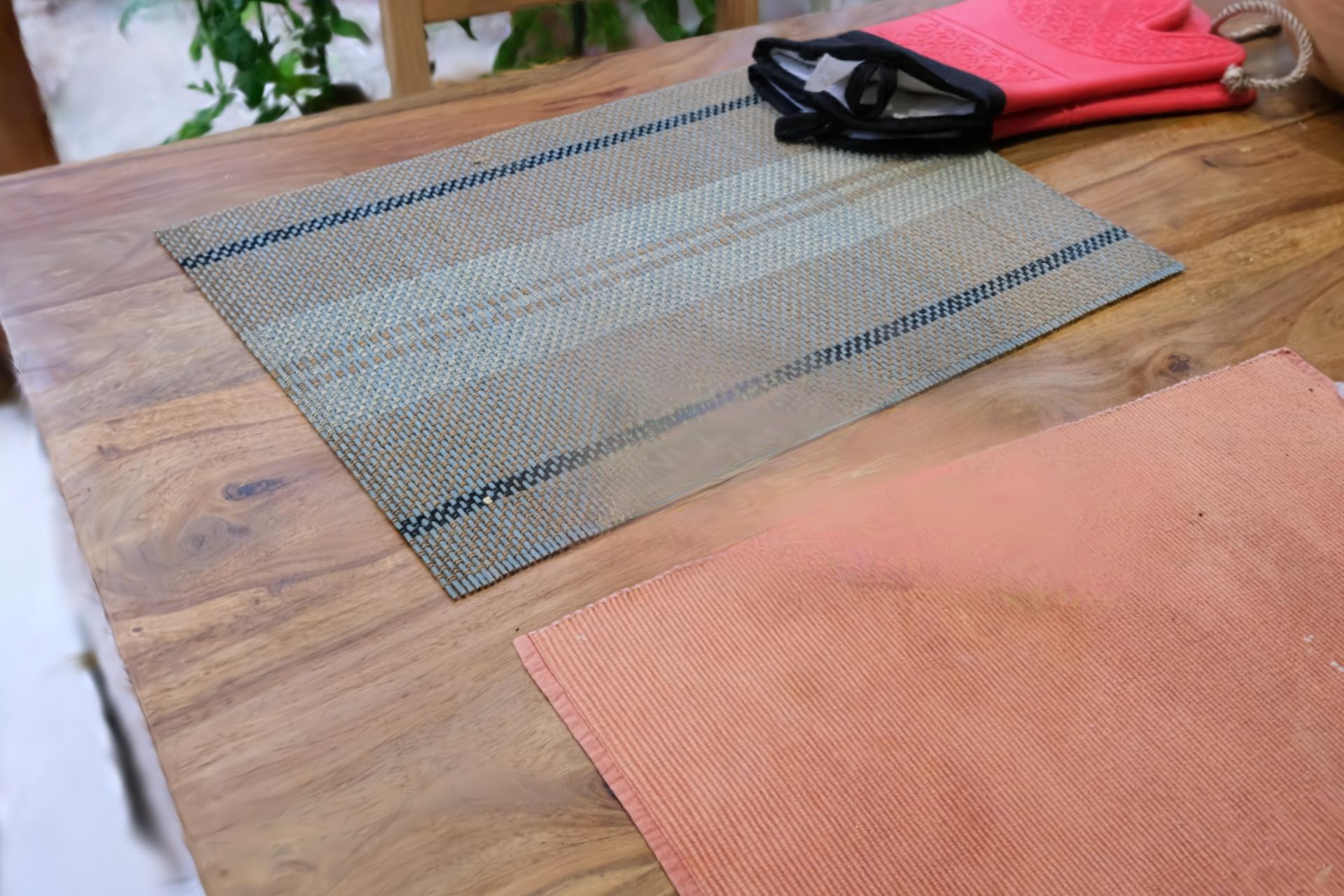}
            \caption*{Inpainted}
        \end{subfigure}
    \end{minipage}
    \vspace{-2mm}
    \caption{\textbf{3D inpainting for object removal.} Typically, removing the target object based on a Gaussian semantic mask generates artifacts at the interface between the target object and the scene. To address this, we generate a repaired image using a 2D inpainting method and employ Mean Squared Error (MSE) loss for supervision. The whole process takes merely two minutes.}
    \vspace{-5mm}
\end{figure}

To obtain the semantic label for each Gaussian, we unproject the posed 2D semantic label back to the Gaussians with inverse rendering. Concretely, we maintain a weight and a counter for each Gaussian. For pixel $\boldsymbol{p}$ on the semantic maps, we unproject the semantic label back to the Gaussians that affects it by

\begin{equation}
    w_{i}^j = \sum o_i(\boldsymbol{p}) * T_{i}^j(\boldsymbol{p}) * \mathcal{M}^j(\boldsymbol{p}),
\end{equation}

where $w_{i}^j$ represents the weight of the $i$-th Gaussian for the $j$-th semantic label, while $o_i(\boldsymbol{p})$, $T_{i}^j(\boldsymbol{p})$, and $\mathcal{M}^j(\boldsymbol{p})$ denote the opacity, transmittance from pixel $\boldsymbol{p}$, and semantic mask of pixel $\boldsymbol{p}$ for the $i$-th Gaussian, respectively.
After updating all the Gaussian weights and counters, we determine whether a Gaussian belongs to the $j$-th semantic class based on whether its average weight exceeds a manually set threshold.

The entire labeling process is remarkably fast, typically taking less than a second. Once this semantic label assignment is completed, the entire Gaussian scene becomes parsed by us, making a variety of operations possible. These include manually changing colors, moving properties of a specific category, and deleting certain categories. Notably, 2D diffusion guidance often struggles to effectively edit small objects in complex scenes. Thanks to Gaussian semantic tracing, we can now render these small objects independently and input them into the 2D diffusion model, thereby achieving more precise supervision.

\subsection{Hierarchical Gaussian Splatting}
\label{sec: Hierarchical gs}

The effectiveness of vanilla GS~\cite{kerbl3Dgaussians} in reconstruction tasks lies in the high-quality initialization provided by point clouds derived from SFM~\cite{schonberger2016structure}, coupled with stable supervision from ground truth datasets.

However, the scenario changes in the field of generation. In previous work involving GS in text-to-3D and image-to-3D~\cite{tang2023dreamgaussian,chen2023text,yi2023gaussiandreamer}, GS has shown limitations when facing the randomness of generative guidance due to its nature as a point cloud-like representation. This instability in GS is mainly due to their direct exposure to the randomness of loss functions,  unlike neural network-based implicit representations. GS models, which update a large number of Gaussian points each training step, lack the memorization and moderating ability of neural networks. This leads to erratic updates and prevents GS from achieving the detailed results seen in neural network-based implicit representations, as GS's excessive fluidity hampers its convergence in generative training.

\begin{figure*}[ht]
    \centering
    \vspace{-8mm}
    \includegraphics[width=1.0\textwidth]{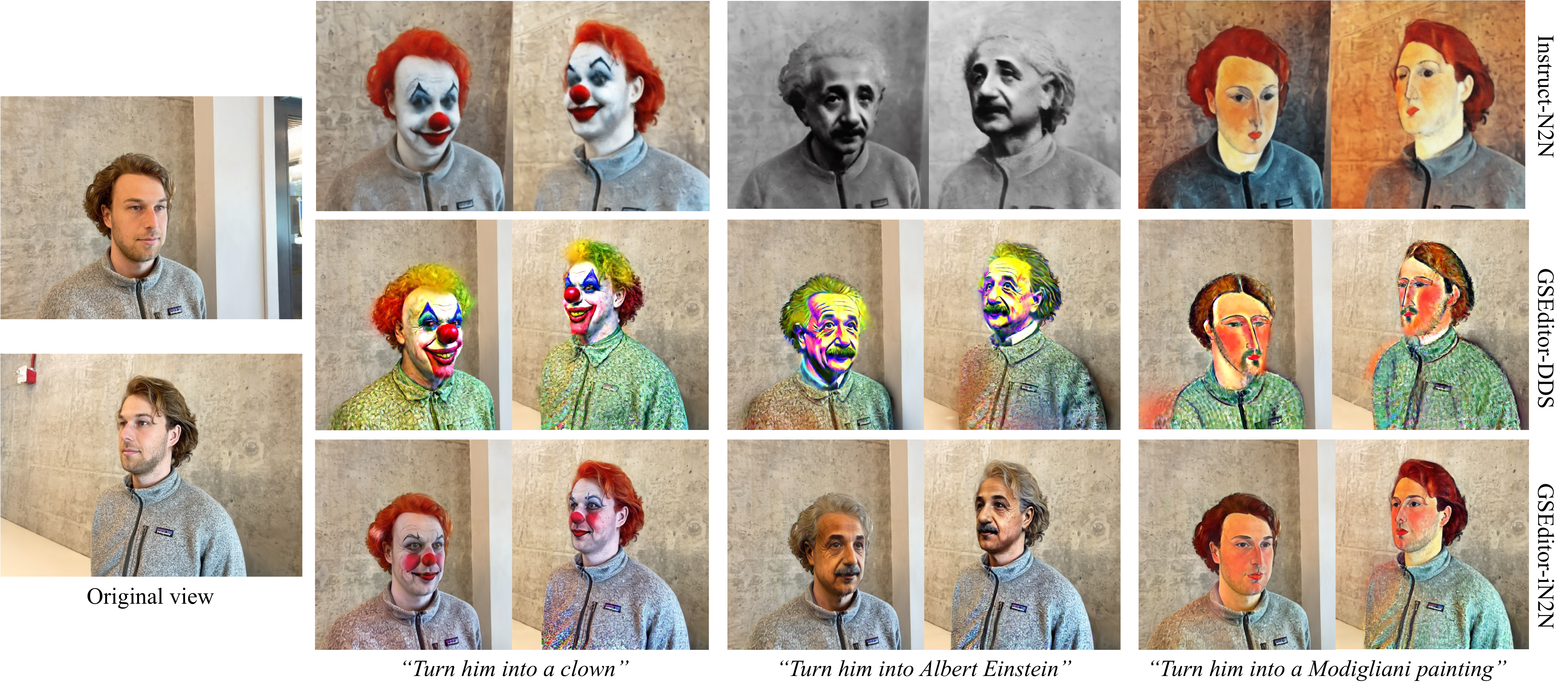}
    \vspace{-7mm}
    \caption{\textbf{Qualitative comparison.} It's important to note the level of control we maintain over the editing area (the whole body of the man). Background and other non-target regions are essentially unaffected, in contrast to Instruct-Nerf2Nerf~\cite{haque2023instruct} where the entire scene undergoes changes. 
GaussianEditor-DDS and GaussianEditor-iN2N indicate that we utilize delta denoising score~\cite{hertz2023delta} and Instruct-Nerf2Nerf~\cite{haque2023instruct} respectively, as guidance for editing.}
    \label{fig:comparison}
    \vspace{-2mm}
\end{figure*}
To address these challenges, we introduce Hierarchical Gaussian Splatting (HGS), a structured representation of GS that is more suitable for generative and editing scenarios. HGS categorizes GS into different generations based on the densification round in which a particular Gaussian point is produced. The initial Gaussians ${\Theta}$, are all assigned a generation of $0$. During the training process for editing, points generated in the $k$-th densification round are marked as generation $k$.

Subsequently, we impose varying constraints on Gaussians from different generations to control their degree of flexibility. The older the generation, the stronger the constraints applied. Anchor loss is utilized to enforce these constraints. At the beginning of training, HGS records the attributes of all Gaussians as anchors. These anchors are then updated to reflect the current state of the Gaussians at each densification process. During training, MSE loss between the anchor state and the current state is employed to ensure that the Gaussians do not deviate too far from their respective anchors:
\begin{equation}
	\mathcal{L}^{P}_{anchor} =   \sum_{i=0}^{n}\lambda_i(P_i - \hat{P}_i)^2
\end{equation}
where $n$ represents the total number of Gaussians and $P$ denotes a certain property of the current Gaussian, including elements from the set ${x, s, q, \alpha, c}$. Here, $\hat{P}$ refers to the same property recorded in the anchor state. The term $\lambda_i$ indicates the strength of the anchor loss applied to the $i$-th Gaussian, which varies based on its generation.
The overall training loss is defined as:
\begin{equation}
	\mathcal{L} =  \mathcal{L}_{Edit} + \sum_{P \in \{x, s, q, \alpha, c\}} \lambda_P \mathcal{L}^{P}_{anchor}
\end{equation}
In this equation, $\lambda_P$ signifies the strength of the anchor loss applied to property $P$, and $\mathcal{L}_{Edit}$ is the edit loss defined in Sec.~\ref{pre: Editing Guidance}.

This generational design in HGS prevents the issue of excessive flexibility in GS when faced with stochastic loss. With each densification, the anchor loss weight $\lambda_i$ for all previous generations of Gaussians is increased. As a result, the fluidity of the existing generations gradually decreases until it nearly solidifies. This approach ensures stable geometry formation under stochastic losses, relying on the almost unconstrained Gaussians from new densifications to carve out details. Furthermore, this method of applying anchor loss can effectively meet various editing needs. For instance, to limit changes in the original GS, one can increase the anchor loss weight for generation $0$. Similarly, if there is no desire to alter color or geometry during editing, a stronger anchor loss can be applied to these specific properties.
\begin{figure*}
    \centering
    \vspace{-8mm}
    \begin{minipage}{0.243\textwidth}
        \centering
        \begin{subfigure}{\linewidth}
            \centering
            \includegraphics[width=1.0\linewidth]{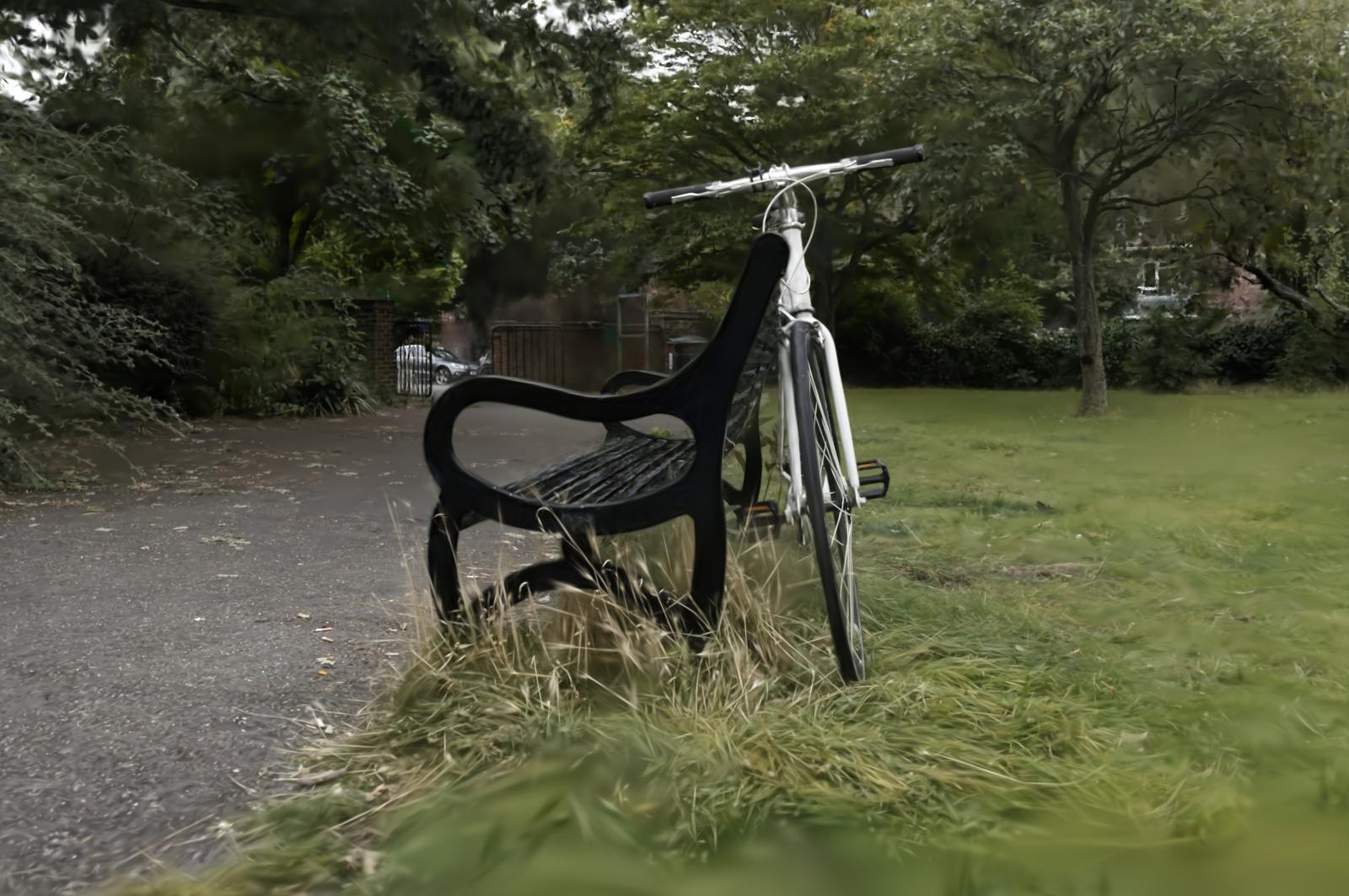}
        \end{subfigure}
    \end{minipage}
    \hfill
    \begin{minipage}{0.243\textwidth}
        \centering
        \begin{subfigure}{\linewidth}
            \centering
            \includegraphics[width=1.0\linewidth]{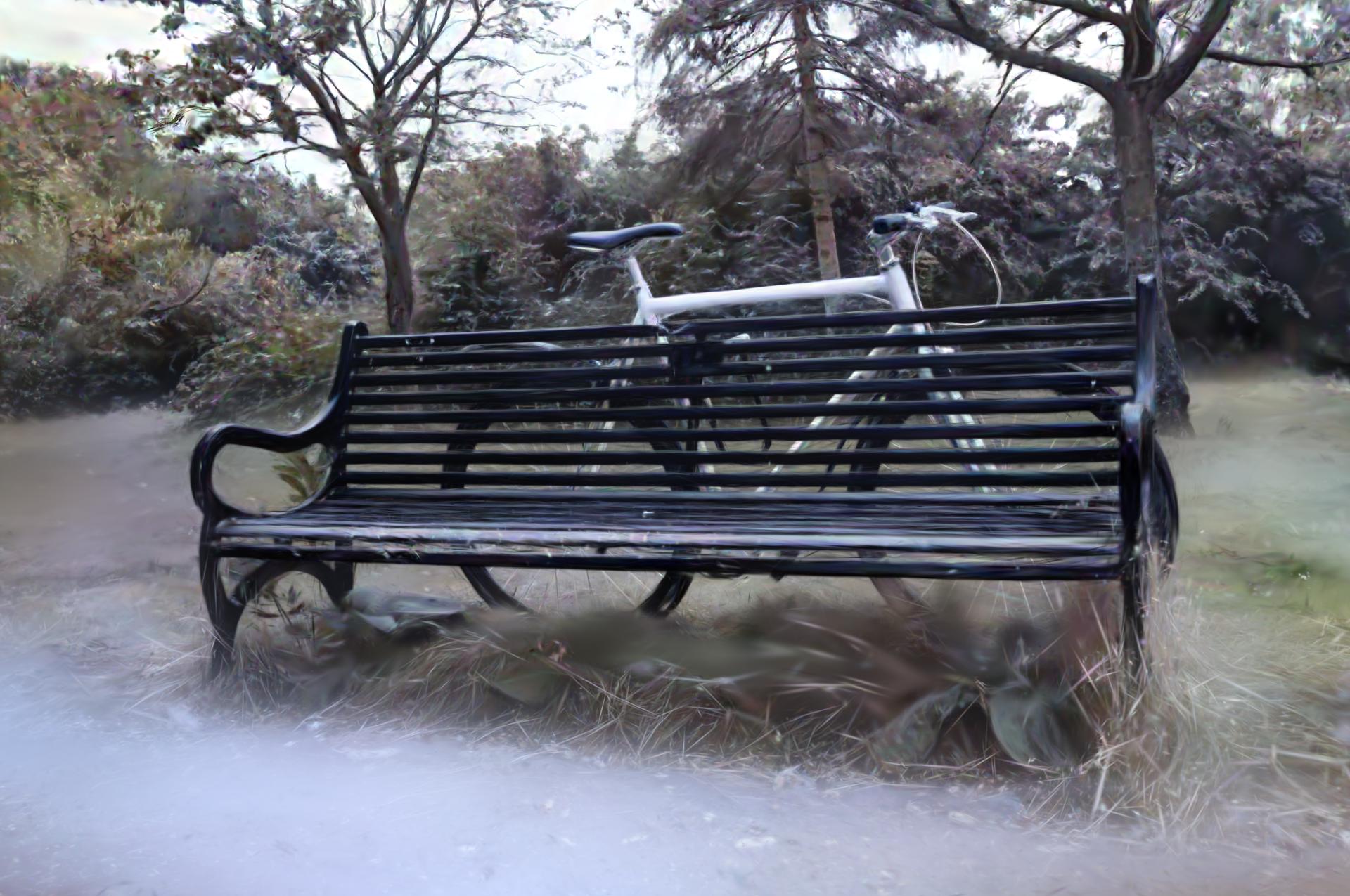}
        \end{subfigure}
    \end{minipage}
    \hfill
    \begin{minipage}{0.243\textwidth}
        \centering
        \begin{subfigure}{\linewidth}
            \centering
            \includegraphics[width=1.0\linewidth]{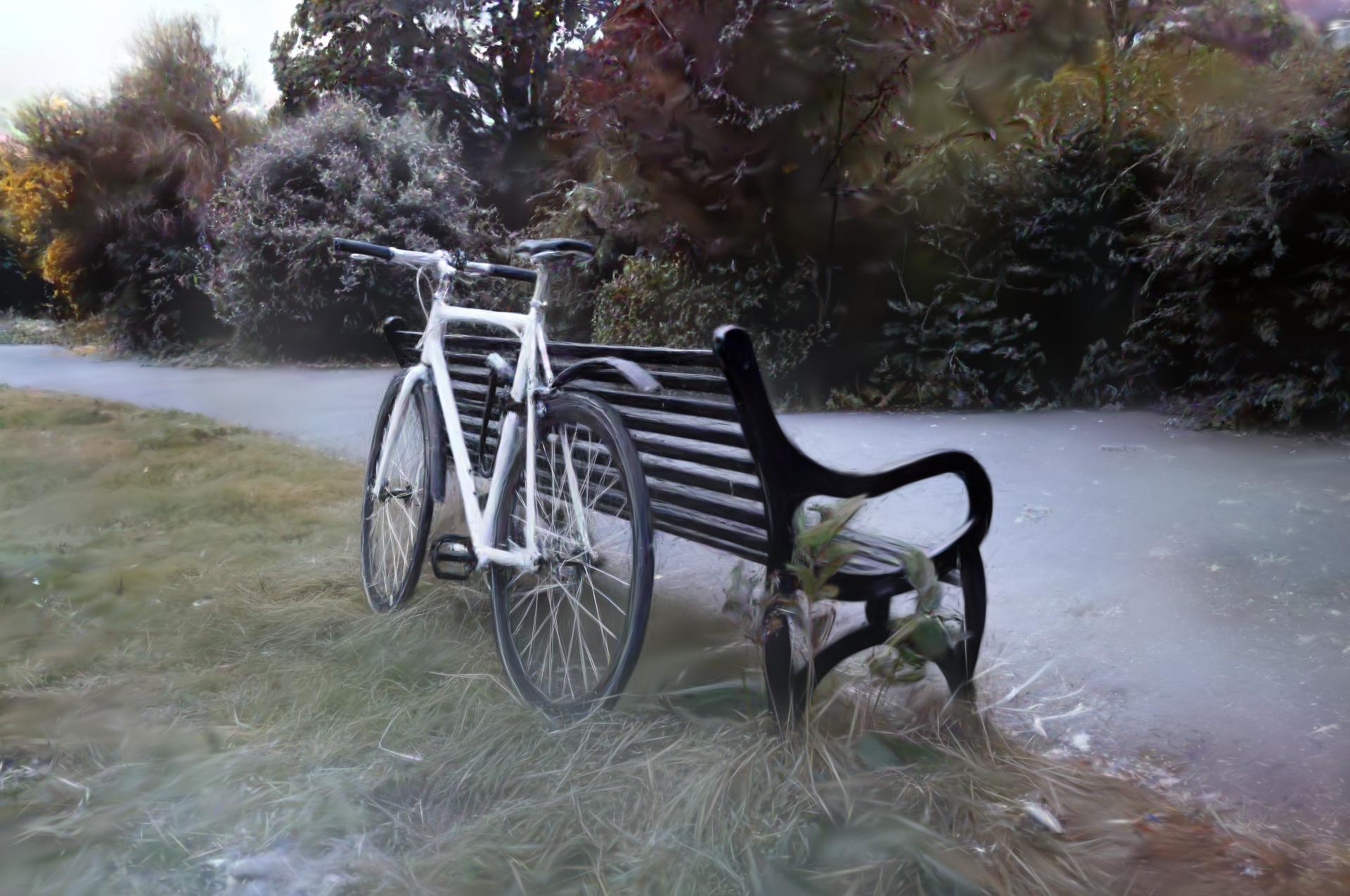}
        \end{subfigure}
    \end{minipage}
    \hfill
    \begin{minipage}{0.243\textwidth}
        \centering
        \begin{subfigure}{\linewidth}
            \centering
            \includegraphics[width=1.0\linewidth]{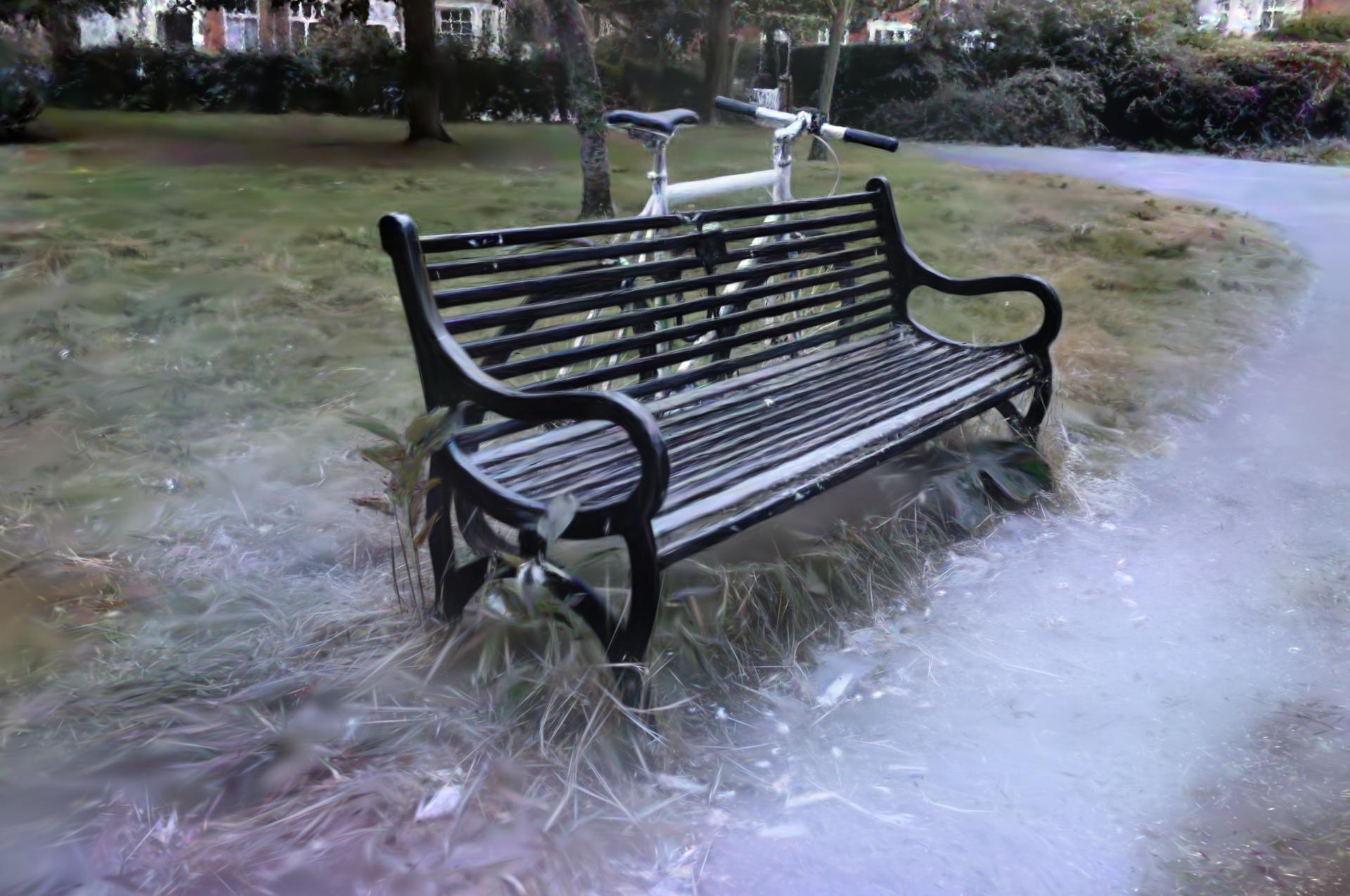}
        \end{subfigure}
    \end{minipage}
    \\ 
    \vspace{1mm}
    \textit{``Make it snowy"}
    \vspace{1mm}
    
    \begin{minipage}{0.243\textwidth}
        \centering
        \begin{subfigure}{\linewidth}
            \centering
            \includegraphics[width=1.0\linewidth]{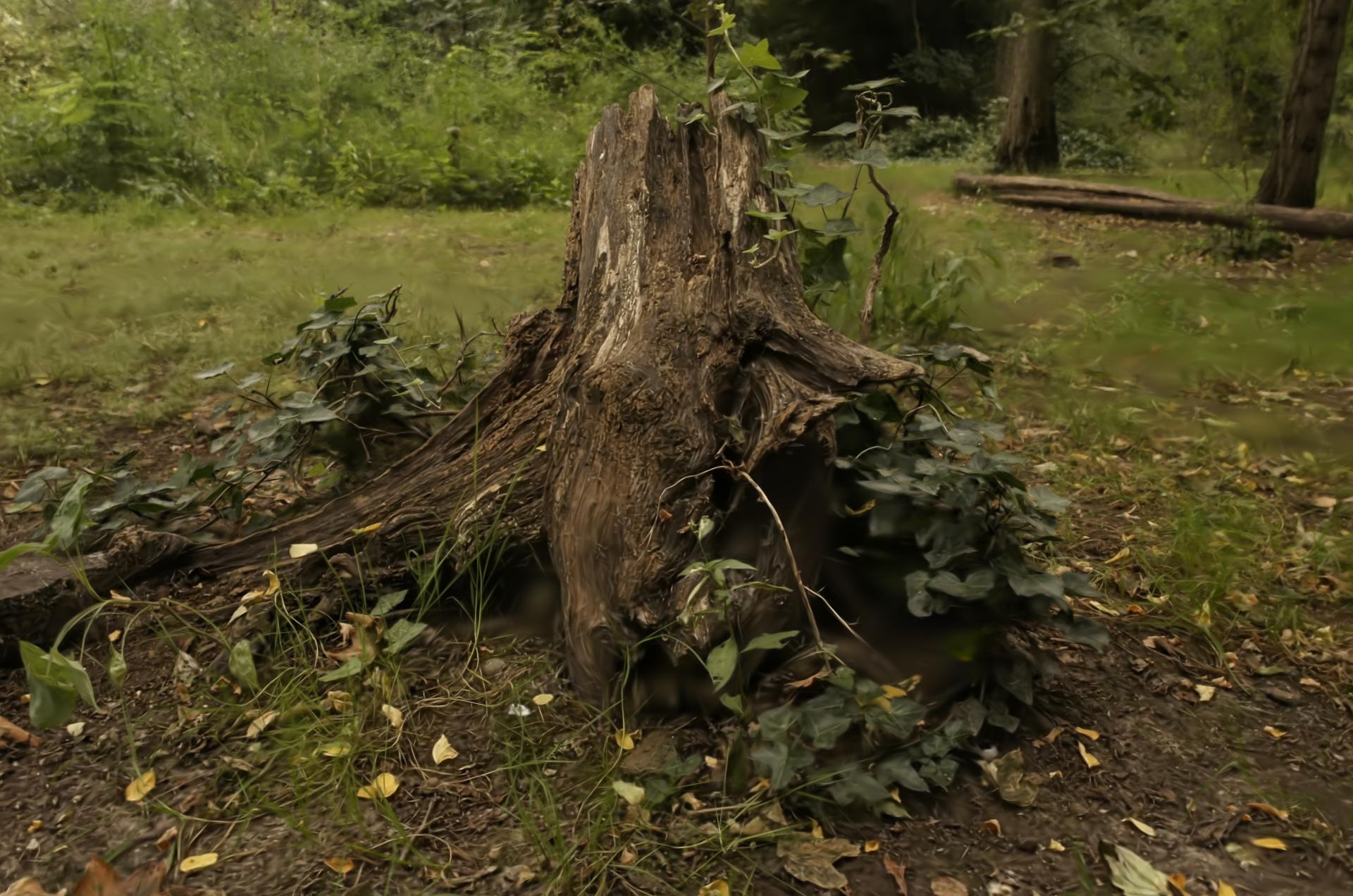}
        \end{subfigure}
    \end{minipage}
    \hfill
    \begin{minipage}{0.243\textwidth}
        \centering
        \begin{subfigure}{\linewidth}
            \centering
            \includegraphics[width=1.0\linewidth]{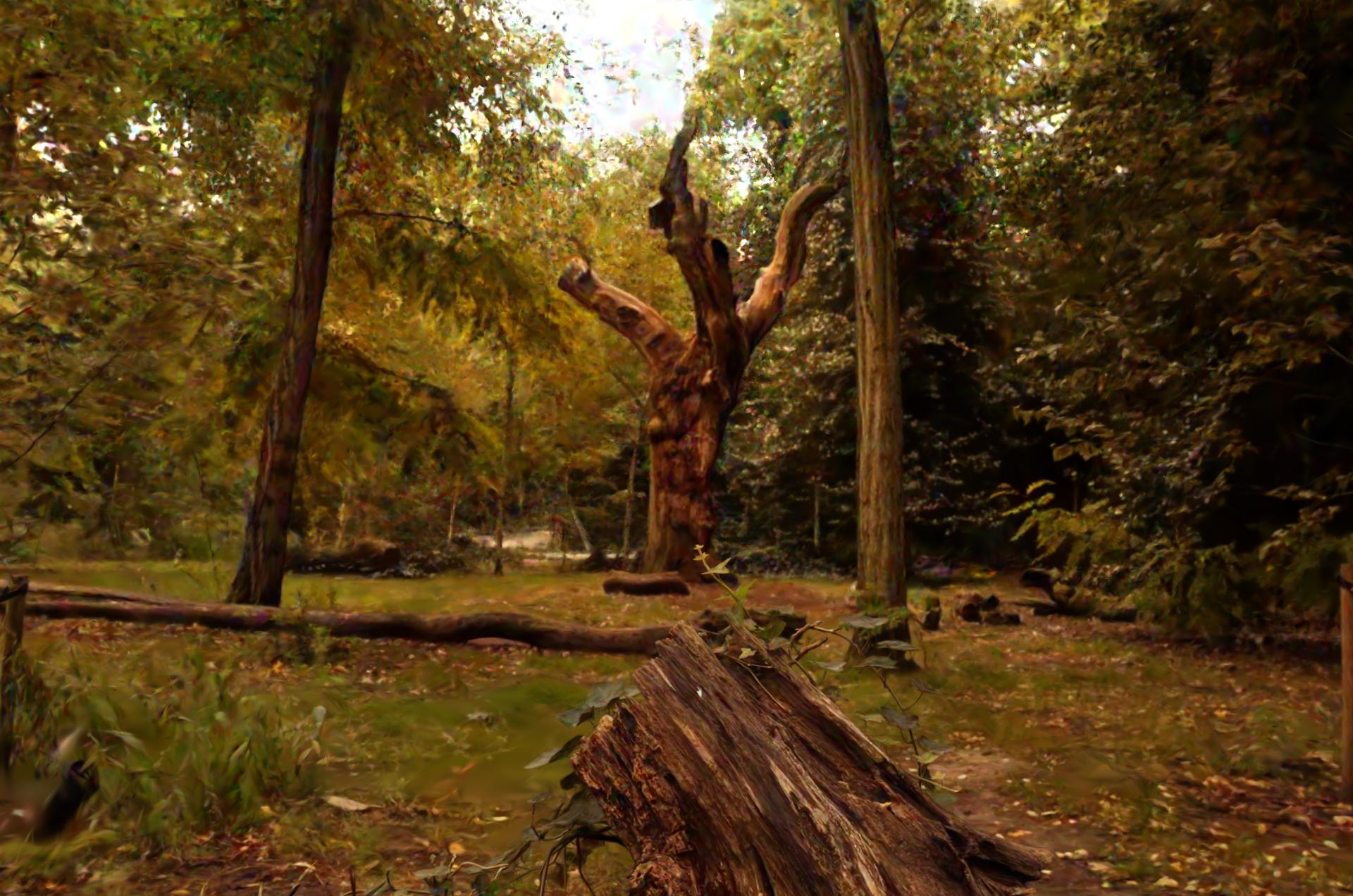}
        \end{subfigure}
    \end{minipage}
    \hfill
    \begin{minipage}{0.243\textwidth}
        \centering
        \begin{subfigure}{\linewidth}
            \centering
            \includegraphics[width=1.0\linewidth]{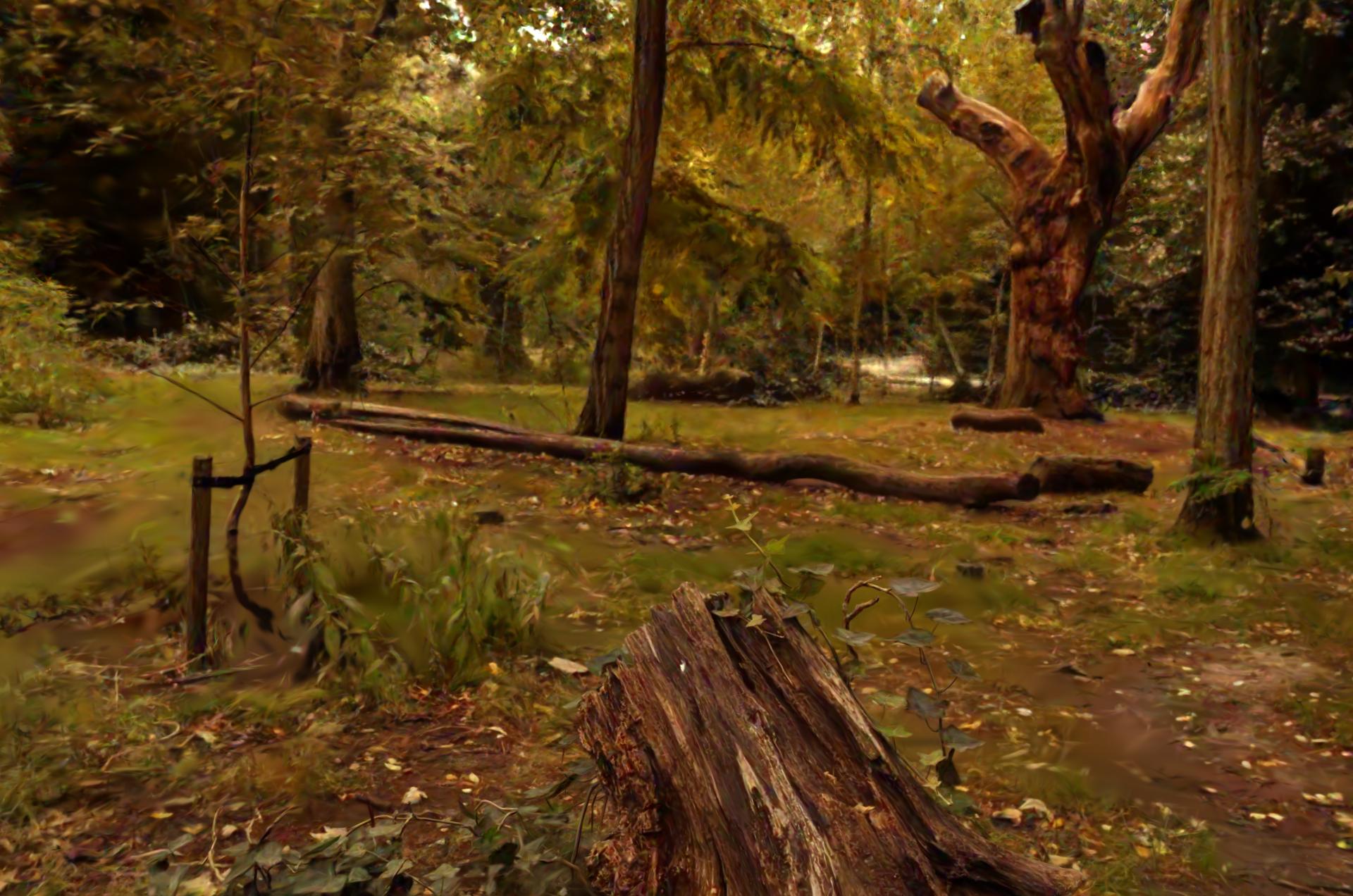}
        \end{subfigure}
    \end{minipage}
    \hfill
    \begin{minipage}{0.243\textwidth}
        \centering
        \begin{subfigure}{\linewidth}
            \centering
            \includegraphics[width=1.0\linewidth]{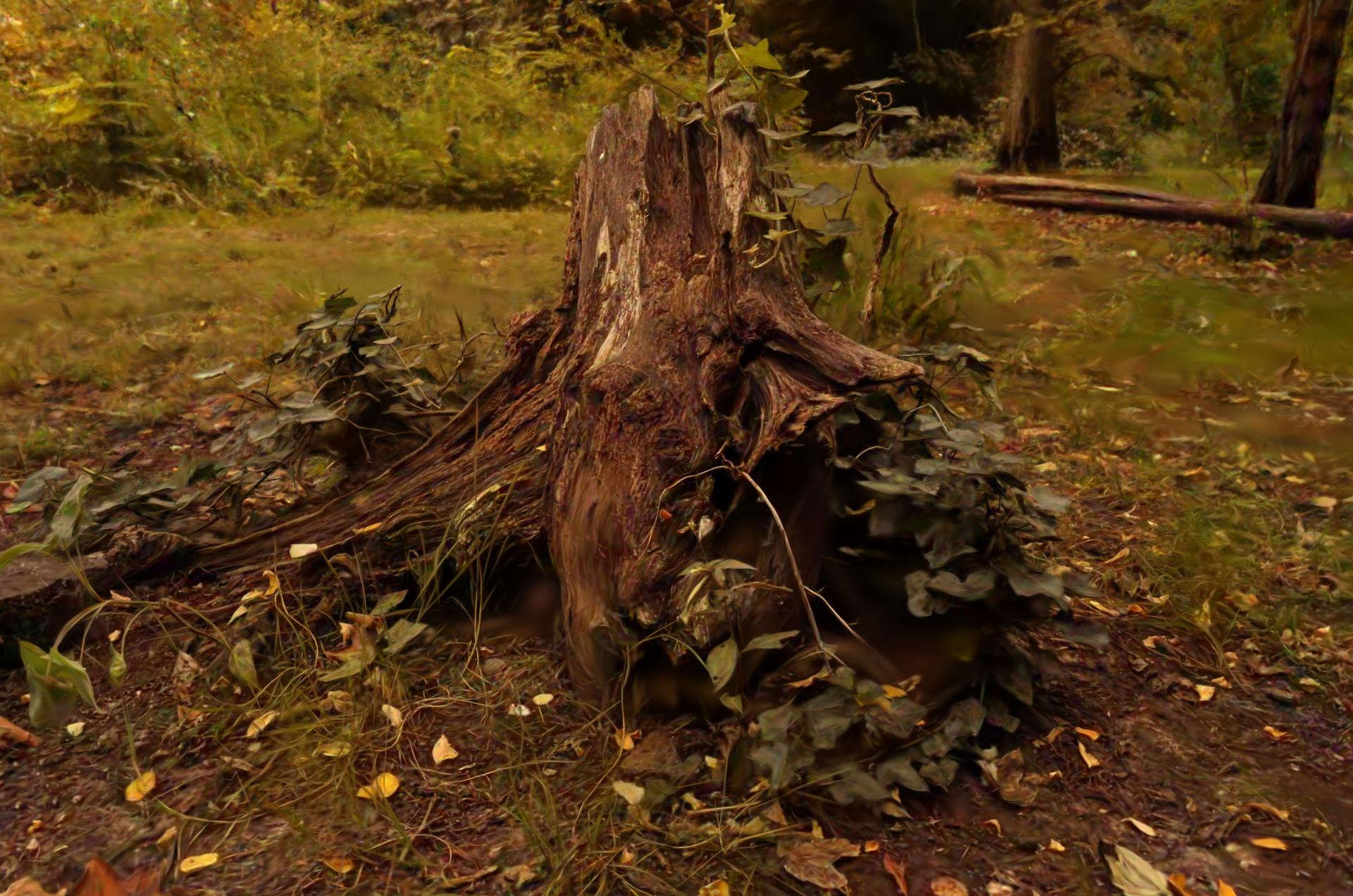}
        \end{subfigure}
    \end{minipage}
    \\ 
    \vspace{1mm}
    \textit{``Make it Autumn"}
    \vspace{1mm}

    \begin{minipage}{0.243\textwidth}
        \centering
        \begin{subfigure}{\linewidth}
            \centering
            \includegraphics[width=1.0\linewidth]{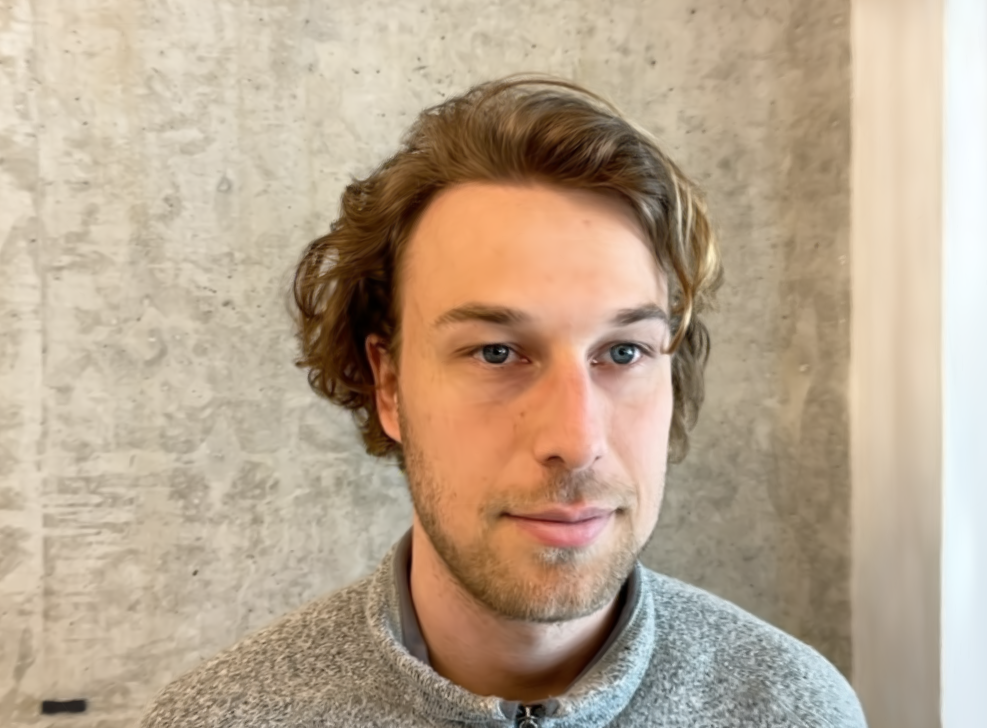}
        \end{subfigure}
    \end{minipage}
    \hfill
    \begin{minipage}{0.243\textwidth}
        \centering
        \begin{subfigure}{\linewidth}
            \centering
            \includegraphics[width=1.0\linewidth]{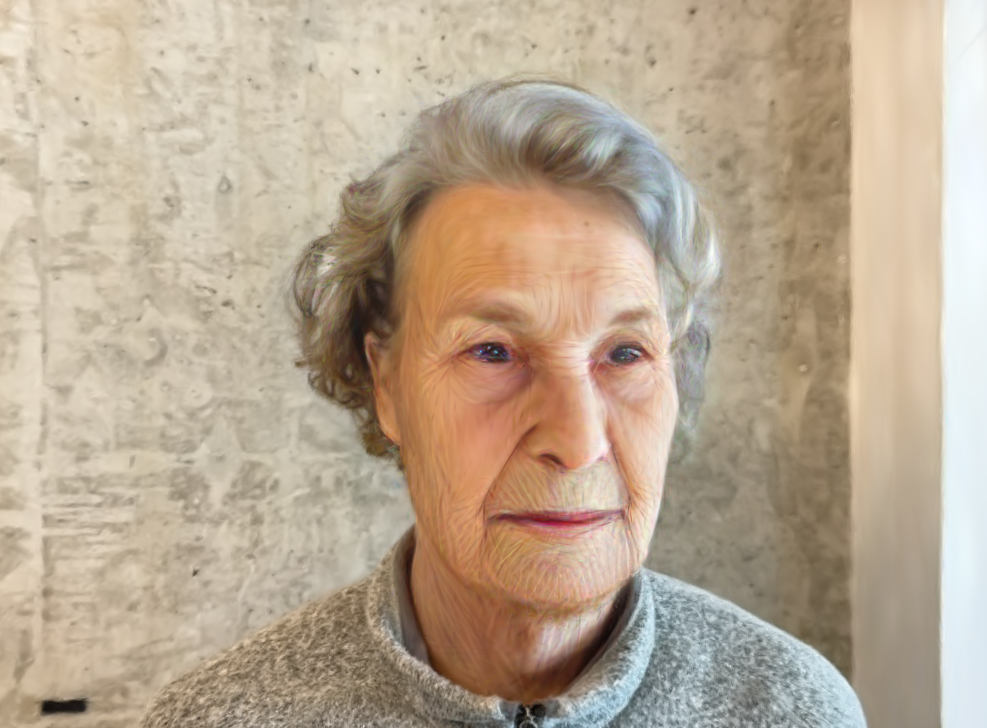}
        \end{subfigure}
    \end{minipage}
    \hfill
    \begin{minipage}{0.243\textwidth}
        \centering
        \begin{subfigure}{\linewidth}
            \centering
            \includegraphics[width=1.0\linewidth]{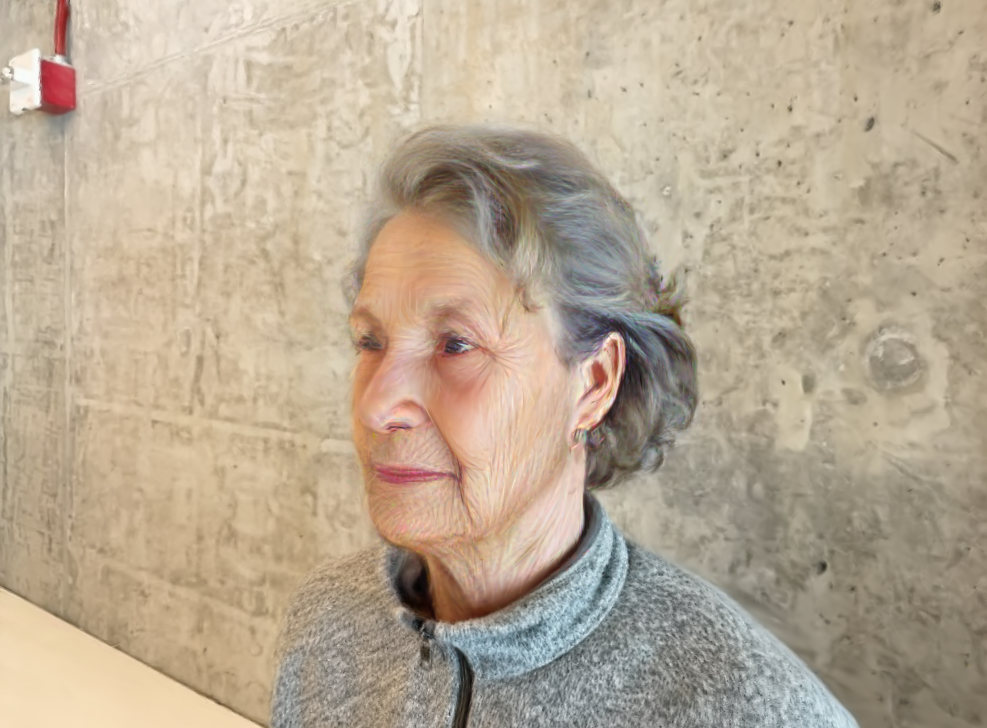}
        \end{subfigure}
    \end{minipage}
    \hfill
    \begin{minipage}{0.243\textwidth}
        \centering
        \begin{subfigure}{\linewidth}
            \centering
            \includegraphics[width=1.0\linewidth]{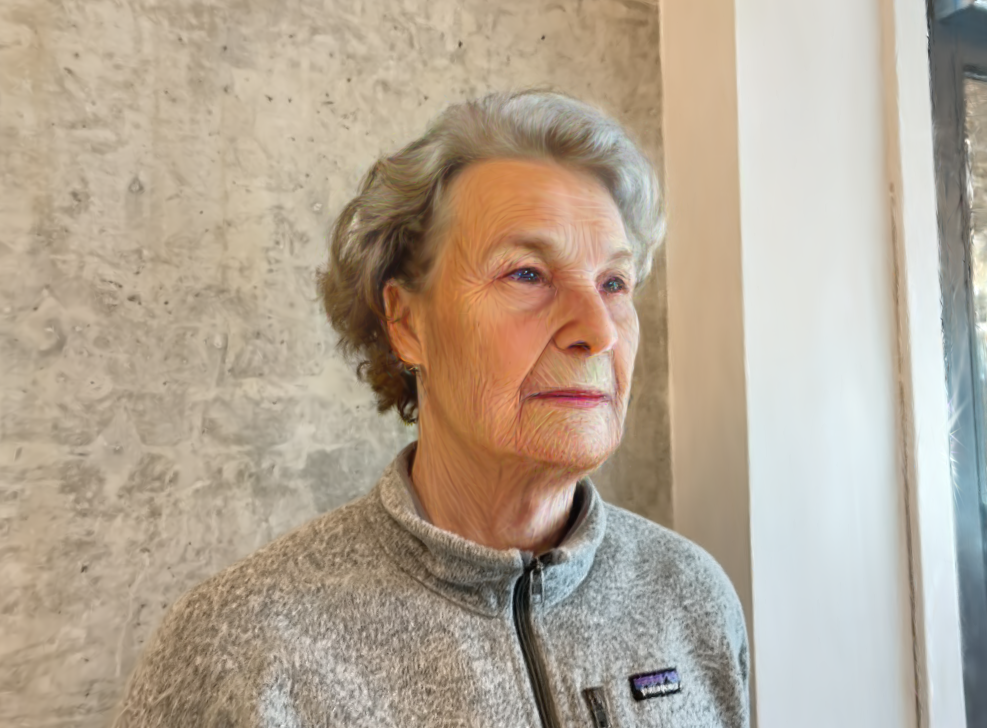}
        \end{subfigure}
    \end{minipage}
    \\ 
    \vspace{1mm}
    \textit{``Turn him into an old lady"}
    \vspace{1mm}

    \begin{minipage}{0.243\textwidth}
        \centering
        \begin{subfigure}{\linewidth}
            \centering
            \includegraphics[width=1.0\linewidth]{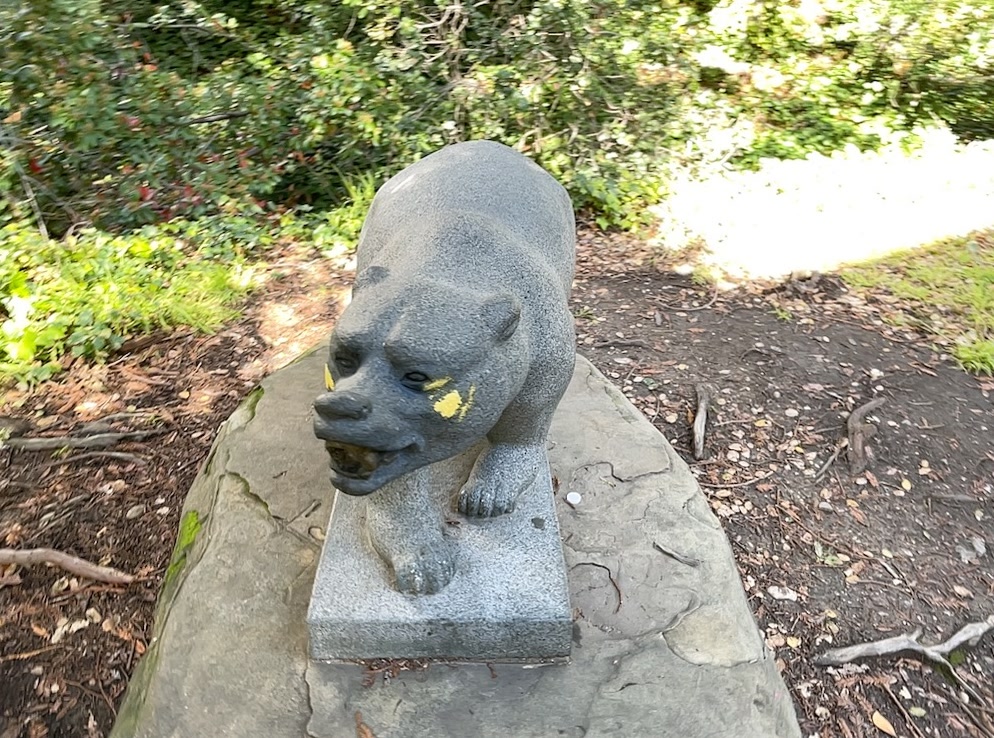}
        \end{subfigure}
    \end{minipage}
    \hfill
    \begin{minipage}{0.243\textwidth}
        \centering
        \begin{subfigure}{\linewidth}
            \centering
            \includegraphics[width=1.0\linewidth]{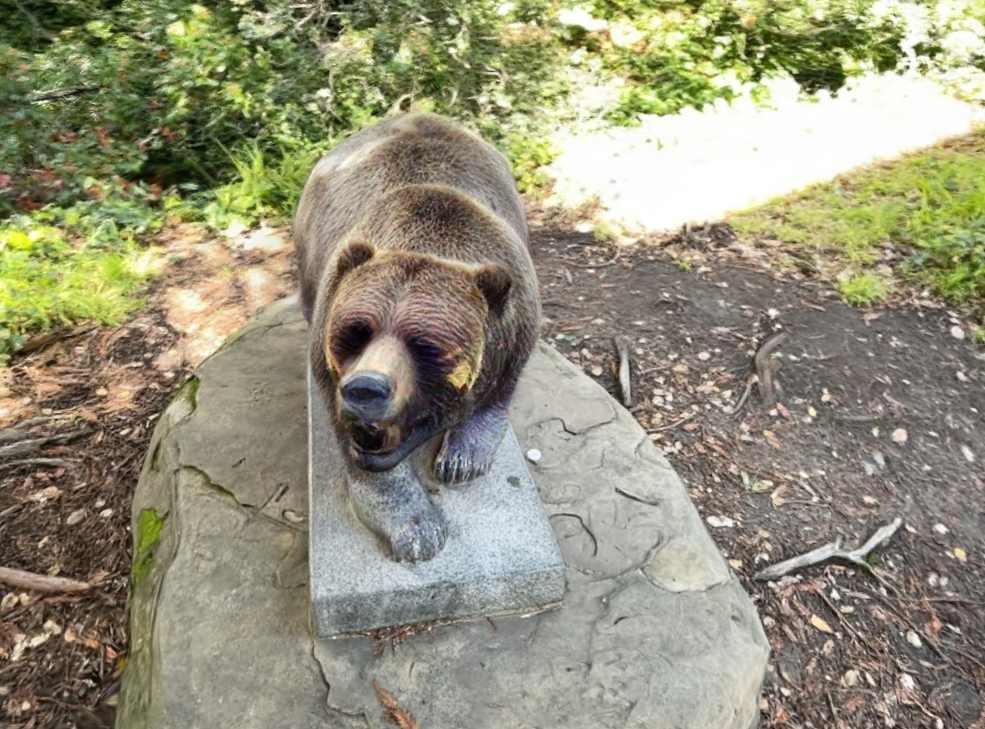}
        \end{subfigure}
    \end{minipage}
    \hfill
    \begin{minipage}{0.243\textwidth}
        \centering
        \begin{subfigure}{\linewidth}
            \centering
            \includegraphics[width=1.0\linewidth]{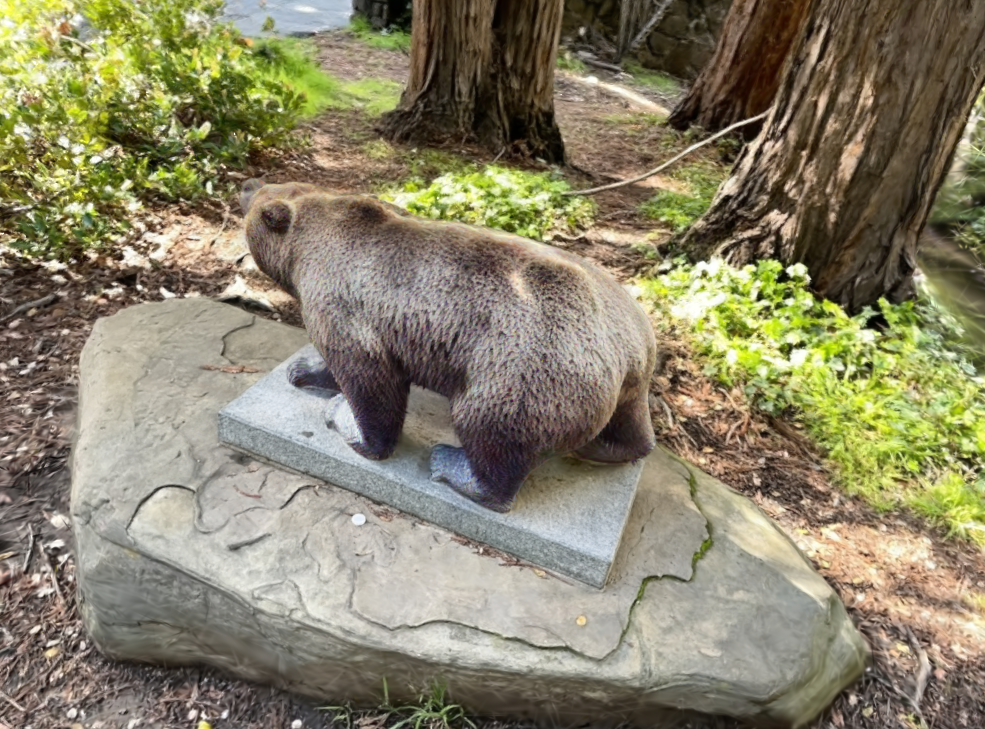}
        \end{subfigure}
    \end{minipage}
    \hfill
    \begin{minipage}{0.243\textwidth}
        \centering
        \begin{subfigure}{\linewidth}
            \centering
            \includegraphics[width=1.0\linewidth]{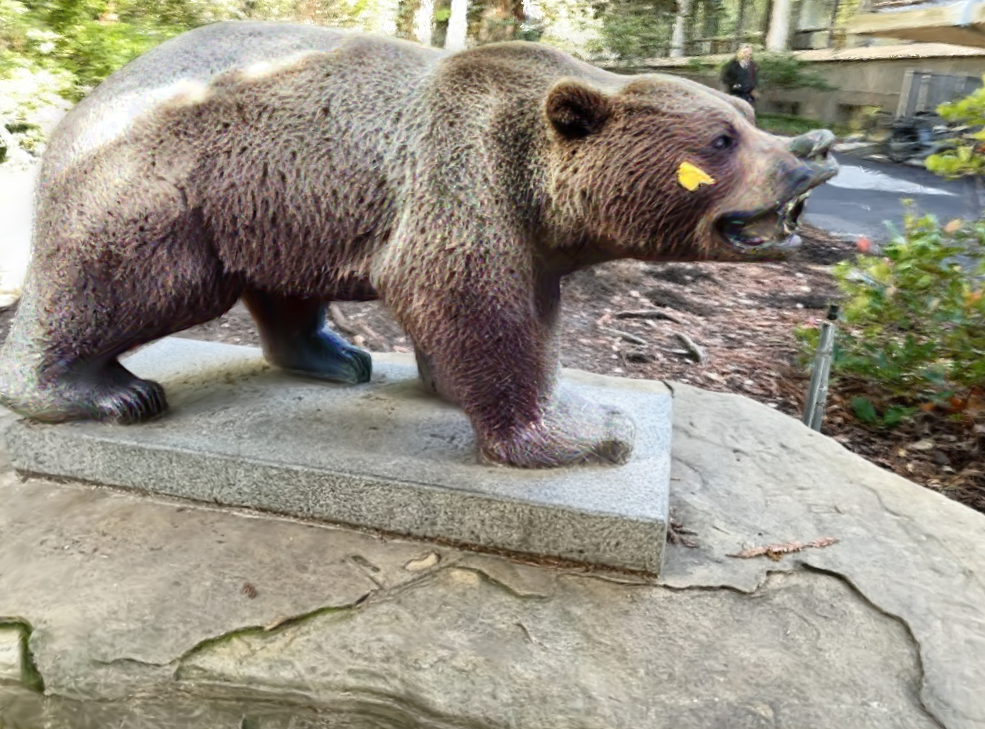}
        \end{subfigure}
    \end{minipage}
    \\ 
    \vspace{1mm}
    \textit{``Turn the bear into a grizzly bear"}
    \caption{\textbf{Extensive Results of GaussianEditor.} Our method is capable of various editing tasks, including face and scene editing. In face and bear editing, we restrict the editing area to the face using Gaussian semantic tracing, ensuring that undesired areas remain unchanged. The leftmost column demonstrates the original view, while the right three columns show the images after editing.}
    \label{fig:qualitive_others}
\end{figure*}

Additionally, to address the challenge of manually determining a densification threshold, we regulate the densification process based on a percentage criterion. In this method, during each densification step, we selectively densify only those Gaussians whose 3D position gradients are within the top $k\%$. This strategy proves to be more manageable and intuitive than directly setting a threshold value in the Hierarchical Gaussian Splatting (HGS) framework.

\subsection{3D Inpainting}
\label{sec: 3D Inpainting}

\textbf{Object Removal.} Simply removing Gaussians identified by a mask can lead to artifacts, especially at the interfaces where the object intersects with other Gaussians. To address this, we employ 2D inpainting techniques to provide guidance for filling these areas. However, effective 2D inpainting requires precise masks to offer better guidance. To generate these masks, after deletion, we use the KNN algorithm to identify Gaussians nearest to the ones removed, which are likely at the interface. These are then projected onto various views. We subsequently dilate the mask and fix any holes to accurately represent the interface area, thus creating a refined mask for the boundary zones. The whole object removal procedure typically takes only two minutes. 

\noindent\textbf{Object Incorporation.} We define this task as follows: Within the 3D Gaussians ${\theta}$, given a camera pose $p$ and the corresponding rendering $I$ from this viewpoint, the user provides a 2D mask $M$ on $I$ indicating the area they wish to inpaint. Additionally, a prompt $y$ is provided to specify the content of the inpainting. We then update ${\theta}$ to fulfill the inpainting request.

Given $I$, $M$, and $y$, the process begins with generating a 2D inpainted image $I^{M}{y}$, utilizing a 2D inpainting diffusion model as per \cite{podell2023sdxl}. Subsequently, the foreground object from $I^{M}{y}$, created by \cite{podell2023sdxl}, is segmented and input into the image-to-3D method referenced in \cite{long2023wonder3d} to generate a coarse 3D mesh. This coarse mesh is then transformed into 3D Gaussians ${\theta}_y$, and refined with HGS detailed in Sec.~\ref{sec: Hierarchical gs}.

For aligning the coordinate system of ${\theta}_y$ with ${\theta}$, the depth of $I^{M}_{y}$ is first estimated using the technique from \cite{ranftl2021vision}. This depth is then aligned with the depth map rendered by ${\theta}$ at camera pose $p$, using the least squares method. With this alignment, we can accurately determine the coordinates and scale of the inpainted foreground object in the coordinate system of ${\theta}$. After transforming ${\theta}_y$ into the coordinate system of ${\theta}$, we simply concatenate them to produce the final inpainted 3D Gaussians.

It is important to note that due to our efficient design, the entire object incorporation procedure can be completed in approximately 5 minutes.

\section{Experiments}
\subsection{Implementation Details}
\label{exp: implement}

\begin{figure}[ht]
    \centering
    \begin{minipage}{0.155\textwidth}
        \centering
        \begin{subfigure}{\linewidth}
            \centering
            \includegraphics[width=1.0\linewidth]{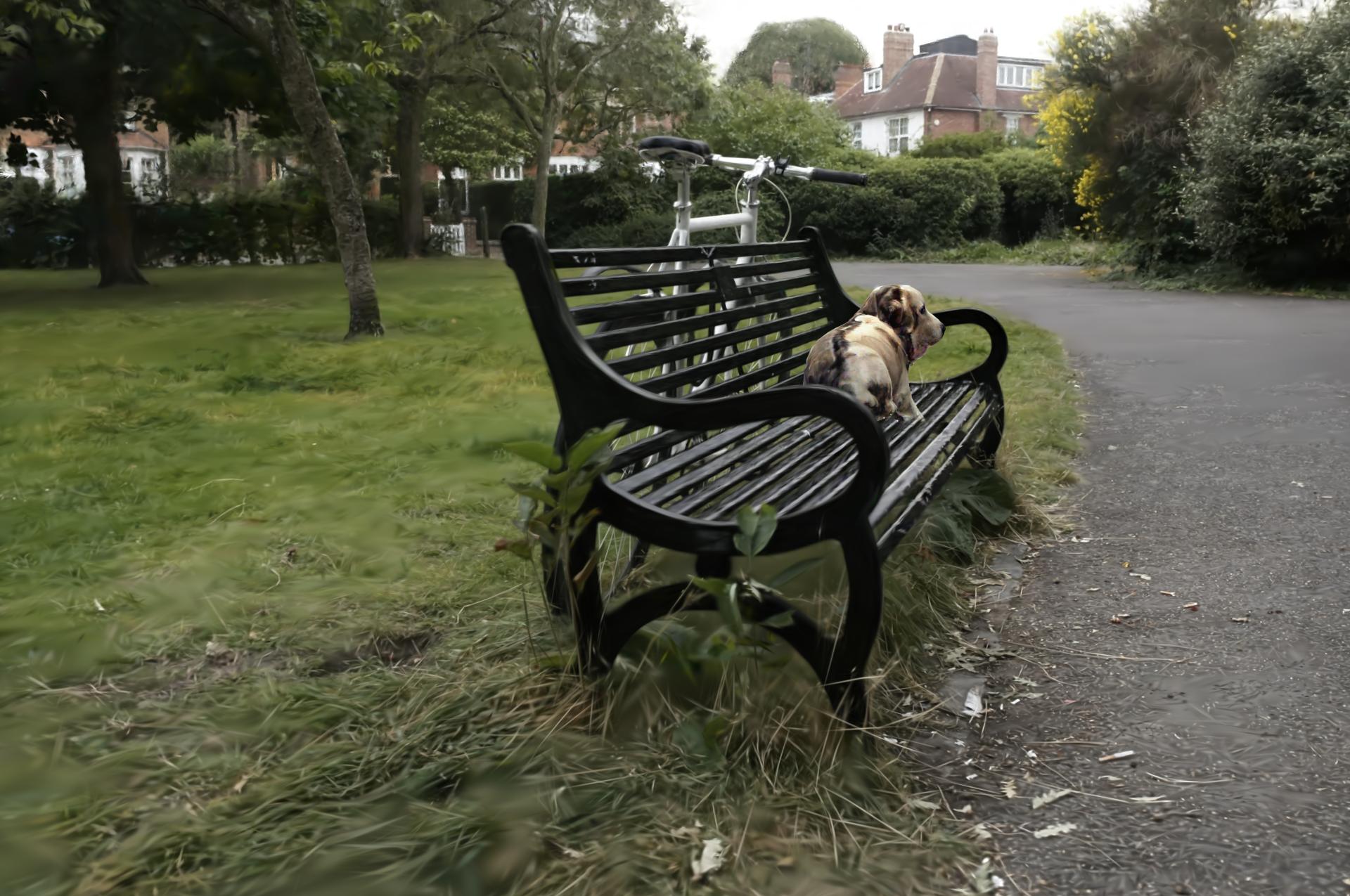}
        \end{subfigure}
    \end{minipage}
    \hfill
    \begin{minipage}{0.155\textwidth}
        \centering
        \begin{subfigure}{\linewidth}
            \centering
            \includegraphics[width=1.0\linewidth]{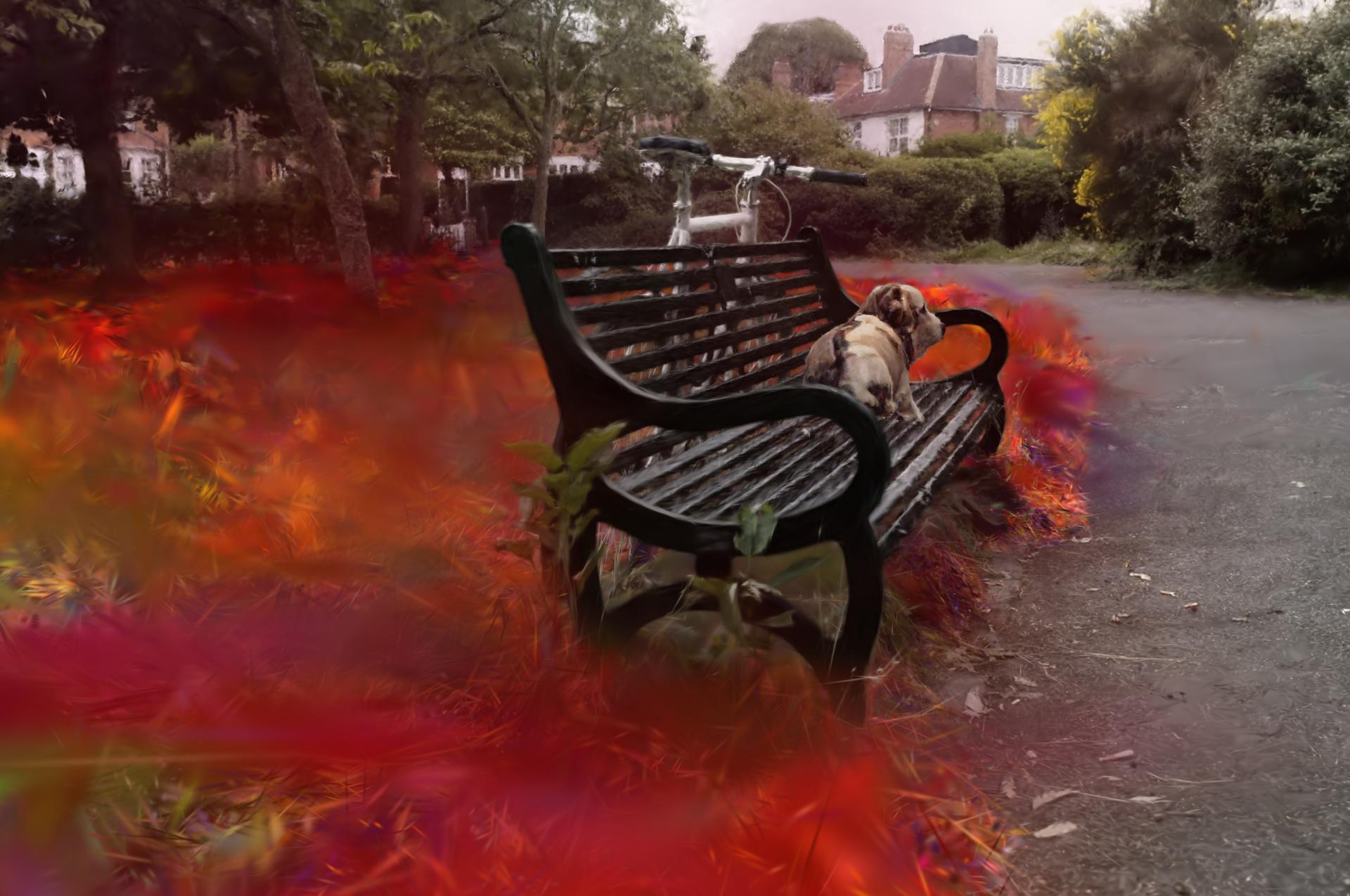}
        \end{subfigure}
    \end{minipage}
    \hfill
    \begin{minipage}{0.155\textwidth}
        \centering
        \begin{subfigure}{\linewidth}
            \centering
            \includegraphics[width=1.0\linewidth]{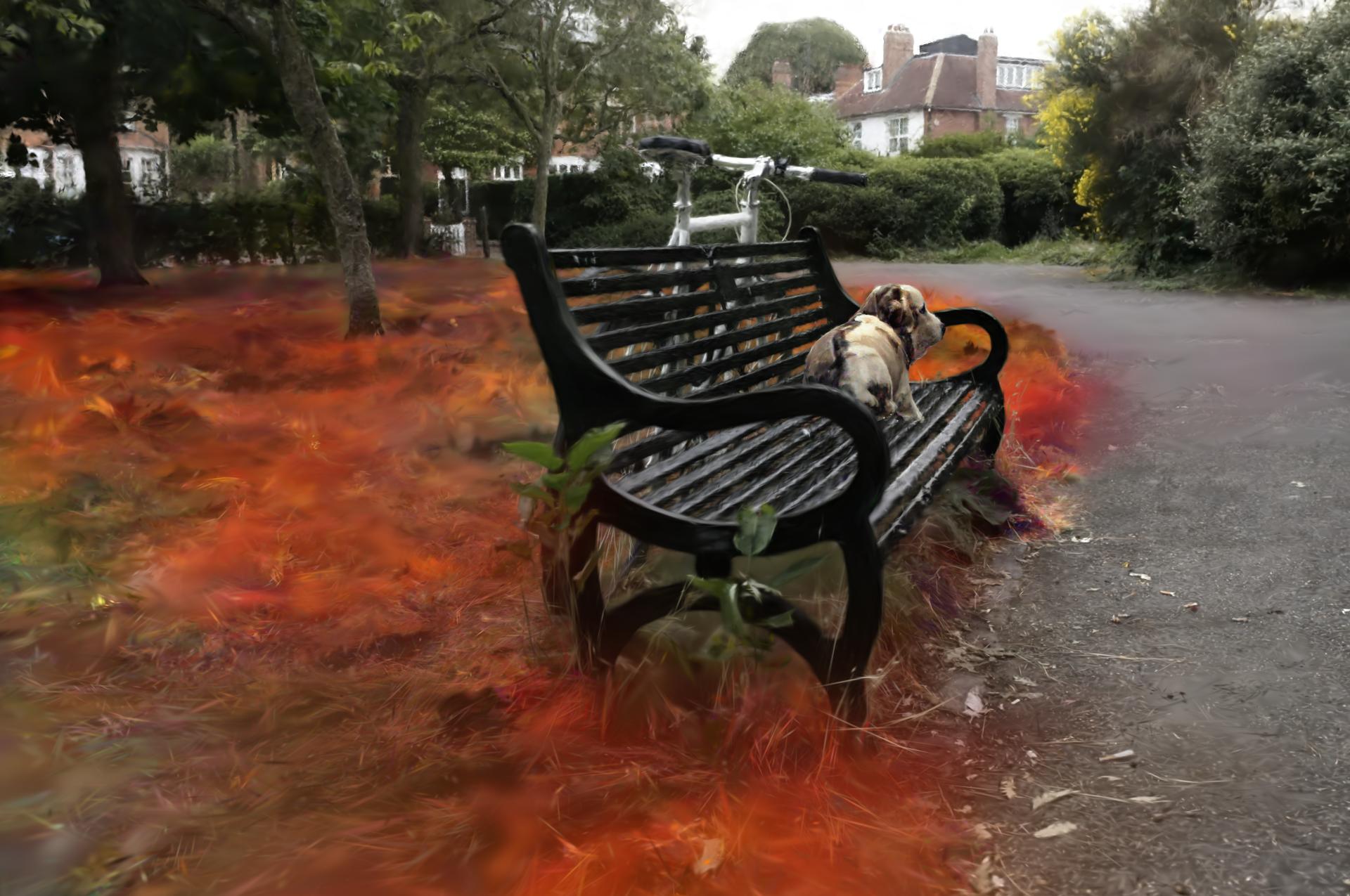}
        \end{subfigure}
    \end{minipage}

    \begin{minipage}{0.155\textwidth}
        \centering
        \begin{subfigure}{\linewidth}
            \centering
            \includegraphics[width=1.0\linewidth]{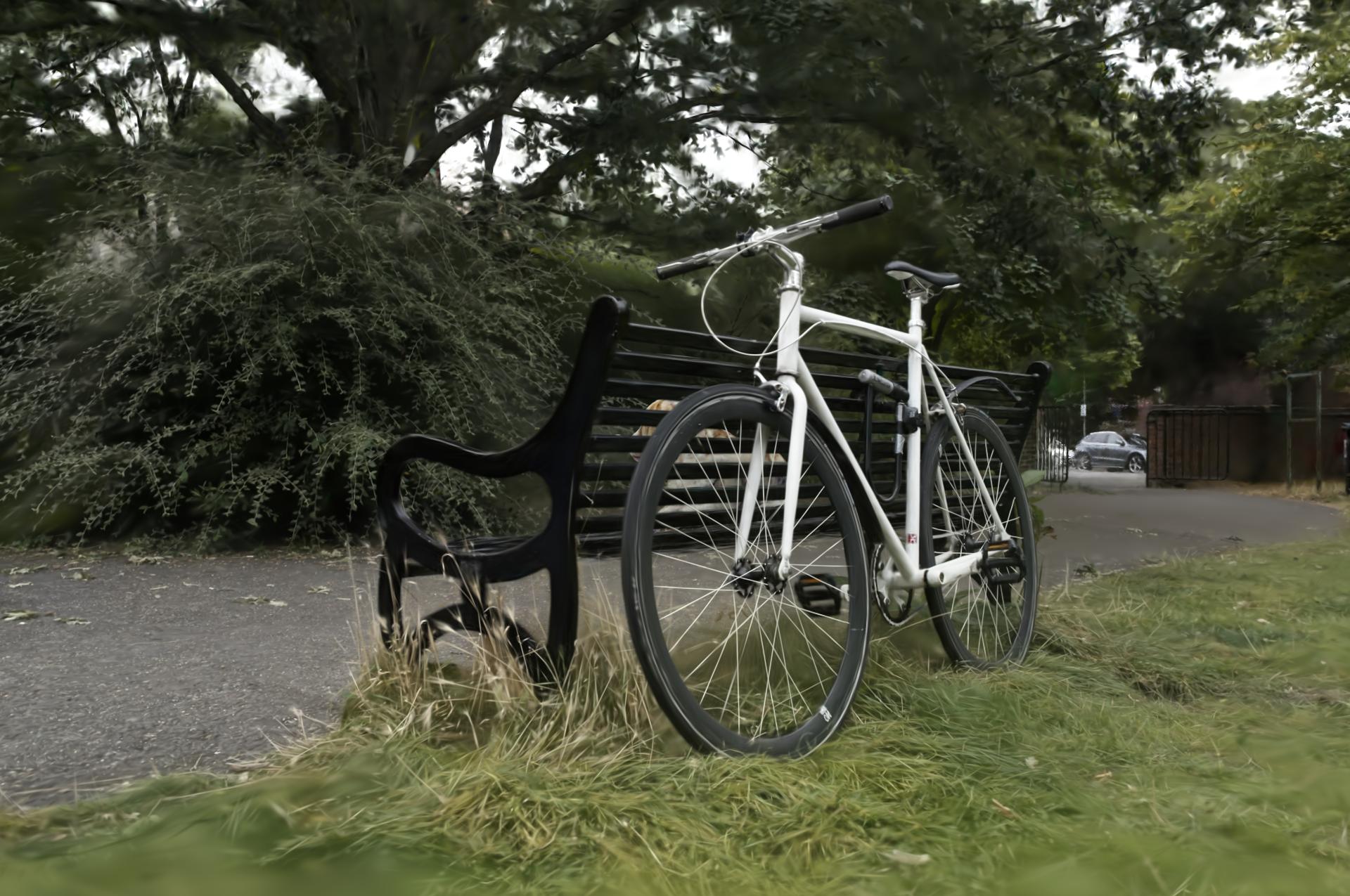}
            \caption*{Original View}
        \end{subfigure}
    \end{minipage}
    \hfill
    \begin{minipage}{0.155\textwidth}
        \centering
        \begin{subfigure}{\linewidth}
            \centering
            \includegraphics[width=1.0\linewidth]{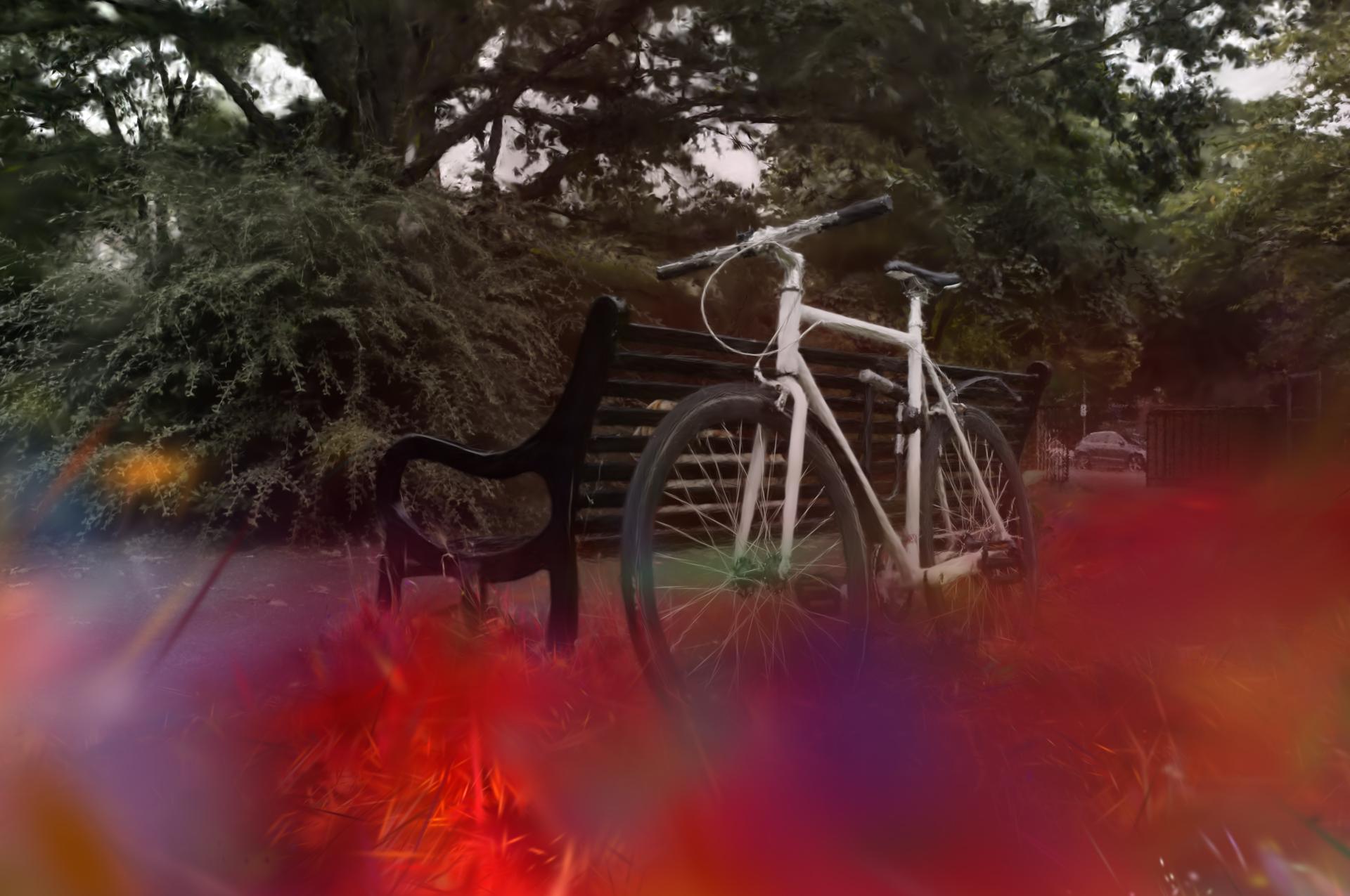}
            \caption*{GS}
        \end{subfigure}
    \end{minipage}
    \hfill
    \begin{minipage}{0.155\textwidth}
        \centering
        \begin{subfigure}{\linewidth}
            \centering
            \includegraphics[width=1.0\linewidth]{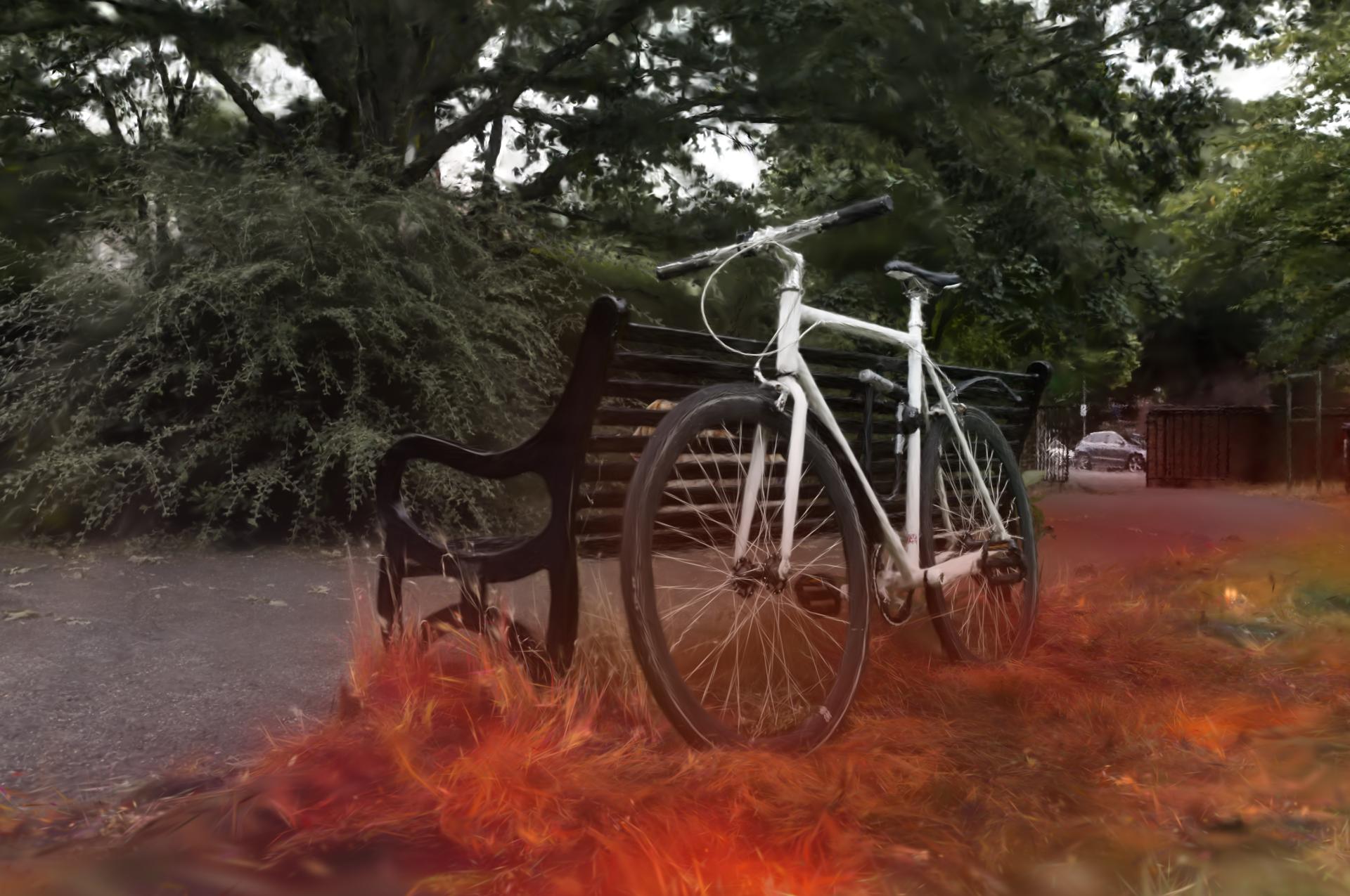}
            \caption*{HGS}
        \end{subfigure}
    \end{minipage}

    \caption{\textbf{Ablation study on Hierarchical Gaussian Splatting (HGS). Prompt: make the grass on fire.} 
Even when specifying the editing area with prompts, generative methods like Instruct-Pix2Pix~\cite{brooks2022instructpix2pix} tend to edit the entire 2D image. Without HGS, Gaussians tend to conform to this whole-image editing by spreading and densifying across the entire scene, leading to uncontrollable densification and blurring of the image. With HGS, however, this kind of diffusion is effectively restrained.}
    \label{fig:ablation}
    \vspace{-3mm}
\end{figure}

\begin{figure}[ht]
    \centering
    \begin{minipage}{0.155\textwidth}
        \centering
        \begin{subfigure}{\linewidth}
            \centering
            \includegraphics[width=1.0\linewidth]{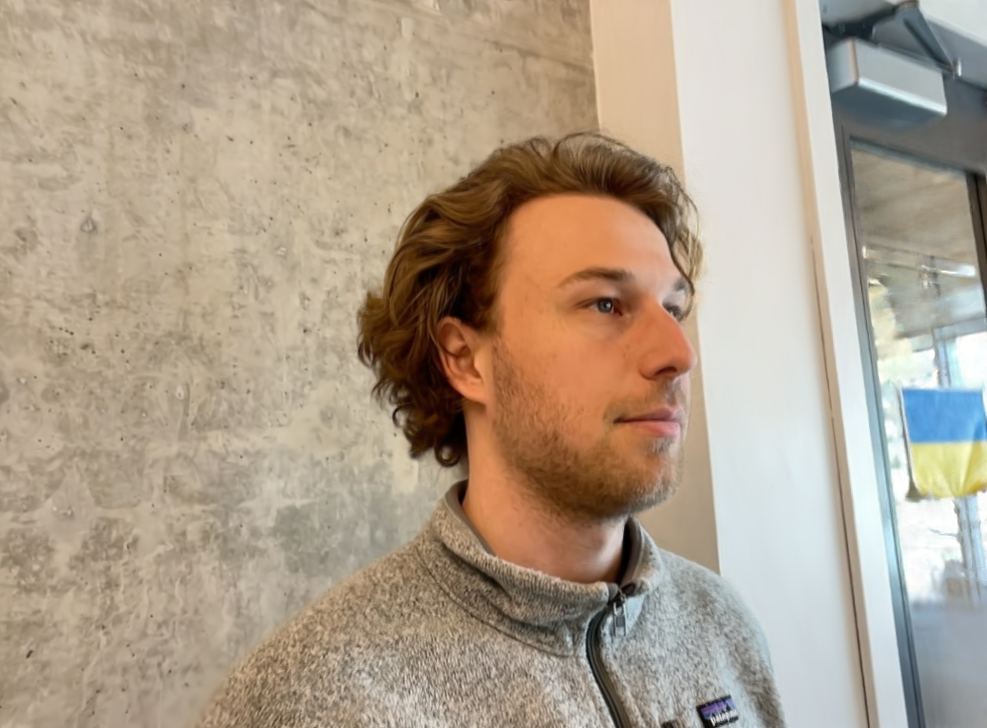}

        \end{subfigure}
    \end{minipage}
    \hfill
    \begin{minipage}{0.155\textwidth}
        \centering
        \begin{subfigure}{\linewidth}
            \centering
            \includegraphics[width=1.0\linewidth]{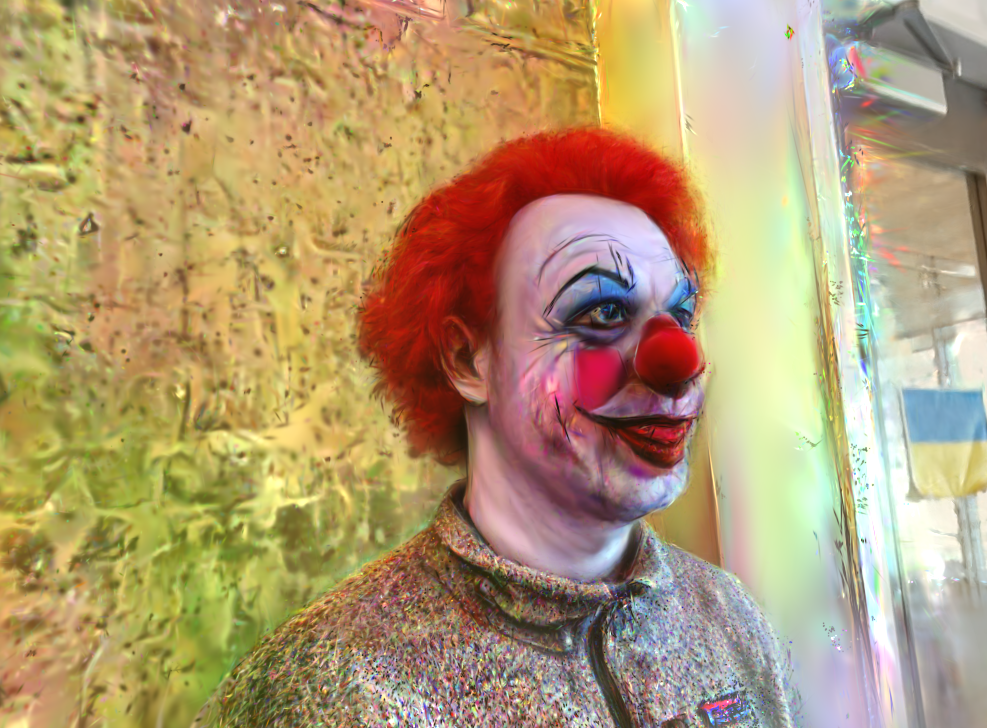}
        \end{subfigure}
    \end{minipage}
    \hfill
    \begin{minipage}{0.155\textwidth}
        \centering
        \begin{subfigure}{\linewidth}
            \centering
            \includegraphics[width=1.0\linewidth]{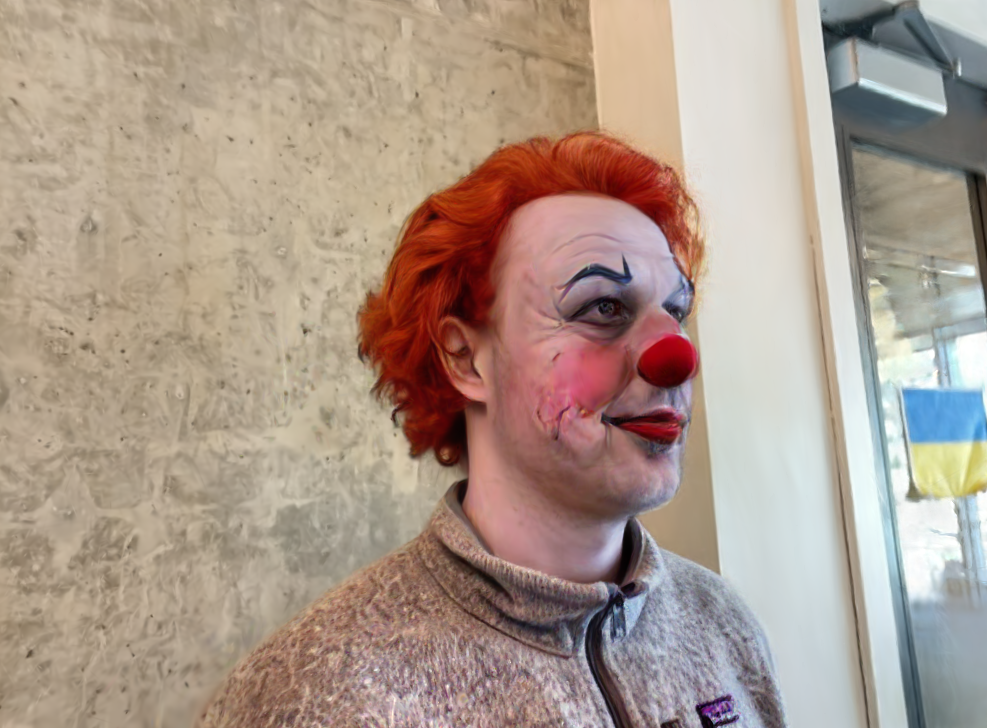}
        \end{subfigure}
    \end{minipage}

    \begin{minipage}{0.155\textwidth}
        \centering
        \begin{subfigure}{\linewidth}
            \centering
            \includegraphics[width=1.0\linewidth]{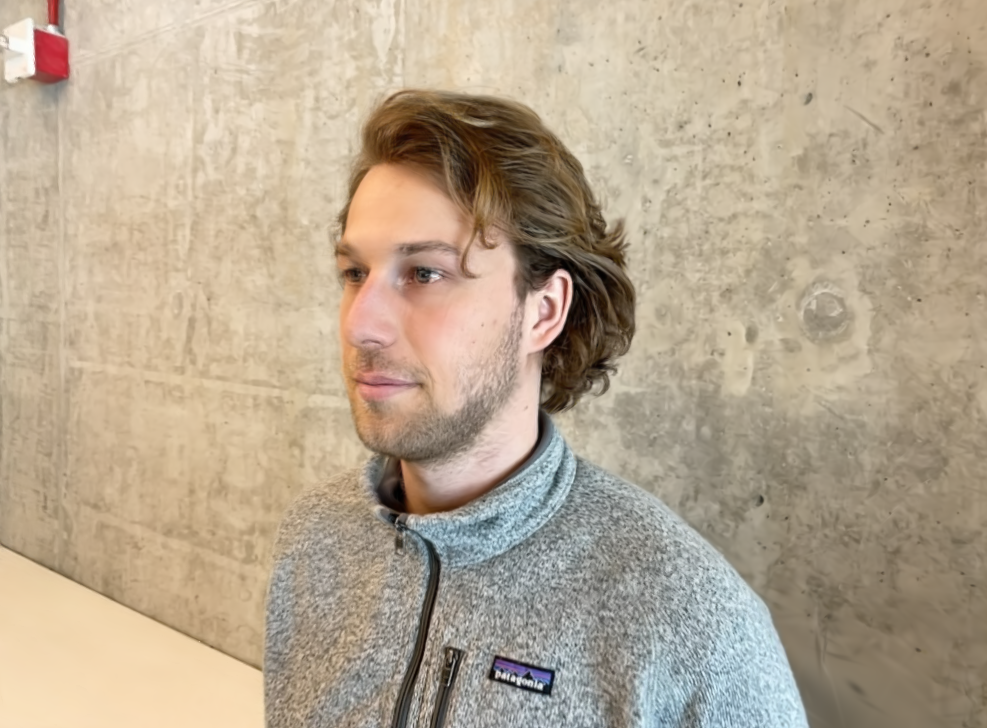}
            \caption*{Original View}
        \end{subfigure}
    \end{minipage}
    \hfill
    \begin{minipage}{0.155\textwidth}
        \centering
        \begin{subfigure}{\linewidth}
            \centering
            \includegraphics[width=1.0\linewidth]{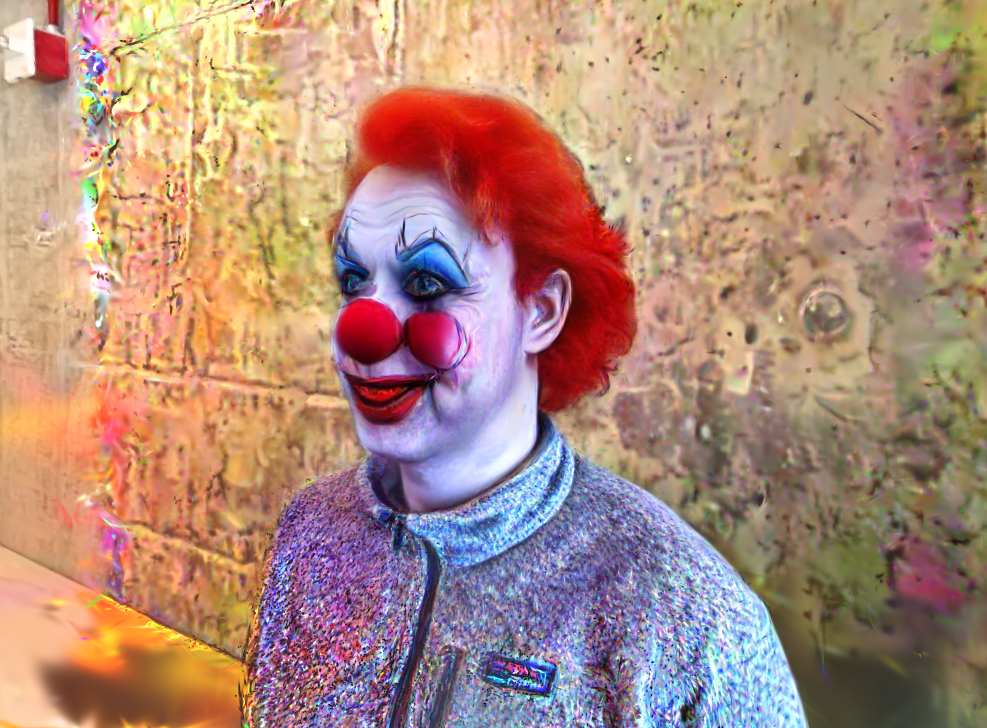}
            \caption*{W/O Semantic Tracing}
        \end{subfigure}
    \end{minipage}
    \hfill
    \begin{minipage}{0.155\textwidth}
        \centering
        \begin{subfigure}{\linewidth}
            \centering
            \includegraphics[width=1.0\linewidth]{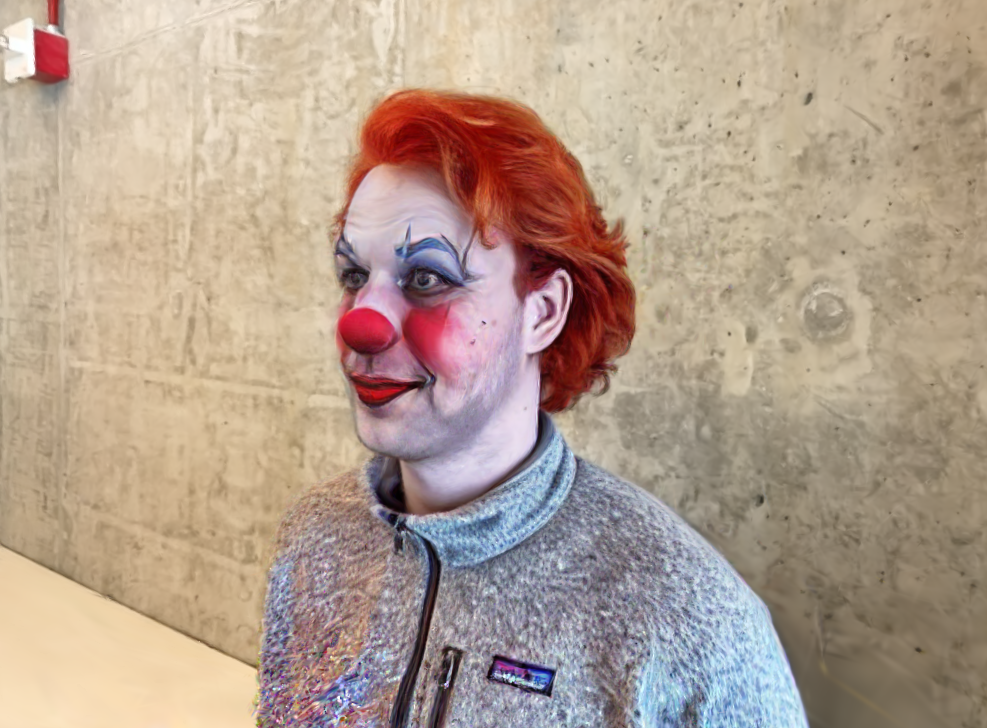}
            \caption*{Semantic Tracing}
        \end{subfigure}
    \end{minipage}

    \caption{\textbf{Ablation study on Semantic Tracing. Tracing Prompt: Man} 
As shown in Fig.\ref{fig:ablation}, generative guidance techniques such as Instruct-Pix2Pix\cite{brooks2022instructpix2pix} exhibit a tendency to modify the entire 2D image. With the assistance of semantic tracing, we can confine the editing region to the desired area.}
    \label{fig:abla_mask}
    \vspace{-3mm}
\end{figure}

We utilize the highly optimized renderer implementation from \cite{kerbl20233d} for Gaussian rendering and base our implementation on Threestudio~\cite{threestudio2023}. All the original 3D Gaussians used in this work are trained using the methods described in \cite{kerbl20233d}. Our experiments are conducted on a single RTX A6000 GPU. As detailed in Sec.~\ref{sec:gs seg}, once we obtain segmentation masks from the 2D segmentation method outlined in \cite{kirillov2023segment}, segmenting the 3D Gaussians takes only about 1 second. 

For editing large scenes, the camera poses employed during the editing process are selected from a subset of the multi-view image dataset initially used for reconstruction. In the case of editing targeted objects, as facilitated by the GS segmentation detailed in Sec.~\ref{sec:gs seg}, we generate a set of camera poses closely surrounding the segmented object. This approach is adopted to increase the resolution of the object in the rendering, thereby enhancing the effectiveness of the editing process. Moreover, when the target object has a low degree of association with the scene, we opt to render only the target object to reduce computational load.

Depending on the complexity of the scene, the number of camera poses used in our experiments varies from 24 to 96. The editing process, influenced by the specified prompt and the complexity of the scene, typically involves optimizing for 500-1000 steps, taking about 5-10 minutes in total.

Regarding 3D inpainting for object incorporation, as detailed in Sec.~\ref{sec: 3D Inpainting}, it takes approximately 3 minutes to generate a 3D mesh using the method from \cite{long2023wonder3d} and an additional 2 minutes to transfer this mesh into 3D Gaussians and refine it, while the composition process of two Gaussians takes less than $1$ second.

\subsection{Qualitative Comparisons}
\label{exp: Qualitative}

As illustrated in Fig.~\ref{fig:comparison}, GaussianEditor-iN2N surpasses other methods in both the quality of edits and controllability. Instruct-Nerf2Nerf, while producing edits with insufficient detail, cannot also control the editing area. GaussianEditor-DDS, due to the more challenging control of guidance offered by DDS loss~\cite{hertz2023delta} compared to Instruct-pix2pix~\cite{brooks2022instructpix2pix}, tends to result in oversaturated colors and less precise editing outcomes.

Additionally, our method exhibits exceptional control over the editing area. This is achieved through Gaussian semantic tracing, which identifies the Gaussians that require editing at each training step, for example, the entire human body in Fig.~\ref{fig:comparison}. It's important to note that in InstructNerf2Nerf~\cite{haque2023instruct}, the use of static 2D or 3D masks restricts the spatial freedom of the edits, as the permissible area for the edited subject is limited by these masks. Furthermore, the effectiveness of static masks diminishes as the geometries and appearances of 3D models evolve during training.

In Fig.~\ref{fig:qualitive_others}, we demonstrate that \name can accommodate a variety of scenarios, such as editing in large-scale scenes and facial swaps. In the case of large scenes, we did not apply Gaussian semantic tracing. However, for facial swaps, we traced the Gaussians corresponding to facial regions, achieving controllable and realistic editing.

We will include additional qualitative results in our supplemental materials to demonstrate the advantages of our method.

\subsection{Quantitative Comparisons}
\label{exp: Quantitative}

\vspace{-3mm}
\begin{table}[ht]
\centering
\resizebox{0.48\textwidth}{!}{
\begin{tabular}{lccc}
\toprule[2pt]
 & iN2N~\cite{haque2023instruct} & Ours~(DDS) & Ours~(iN2N) \\
 \midrule
User study & 15.45\% & 12.27\% & \textbf{72.28\%} \\
CLIP Directional Similarity & 0.1600 & 0.1813 & \textbf{0.2071} \\ 
\bottomrule[2pt]
\end{tabular}
}
\caption{\textbf{Quantitative Comparation.} 
GaussianEditor-iN2N outperforms in both user study evaluations and CLIP Directional Similarity~\cite{gal2022stylegan} metrics.}
\label{table: quantitive_small}
\end{table}
As shown in Table~\ref{table: quantitive_small}, we conduct quantitative comparisons on user study and CLIP directional similarity (as shown in InstructPix2Pix~\cite{brooks2022instructpix2pix} and StyleGAN-Nada~\cite{gal2022stylegan}). GaussianEditor-iN2N not only demonstrates superior outcomes in user studies but also excels in CLIP Directional Similarity. Besides, Instruct-Nerf2Nerf~\cite{haque2023instruct} typically requires more than 30 minutes to complete the editing of a scene, whereas our method only takes between 5 to 10 minutes.

\subsection{Ablation Study}
\label{exp: abla}
As demonstrated in Fig.~\ref{fig:ablation} and Fig.~\ref{fig:abla_mask}, We conducted ablation experiments on Hierarchical Gaussian Splatting(HGS) and Semantic Tracing.
Without HGS, Gaussians tend to spread and densify across the scene, leading to uncontrolled densification and image blurring. 
This is typically caused by the tendency of methods like Instruct-Pix2Pix to edit the entire 2D image when prompts are used to define editing areas. However, HGS effectively circumvents this issue by constraining the mobility of the Gaussian in the old generation, ensuring that the overall scene does not exhibit excessive mobility. On the other hand, Semantic tracing assists GaussianEditor in limiting editing to a specified area without restricting the expansiveness of the editing region.

\section{Conclusion}

In our research, we introduce \name, an innovative 3D editing algorithm based on Gaussian Splatting, designed for enhanced control and efficiency.
Our method employs Gaussian semantic tracing for precise identification and targeting of editing areas, followed by Hierarchical Gaussian Splatting (HGS) to balance fluidity and stability in achieving detailed results under stochastic guidance. Additionally, we developed a specialized 3D inpainting algorithm for Gaussian Splatting, streamlining object removal and integration, and greatly reducing editing time.

\textbf{Limitation.} Similar to previous 3D editing works based on 2D diffusion models, \name relies on these models to provide effective supervision. However, current 2D diffusion models struggle to offer effective guidance for certain complex prompts, leading to limitations in 3D editing.

{
    \small
    \bibliographystyle{ieeenat_fullname}
    \bibliography{main}

\begin{thebibliography}{58}
\providecommand{\natexlab}[1]{#1}
\providecommand{\url}[1]{\texttt{#1}}
\expandafter\ifx\csname urlstyle\endcsname\relax
  \providecommand{\doi}[1]{doi: #1}\else
  \providecommand{\doi}{doi: \begingroup \urlstyle{rm}\Url}\fi

\bibitem[Bao et~al.(2023)Bao, Zhang, and Yang]{bao2023sine}
Chong Bao, Yinda Zhang, and Bangbang et~al. Yang.
\newblock Sine: Semantic-driven image-based nerf editing with prior-guided editing field.
\newblock In \emph{CVPR 2023}, pages 20919--20929, 2023.

\bibitem[Barron et~al.(2021)Barron, Mildenhall, Tancik, Hedman, Martin-Brualla, and Srinivasan]{barron2021mip}
Jonathan~T Barron, Ben Mildenhall, Matthew Tancik, Peter Hedman, Ricardo Martin-Brualla, and Pratul~P Srinivasan.
\newblock Mip-nerf: A multiscale representation for anti-aliasing neural radiance fields.
\newblock In \emph{Proceedings of the IEEE/CVF International Conference on Computer Vision}, pages 5855--5864, 2021.

\bibitem[Barron et~al.(2022)Barron, Mildenhall, Verbin, Srinivasan, and Hedman]{barron2022mipnerf360}
Jonathan~T. Barron, Ben Mildenhall, Dor Verbin, Pratul~P. Srinivasan, and Peter Hedman.
\newblock Mip-nerf 360: Unbounded anti-aliased neural radiance fields.
\newblock \emph{CVPR}, 2022.

\bibitem[Brooks et~al.(2022)Brooks, Holynski, and Efros]{brooks2022instructpix2pix}
Tim Brooks, Aleksander Holynski, and Alexei~A Efros.
\newblock Instructpix2pix: Learning to follow image editing instructions.
\newblock \emph{arXiv preprint arXiv:2211.09800}, 2022.

\bibitem[Chen et~al.(2023{\natexlab{a}})Chen, Zhang, Yang, Cai, Yu, Yang, and Lin]{chen2023it3d}
Yiwen Chen, Chi Zhang, Xiaofeng Yang, Zhongang Cai, Gang Yu, Lei Yang, and Guosheng Lin.
\newblock It3d: Improved text-to-3d generation with explicit view synthesis.
\newblock \emph{arXiv preprint arXiv:2308.11473}, 2023{\natexlab{a}}.

\bibitem[Chen et~al.(2022)Chen, Funkhouser, Hedman, and Tagliasacchi]{chen2022mobilenerf}
Zhiqin Chen, Thomas Funkhouser, Peter Hedman, and Andrea Tagliasacchi.
\newblock Mobilenerf: Exploiting the polygon rasterization pipeline for efficient neural field rendering on mobile architectures.
\newblock \emph{arXiv preprint arXiv:2208.00277}, 2022.

\bibitem[Chen et~al.(2023{\natexlab{b}})Chen, Wang, and Liu]{chen2023text}
Zilong Chen, Feng Wang, and Huaping Liu.
\newblock Text-to-3d using gaussian splatting.
\newblock \emph{arXiv preprint arXiv:2309.16585}, 2023{\natexlab{b}}.

\bibitem[Cheng et~al.(2023)Cheng, Yang, Wang, Li, Zhang, Zhang, and Yuan]{cheng2023progressive3d}
Xinhua Cheng, Tianyu Yang, Jianan Wang, Yu Li, Lei Zhang, Jian Zhang, and Li Yuan.
\newblock Progressive3d: Progressively local editing for text-to-3d content creation with complex semantic prompts.
\newblock \emph{arXiv preprint arXiv:2310.11784}, 2023.

\bibitem[Gal et~al.(2022)Gal, Patashnik, Maron, Bermano, Chechik, and Cohen-Or]{gal2022stylegan}
Rinon Gal, Or Patashnik, Haggai Maron, Amit~H Bermano, Gal Chechik, and Daniel Cohen-Or.
\newblock Stylegan-nada: Clip-guided domain adaptation of image generators.
\newblock \emph{ACM Transactions on Graphics (TOG)}, 41\penalty0 (4):\penalty0 1--13, 2022.

\bibitem[Gao et~al.(2023)Gao, Aigerman, Groueix, Kim, and Hanocka]{gao2023textdeformer}
William Gao, Noam Aigerman, Thibault Groueix, Vladimir~G Kim, and Rana Hanocka.
\newblock Textdeformer: Geometry manipulation using text guidance.
\newblock \emph{arXiv preprint arXiv:2304.13348}, 2023.

\bibitem[Guo et~al.(2023)Guo, Liu, Wang, Zou, Luo, Chen, Cao, and Zhang]{threestudio2023}
Yuan-Chen Guo, Ying-Tian Liu, Chen Wang, Zi-Xin Zou, Guan Luo, Chia-Hao Chen, Yan-Pei Cao, and Song-Hai Zhang.
\newblock threestudio: A unified framework for 3d content generation.
\newblock \url{https://github.com/threestudio-project/threestudio}, 2023.

\bibitem[Haque et~al.(2023)Haque, Tancik, Efros, Holynski, and Kanazawa]{haque2023instruct}
Ayaan Haque, Matthew Tancik, Alexei~A Efros, Aleksander Holynski, and Angjoo Kanazawa.
\newblock Instruct-nerf2nerf: Editing 3d scenes with instructions.
\newblock \emph{arXiv preprint arXiv:2303.12789}, 2023.

\bibitem[Hedman et~al.(2021)Hedman, Srinivasan, Mildenhall, Barron, and Debevec]{hedman2021snerg}
Peter Hedman, Pratul~P. Srinivasan, Ben Mildenhall, Jonathan~T. Barron, and Paul Debevec.
\newblock Baking neural radiance fields for real-time view synthesis.
\newblock \emph{ICCV}, 2021.

\bibitem[Hertz et~al.(2023)Hertz, Aberman, and Cohen-Or]{hertz2023delta}
Amir Hertz, Kfir Aberman, and Daniel Cohen-Or.
\newblock Delta denoising score.
\newblock In \emph{Proceedings of the IEEE/CVF International Conference on Computer Vision}, pages 2328--2337, 2023.

\bibitem[Kerbl et~al.(2023{\natexlab{a}})Kerbl, Kopanas, Leimk{\"u}hler, and Drettakis]{3dgs}
Bernhard Kerbl, Georgios Kopanas, Thomas Leimk{\"u}hler, and George Drettakis.
\newblock 3d gaussian splatting for real-time radiance field rendering.
\newblock \emph{ACM Transactions on Graphics (ToG)}, 42\penalty0 (4):\penalty0 1--14, 2023{\natexlab{a}}.

\bibitem[Kerbl et~al.(2023{\natexlab{b}})Kerbl, Kopanas, Leimk{\"u}hler, and Drettakis]{kerbl20233d}
Bernhard Kerbl, Georgios Kopanas, Thomas Leimk{\"u}hler, and George Drettakis.
\newblock 3d gaussian splatting for real-time radiance field rendering.
\newblock \emph{ToG}, 42\penalty0 (4):\penalty0 1--14, 2023{\natexlab{b}}.

\bibitem[Kerbl et~al.(2023{\natexlab{c}})Kerbl, Kopanas, Leimk{\"u}hler, and Drettakis]{kerbl3Dgaussians}
Bernhard Kerbl, Georgios Kopanas, Thomas Leimk{\"u}hler, and George Drettakis.
\newblock 3d gaussian splatting for real-time radiance field rendering.
\newblock \emph{ACM Transactions on Graphics}, 42\penalty0 (4), 2023{\natexlab{c}}.

\bibitem[Kirillov et~al.(2023)Kirillov, Mintun, Ravi, Mao, Rolland, Gustafson, Xiao, Whitehead, Berg, Lo, et~al.]{kirillov2023segment}
Alexander Kirillov, Eric Mintun, Nikhila Ravi, Hanzi Mao, Chloe Rolland, Laura Gustafson, Tete Xiao, Spencer Whitehead, Alexander~C Berg, Wan-Yen Lo, et~al.
\newblock Segment anything.
\newblock \emph{arXiv preprint arXiv:2304.02643}, 2023.

\bibitem[Kobayashi et~al.(2022)Kobayashi, Matsumoto, and Sitzmann]{kobayashi2022decomposing}
Sosuke Kobayashi, Eiichi Matsumoto, and Vincent Sitzmann.
\newblock Decomposing nerf for editing via feature field distillation.
\newblock \emph{arXiv preprint arXiv:2205.15585}, 2022.

\bibitem[Li et~al.(2022)Li, Lin, Forsyth, Huang, and Wang]{li2022climatenerf}
Yuan Li, Zhi-Hao Lin, David Forsyth, Jia-Bin Huang, and Shenlong Wang.
\newblock Climatenerf: Physically-based neural rendering for extreme climate synthesis.
\newblock \emph{arXiv e-prints}, pages arXiv--2211, 2022.

\bibitem[Li et~al.(2023)Li, M\"uller, Evans, Taylor, Unberath, Liu, and Lin]{li2023neuralangelo}
Zhaoshuo Li, Thomas M\"uller, Alex Evans, Russell~H Taylor, Mathias Unberath, Ming-Yu Liu, and Chen-Hsuan Lin.
\newblock Neuralangelo: High-fidelity neural surface reconstruction.
\newblock In \emph{CVPR}, 2023.

\bibitem[Lin et~al.(2023)Lin, Gao, Tang, Takikawa, Zeng, Huang, Kreis, Fidler, Liu, and Lin]{lin2023magic3d}
Chen-Hsuan Lin, Jun Gao, Luming Tang, Towaki Takikawa, Xiaohui Zeng, Xun Huang, Karsten Kreis, Sanja Fidler, Ming-Yu Liu, and Tsung-Yi Lin.
\newblock Magic3d: High-resolution text-to-3d content creation.
\newblock In \emph{CVPR}, pages 300--309, 2023.

\bibitem[Liu et~al.(2022)Liu, Shen, Chen, et~al.]{liu2022nerf}
Hao-Kang Liu, I Shen, Bing-Yu Chen, et~al.
\newblock Nerf-in: Free-form nerf inpainting with rgb-d priors.
\newblock \emph{arXiv preprint arXiv:2206.04901}, 2022.

\bibitem[Liu et~al.(2021)Liu, Zhang, Zhang, Zhang, Zhu, and Russell]{liu2021editing}
Steven Liu, Xiuming Zhang, Zhoutong Zhang, Richard Zhang, Jun-Yan Zhu, and Bryan Russell.
\newblock Editing conditional radiance fields.
\newblock In \emph{ICCV 2021}, pages 5773--5783, 2021.

\bibitem[Long et~al.(2023)Long, Guo, Lin, Liu, Dou, Liu, Ma, Zhang, Habermann, Theobalt, et~al.]{long2023wonder3d}
Xiaoxiao Long, Yuan-Chen Guo, Cheng Lin, Yuan Liu, Zhiyang Dou, Lingjie Liu, Yuexin Ma, Song-Hai Zhang, Marc Habermann, Christian Theobalt, et~al.
\newblock Wonder3d: Single image to 3d using cross-domain diffusion.
\newblock \emph{arXiv preprint arXiv:2310.15008}, 2023.

\bibitem[Luiten et~al.(2023)Luiten, Kopanas, Leibe, and Ramanan]{luiten2023dynamic}
Jonathon Luiten, Georgios Kopanas, Bastian Leibe, and Deva Ramanan.
\newblock Dynamic 3d gaussians: Tracking by persistent dynamic view synthesis.
\newblock \emph{arXiv preprint arXiv:2308.09713}, 2023.

\bibitem[Mikaeili et~al.(2023)Mikaeili, Perel, Safaee, Cohen-Or, and Mahdavi-Amiri]{mikaeili2023sked}
Aryan Mikaeili, Or Perel, Mehdi Safaee, Daniel Cohen-Or, and Ali Mahdavi-Amiri.
\newblock Sked: Sketch-guided text-based 3d editing.
\newblock In \emph{Proceedings of the IEEE/CVF International Conference on Computer Vision}, pages 14607--14619, 2023.

\bibitem[Mildenhall et~al.(2020)Mildenhall, Srinivasan, Tancik, Barron, Ramamoorthi, and Ng]{nerf}
Ben Mildenhall, Pratul~P. Srinivasan, Matthew Tancik, Jonathan~T. Barron, Ravi Ramamoorthi, and Ren Ng.
\newblock Nerf: Representing scenes as neural radiance fields for view synthesis.
\newblock In \emph{ECCV}, 2020.

\bibitem[M\"uller et~al.(2022)M\"uller, Evans, Schied, and Keller]{mueller2022instant}
Thomas M\"uller, Alex Evans, Christoph Schied, and Alexander Keller.
\newblock Instant neural graphics primitives with a multiresolution hash encoding.
\newblock \emph{ACM Trans. Graph.}, 41\penalty0 (4):\penalty0 102:1--102:15, 2022.

\bibitem[Noguchi et~al.(2021)Noguchi, Sun, Lin, and Harada]{noguchi2021neural}
Atsuhiro Noguchi, Xiao Sun, Stephen Lin, and Tatsuya Harada.
\newblock Neural articulated radiance field.
\newblock In \emph{ICCV 2021}, pages 5762--5772, 2021.

\bibitem[Park et~al.(2023)Park, Kwon, and Ye]{park2023ed}
Jangho Park, Gihyun Kwon, and Jong~Chul Ye.
\newblock Ed-nerf: Efficient text-guided editing of 3d scene using latent space nerf.
\newblock \emph{arXiv preprint arXiv:2310.02712}, 2023.

\bibitem[Park et~al.(2021)Park, Sinha, Hedman, Barron, Bouaziz, Goldman, Martin-Brualla, and Seitz]{park2021hypernerf}
Keunhong Park, Utkarsh Sinha, Peter Hedman, Jonathan~T Barron, Sofien Bouaziz, Dan~B Goldman, Ricardo Martin-Brualla, and Steven~M Seitz.
\newblock Hypernerf: A higher-dimensional representation for topologically varying neural radiance fields.
\newblock \emph{arXiv preprint arXiv:2106.13228}, 2021.

\bibitem[Peng et~al.(2021)Peng, Zhang, Xu, and et~al.]{peng2021neural}
Sida Peng, Yuanqing Zhang, Yinghao Xu, and et al.
\newblock Neural body: Implicit neural representations with structured latent codes for novel view synthesis of dynamic humans.
\newblock In \emph{CVPR 2021}, pages 9054--9063, 2021.

\bibitem[Podell et~al.(2023)Podell, English, Lacey, Blattmann, Dockhorn, M{\"u}ller, Penna, and Rombach]{podell2023sdxl}
Dustin Podell, Zion English, Kyle Lacey, Andreas Blattmann, Tim Dockhorn, Jonas M{\"u}ller, Joe Penna, and Robin Rombach.
\newblock Sdxl: Improving latent diffusion models for high-resolution image synthesis.
\newblock \emph{arXiv preprint arXiv:2307.01952}, 2023.

\bibitem[Poole et~al.(2022)Poole, Jain, Barron, and Mildenhall]{poole2022dreamfusion}
Ben Poole, Ajay Jain, Jonathan~T Barron, and Ben Mildenhall.
\newblock Dreamfusion: Text-to-3d using 2d diffusion.
\newblock \emph{arXiv preprint arXiv:2209.14988}, 2022.

\bibitem[Poole et~al.(2023)Poole, Jain, Barron, and Mildenhall]{dreamfusion}
Ben Poole, Ajay Jain, Jonathan~T. Barron, and Ben Mildenhall.
\newblock Dreamfusion: Text-to-3d using 2d diffusion.
\newblock In \emph{The Eleventh International Conference on Learning Representations, {ICLR} 2023, Kigali, Rwanda, May 1-5, 2023}. OpenReview.net, 2023.

\bibitem[Raj et~al.(2023)Raj, Kaza, Poole, Niemeyer, Ruiz, Mildenhall, Zada, Aberman, Rubinstein, Barron, et~al.]{raj2023dreambooth3d}
Amit Raj, Srinivas Kaza, Ben Poole, Michael Niemeyer, Nataniel Ruiz, Ben Mildenhall, Shiran Zada, Kfir Aberman, Michael Rubinstein, Jonathan Barron, et~al.
\newblock Dreambooth3d: Subject-driven text-to-3d generation.
\newblock \emph{arXiv preprint arXiv:2303.13508}, 2023.

\bibitem[Ranftl et~al.(2021)Ranftl, Bochkovskiy, and Koltun]{ranftl2021vision}
Ren{\'e} Ranftl, Alexey Bochkovskiy, and Vladlen Koltun.
\newblock Vision transformers for dense prediction.
\newblock In \emph{Proceedings of the IEEE/CVF international conference on computer vision}, pages 12179--12188, 2021.

\bibitem[Rombach et~al.(2022)Rombach, Blattmann, Lorenz, Esser, and Ommer]{rombach2022high}
Robin Rombach, Andreas Blattmann, Dominik Lorenz, Patrick Esser, and Bj{\"o}rn Ommer.
\newblock High-resolution image synthesis with latent diffusion models.
\newblock In \emph{CVPR}, pages 10684--10695, 2022.

\bibitem[{Sara Fridovich-Keil and Alex Yu} et~al.(2022){Sara Fridovich-Keil and Alex Yu}, Tancik, Chen, Recht, and Kanazawa]{yu_and_fridovichkeil2021plenoxels}
{Sara Fridovich-Keil and Alex Yu}, Matthew Tancik, Qinhong Chen, Benjamin Recht, and Angjoo Kanazawa.
\newblock Plenoxels: Radiance fields without neural networks.
\newblock In \emph{CVPR}, 2022.

\bibitem[Schonberger and Frahm(2016)]{schonberger2016structure}
Johannes~L Schonberger and Jan-Michael Frahm.
\newblock Structure-from-motion revisited.
\newblock In \emph{Proceedings of the IEEE conference on computer vision and pattern recognition}, pages 4104--4113, 2016.

\bibitem[Sella et~al.(2023)Sella, Fiebelman, Hedman, and Averbuch-Elor]{sella2023vox}
Etai Sella, Gal Fiebelman, Peter Hedman, and Hadar Averbuch-Elor.
\newblock Vox-e: Text-guided voxel editing of 3d objects.
\newblock In \emph{Proceedings of the IEEE/CVF International Conference on Computer Vision}, pages 430--440, 2023.

\bibitem[Shao et~al.(2023)Shao, Sun, Peng, Zheng, Zhou, Zhang, and Liu]{shao2023control4d}
Ruizhi Shao, Jingxiang Sun, Cheng Peng, Zerong Zheng, Boyao Zhou, Hongwen Zhang, and Yebin Liu.
\newblock Control4d: Dynamic portrait editing by learning 4d gan from 2d diffusion-based editor.
\newblock \emph{arXiv preprint arXiv:2305.20082}, 2023.

\bibitem[Tang et~al.(2023)Tang, Ren, Zhou, Liu, and Zeng]{tang2023dreamgaussian}
Jiaxiang Tang, Jiawei Ren, Hang Zhou, Ziwei Liu, and Gang Zeng.
\newblock Dreamgaussian: Generative gaussian splatting for efficient 3d content creation.
\newblock \emph{arXiv preprint arXiv:2309.16653}, 2023.

\bibitem[Wang et~al.(2022)Wang, Chai, He, Chen, and Liao]{wang2022clip}
Can Wang, Menglei Chai, Mingming He, Dongdong Chen, and Jing Liao.
\newblock Clip-nerf: Text-and-image driven manipulation of neural radiance fields.
\newblock In \emph{Proceedings of the IEEE/CVF Conference on Computer Vision and Pattern Recognition}, pages 3835--3844, 2022.

\bibitem[Wang et~al.(2023)Wang, Jiang, Chai, He, Chen, and Liao]{wang2023nerf}
Can Wang, Ruixiang Jiang, Menglei Chai, Mingming He, Dongdong Chen, and Jing Liao.
\newblock Nerf-art: Text-driven neural radiance fields stylization.
\newblock \emph{IEEE Transactions on Visualization and Computer Graphics}, 2023.

\bibitem[Wu et~al.(2023)Wu, Yi, Fang, Xie, Zhang, Wei, Liu, Tian, and Wang]{wu20234d}
Guanjun Wu, Taoran Yi, Jiemin Fang, Lingxi Xie, Xiaopeng Zhang, Wei Wei, Wenyu Liu, Qi Tian, and Xinggang Wang.
\newblock 4d gaussian splatting for real-time dynamic scene rendering.
\newblock \emph{arXiv preprint arXiv:2310.08528}, 2023.

\bibitem[Xu and Harada(2022)]{xu2022deforming}
Tianhan Xu and Tatsuya Harada.
\newblock Deforming radiance fields with cages.
\newblock In \emph{Computer Vision--ECCV 2022: 17th European Conference, Tel Aviv, Israel, October 23--27, 2022, Proceedings, Part XXXIII}, pages 159--175. Springer, 2022.

\bibitem[Yang et~al.(2022)Yang, Bao, and Zeng]{yang2022neumesh}
Bangbang Yang, Chong Bao, and Junyi et~al. Zeng.
\newblock Neumesh: Learning disentangled neural mesh-based implicit field for geometry and texture editing.
\newblock In \emph{ECCV 2022}, pages 597--614. Springer, 2022.

\bibitem[Yang et~al.(2023{\natexlab{a}})Yang, Gao, Zhou, Jiao, Zhang, and Jin]{yang2023deformable3dgs}
Ziyi Yang, Xinyu Gao, Wen Zhou, Shaohui Jiao, Yuqing Zhang, and Xiaogang Jin.
\newblock Deformable 3d gaussians for high-fidelity monocular dynamic scene reconstruction.
\newblock \emph{arXiv preprint arXiv:2309.13101}, 2023{\natexlab{a}}.

\bibitem[Yang et~al.(2023{\natexlab{b}})Yang, Yang, Pan, Zhu, and Zhang]{yang2023real}
Zeyu Yang, Hongye Yang, Zijie Pan, Xiatian Zhu, and Li Zhang.
\newblock Real-time photorealistic dynamic scene representation and rendering with 4d gaussian splatting.
\newblock \emph{arXiv preprint arXiv:2310.10642}, 2023{\natexlab{b}}.

\bibitem[Yi et~al.(2023)Yi, Fang, Wu, Xie, Zhang, Liu, Tian, and Wang]{yi2023gaussiandreamer}
Taoran Yi, Jiemin Fang, Guanjun Wu, Lingxi Xie, Xiaopeng Zhang, Wenyu Liu, Qi Tian, and Xinggang Wang.
\newblock Gaussiandreamer: Fast generation from text to 3d gaussian splatting with point cloud priors.
\newblock \emph{arXiv preprint arXiv:2310.08529}, 2023.

\bibitem[Yifan et~al.(2019)Yifan, Serena, Wu, {\"O}ztireli, and Sorkine-Hornung]{yifan2019differentiablesplatting}
Wang Yifan, Felice Serena, Shihao Wu, Cengiz {\"O}ztireli, and Olga Sorkine-Hornung.
\newblock Differentiable surface splatting for point-based geometry processing.
\newblock \emph{ACM Transactions on Graphics (TOG)}, 38\penalty0 (6):\penalty0 1--14, 2019.

\bibitem[Yuan et~al.(2022)Yuan, Sun, Lai, and et~al.]{yuan2022nerf}
Yu-Jie Yuan, Yang-Tian Sun, Yu-Kun Lai, and et al.
\newblock Nerf-editing: geometry editing of neural radiance fields.
\newblock In \emph{CVPR 2022}, pages 18353--18364, 2022.

\bibitem[Zhang et~al.(2020)Zhang, Riegler, Snavely, and Koltun]{zhang2020nerf++}
Kai Zhang, Gernot Riegler, Noah Snavely, and Vladlen Koltun.
\newblock Nerf++: Analyzing and improving neural radiance fields.
\newblock \emph{arXiv preprint arXiv:2010.07492}, 2020.

\bibitem[Zhi et~al.(2021)Zhi, Laidlow, Leutenegger, and Davison]{zhi2021place}
Shuaifeng Zhi, Tristan Laidlow, Stefan Leutenegger, and Andrew~J Davison.
\newblock In-place scene labelling and understanding with implicit scene representation.
\newblock In \emph{Proceedings of the IEEE/CVF International Conference on Computer Vision}, pages 15838--15847, 2021.

\bibitem[Zhuang et~al.(2023)Zhuang, Wang, Liu, Lin, and Li]{zhuang2023dreameditor}
Jingyu Zhuang, Chen Wang, Lingjie Liu, Liang Lin, and Guanbin Li.
\newblock Dreameditor: Text-driven 3d scene editing with neural fields.
\newblock \emph{arXiv preprint arXiv:2306.13455}, 2023.

\bibitem[Zwicker et~al.(2001)Zwicker, Pfister, Van~Baar, and Gross]{zwicker2001surfacesplatting}
Matthias Zwicker, Hanspeter Pfister, Jeroen Van~Baar, and Markus Gross.
\newblock Surface splatting.
\newblock In \emph{Proceedings of the 28th annual conference on Computer graphics and interactive techniques}, pages 371--378, 2001.

\end{thebibliography}
}

\clearpage
\appendix
\begin{center}
\textbf{\large Appendix}
\end{center}
\section{Introduction}
The content of our supplementary material is organized as follows:
\begin{itemize}
  \item Firstly, we provide more qualitative results in Section~\ref{sec:more}.
  \item Secondly, we demonstrate WebUI for \name in Section.~\ref{sec:web} along with specifically tailored algorithem for WebUI 3D editing.
  \item We attach the video of using our WebUI, including tracing, editing, deleting, and adding objects.
\end{itemize}

\section{More Results}
\label{sec:more}
\begin{figure*}
    \centering
    \vspace{-8mm}
    \begin{minipage}{0.243\textwidth}
        \centering
        \begin{subfigure}{\linewidth}
            \centering
            \includegraphics[width=1.0\linewidth]{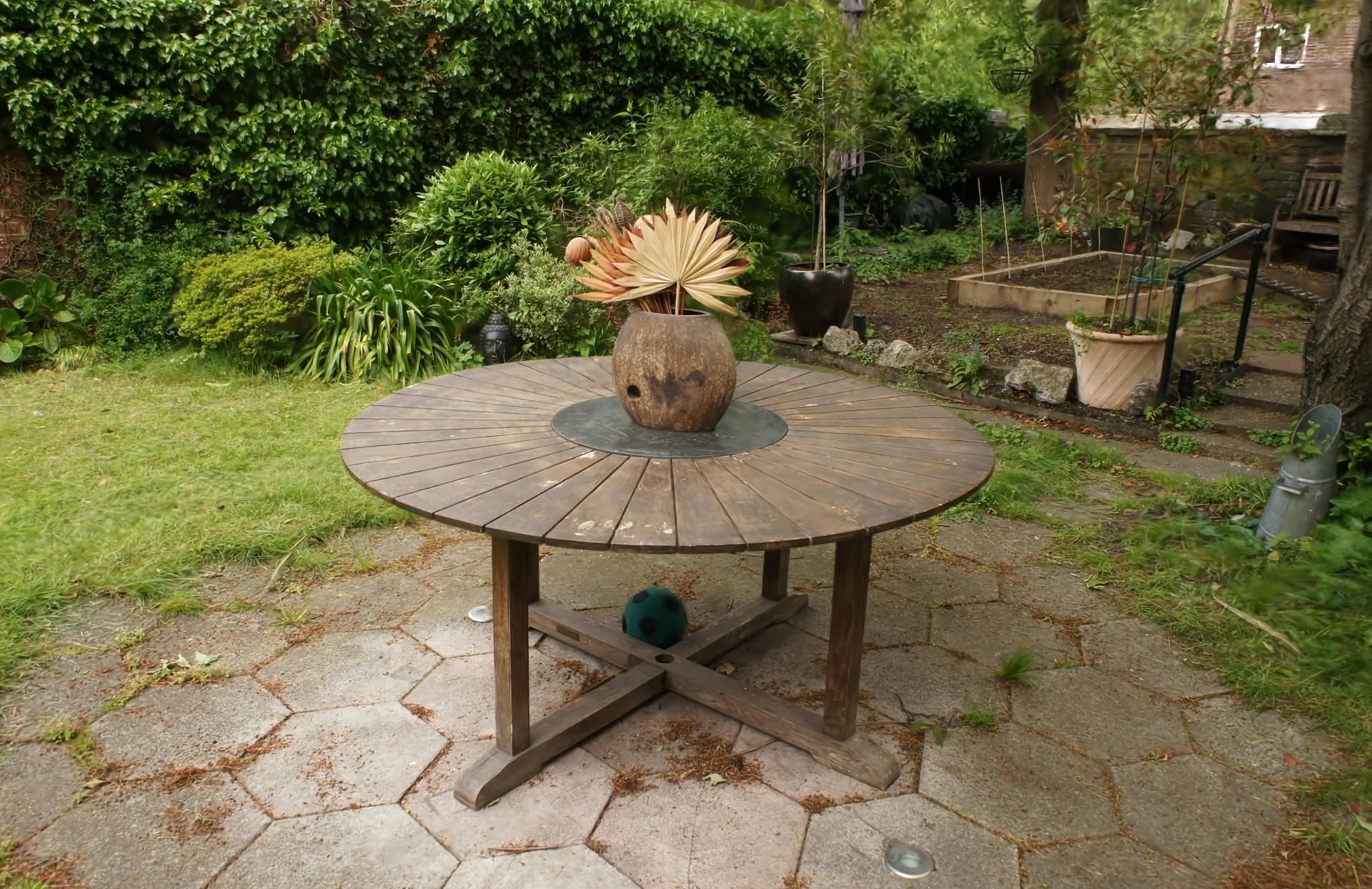}
        \end{subfigure}
    \end{minipage}
    \hfill
    \begin{minipage}{0.243\textwidth}
        \centering
        \begin{subfigure}{\linewidth}
            \centering
            \includegraphics[width=1.0\linewidth]{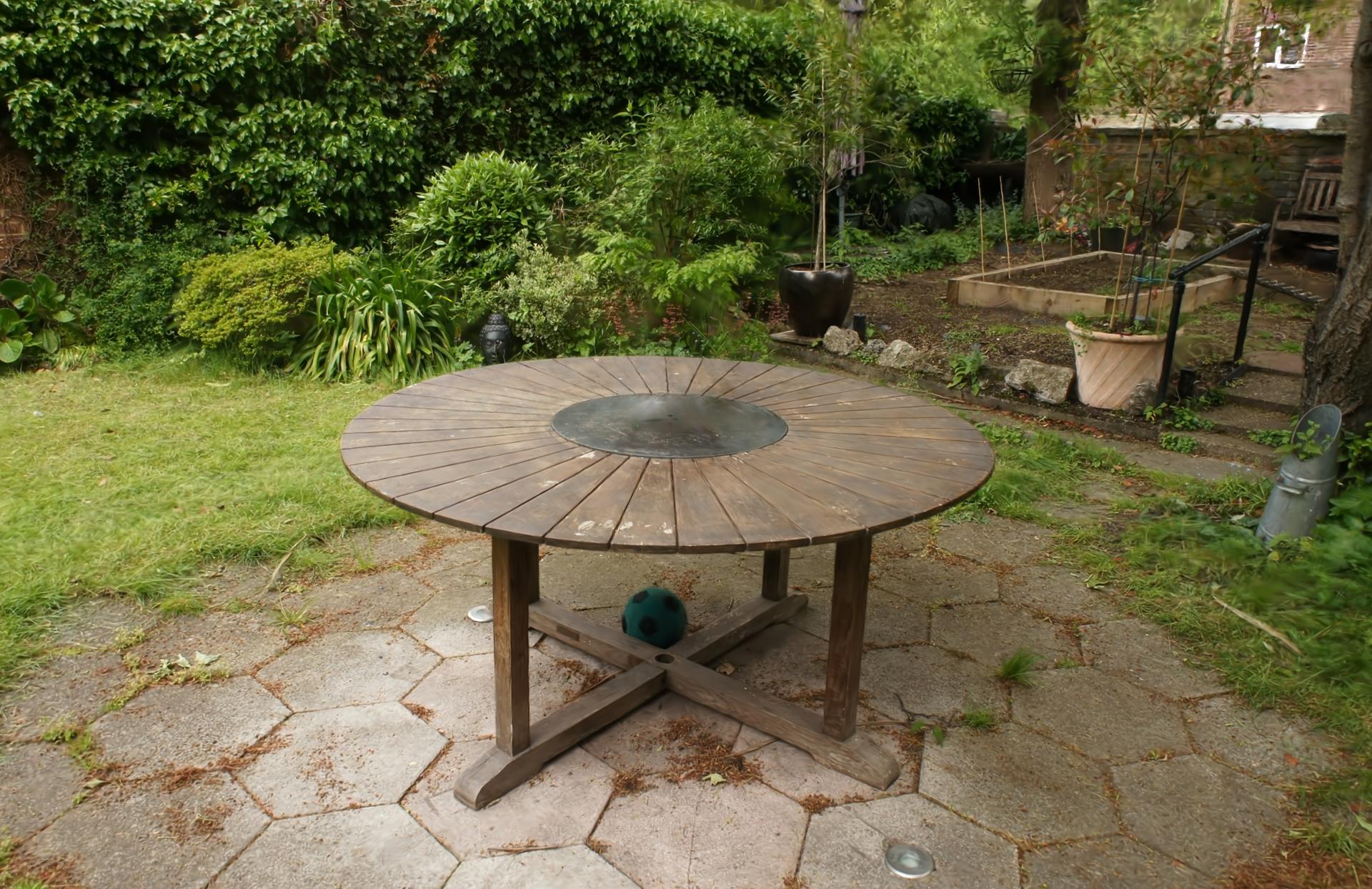}
        \end{subfigure}
    \end{minipage}
    \hfill
    \begin{minipage}{0.243\textwidth}
        \centering
        \begin{subfigure}{\linewidth}
            \centering
            \includegraphics[width=1.0\linewidth]{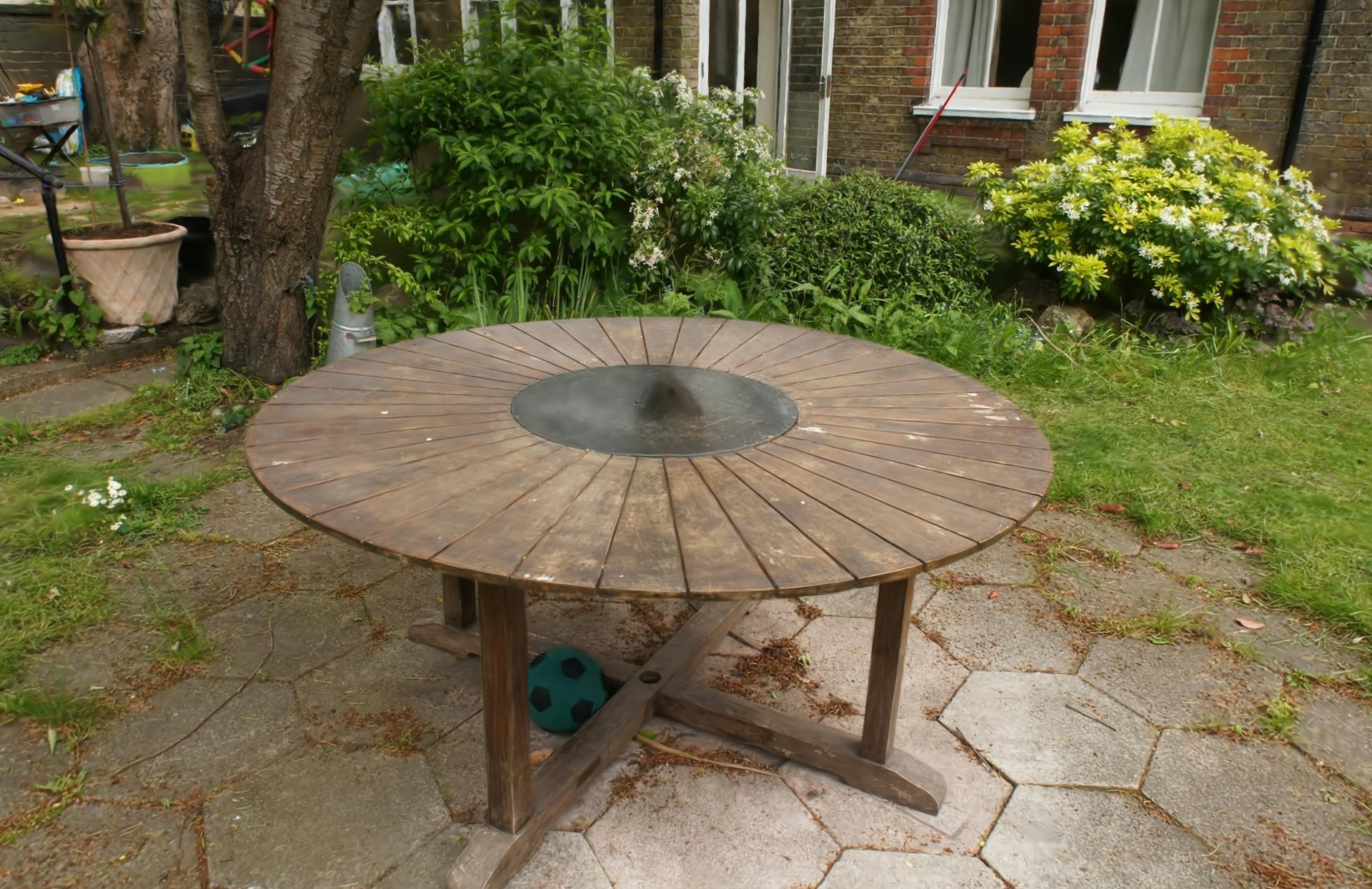}
        \end{subfigure}
    \end{minipage}
    \hfill
    \begin{minipage}{0.243\textwidth}
        \centering
        \begin{subfigure}{\linewidth}
            \centering
            \includegraphics[width=1.0\linewidth]{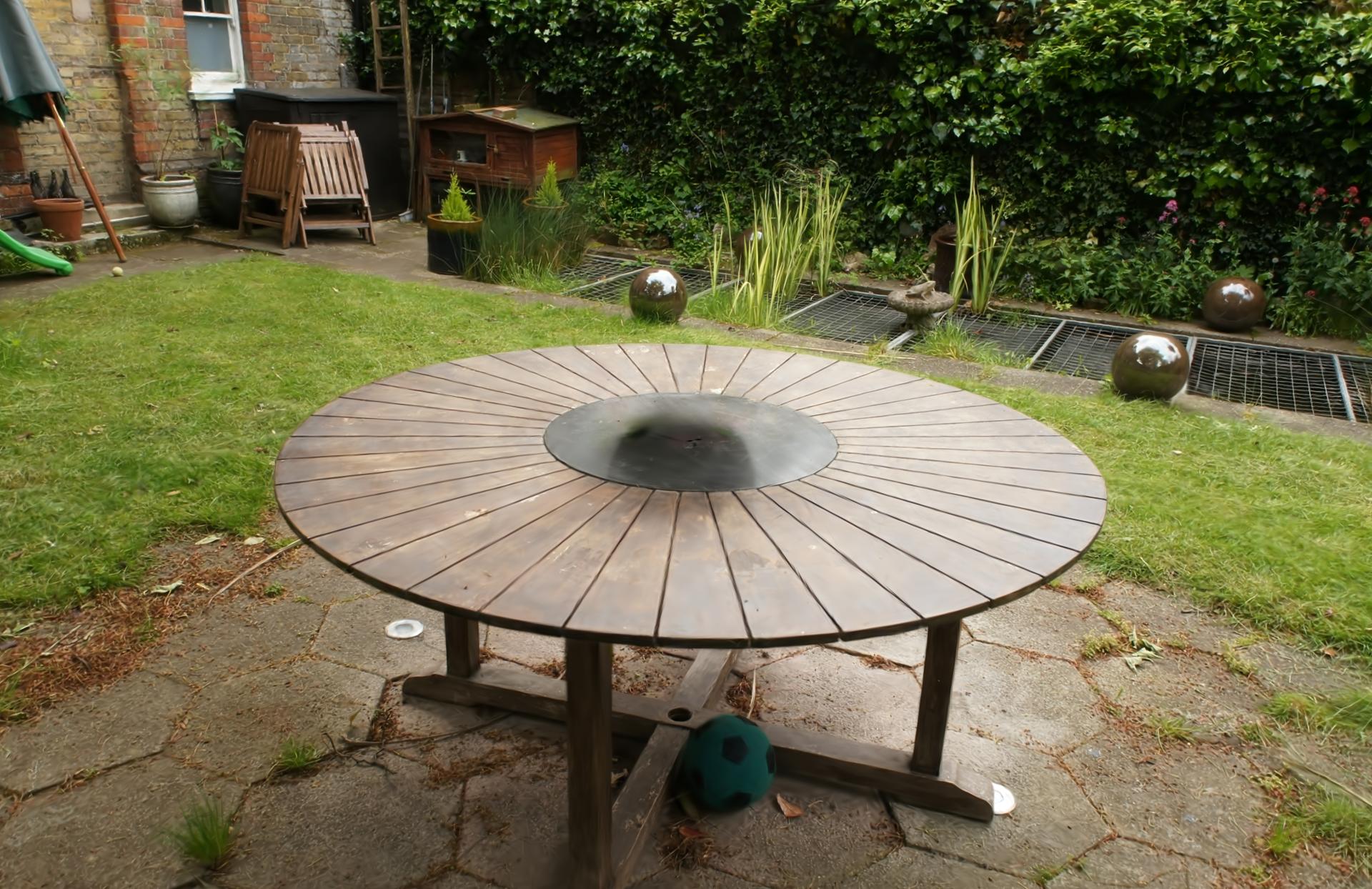}
        \end{subfigure}
    \end{minipage}
    \\ 
    \vspace{1mm}
    \textit{``Remove the vase"}
    \vspace{1mm}

    \begin{minipage}{0.243\textwidth}
        \centering
        \begin{subfigure}{\linewidth}
            \centering
            \includegraphics[width=1.0\linewidth]{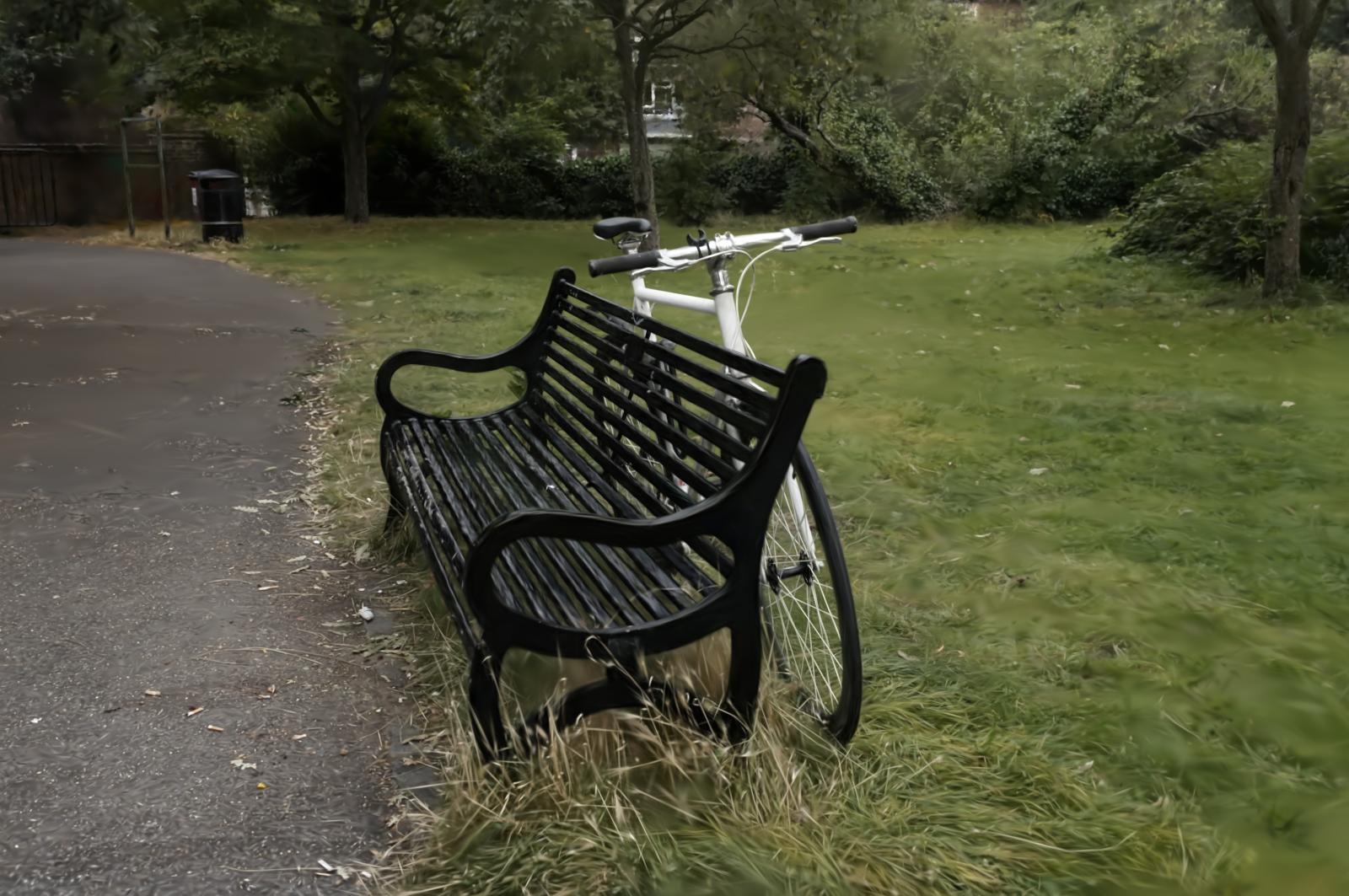}
        \end{subfigure}
    \end{minipage}
    \hfill
    \begin{minipage}{0.243\textwidth}
        \centering
        \begin{subfigure}{\linewidth}
            \centering
            \includegraphics[width=1.0\linewidth]{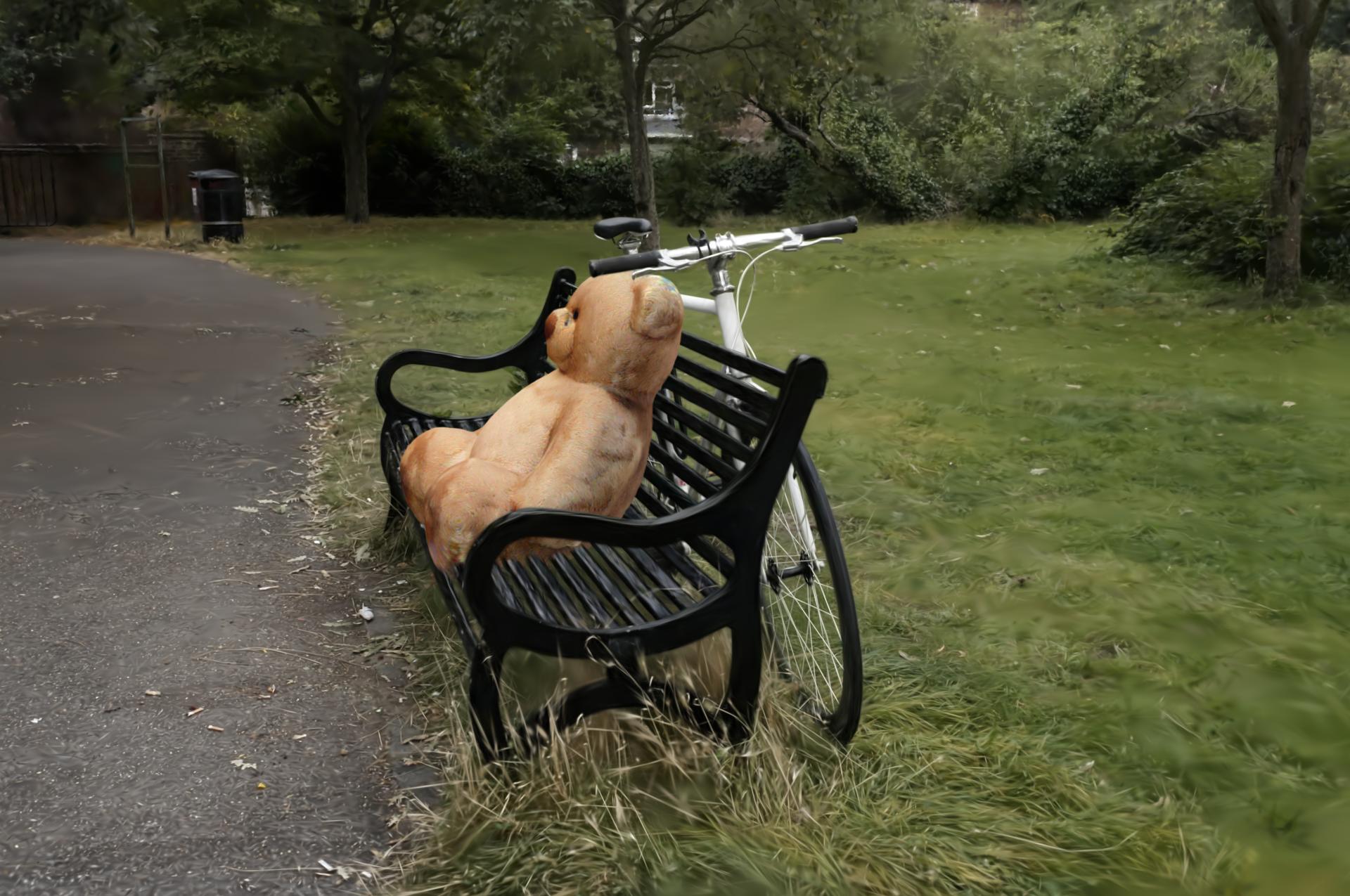}
        \end{subfigure}
    \end{minipage}
    \hfill
    \begin{minipage}{0.243\textwidth}
        \centering
        \begin{subfigure}{\linewidth}
            \centering
            \includegraphics[width=1.0\linewidth]{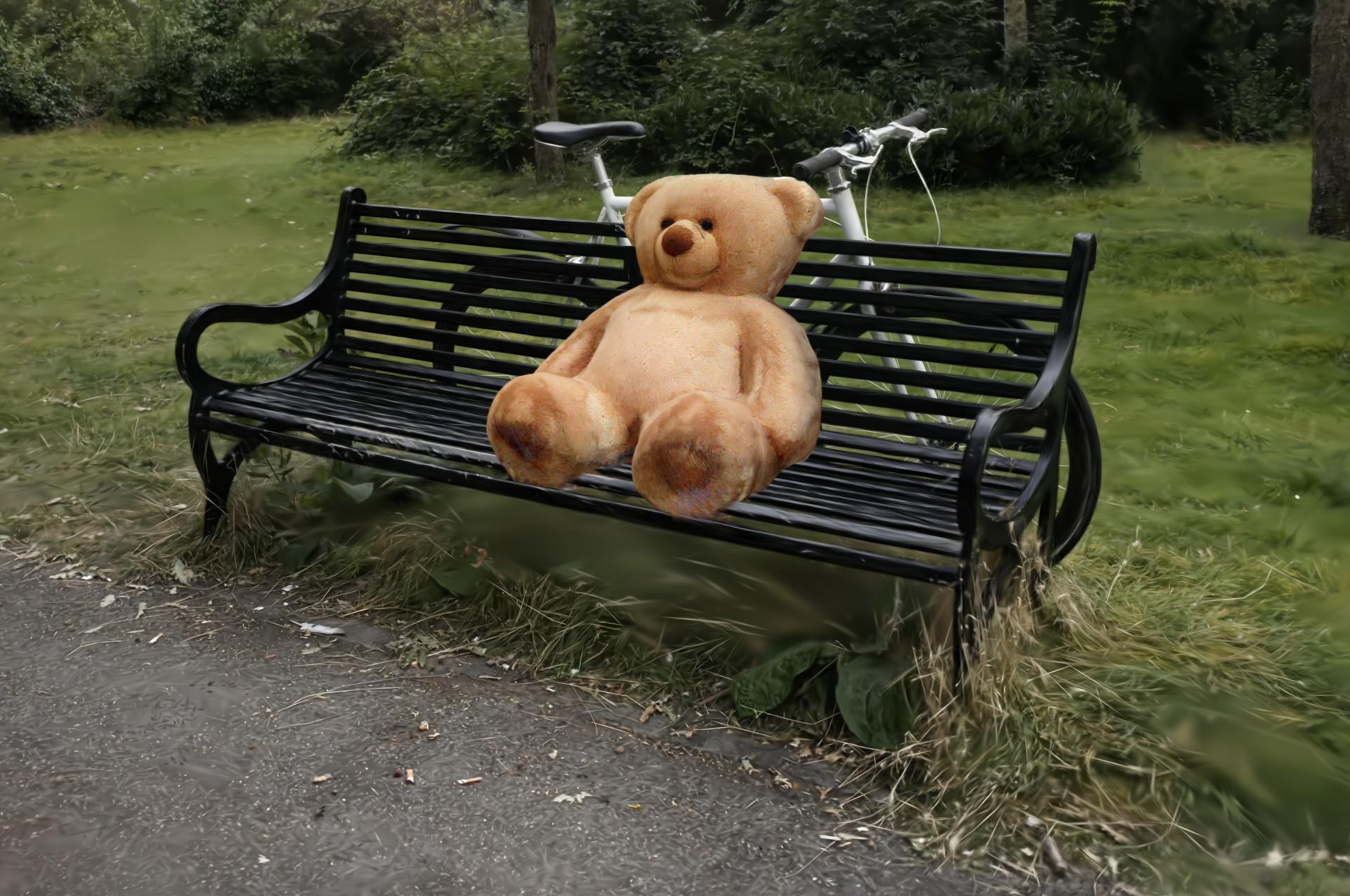}
        \end{subfigure}
    \end{minipage}
    \hfill
    \begin{minipage}{0.243\textwidth}
        \centering
        \begin{subfigure}{\linewidth}
            \centering
            \includegraphics[width=1.0\linewidth]{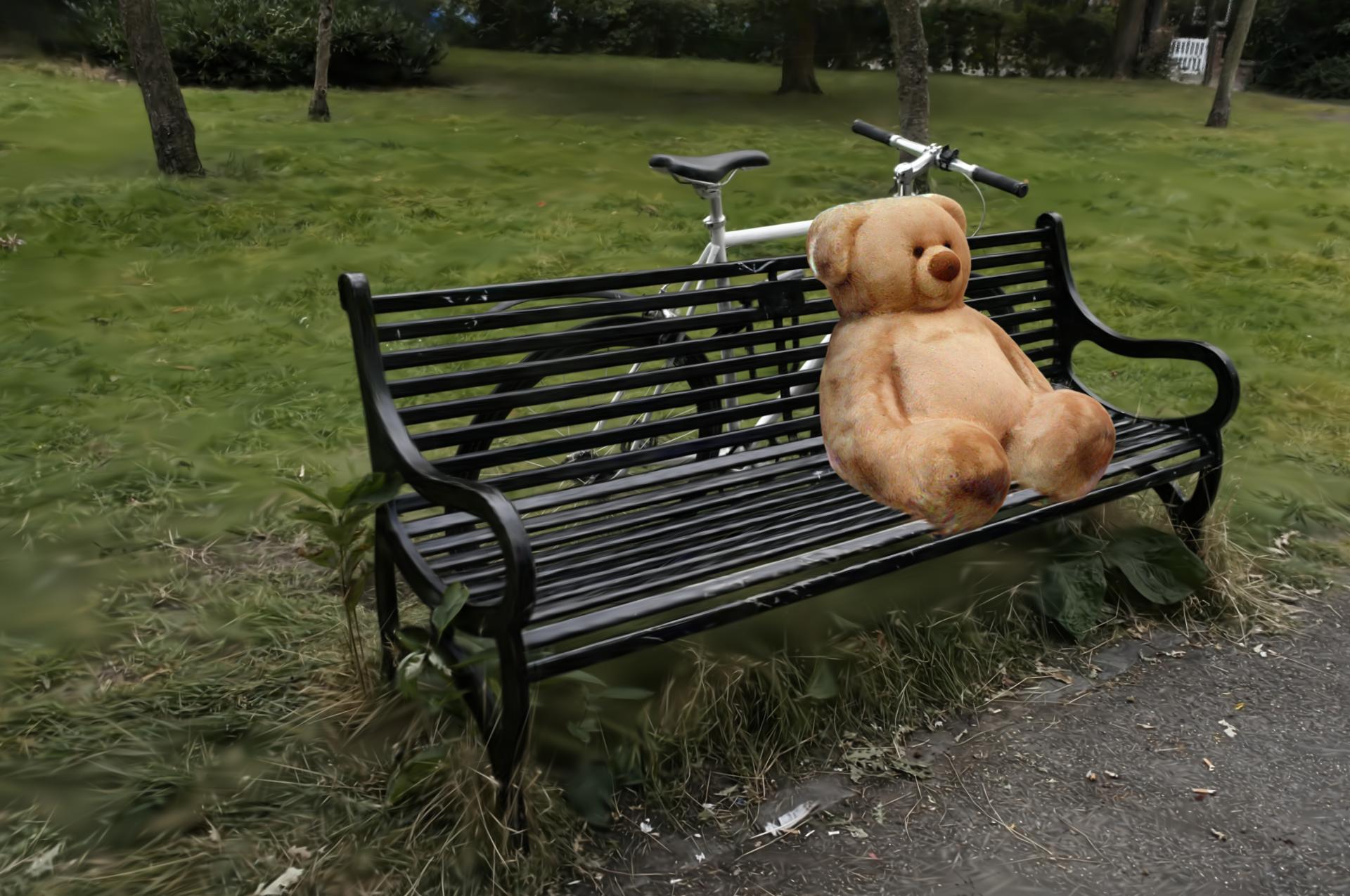}
        \end{subfigure}
    \end{minipage}
    \\ 
    \vspace{1mm}
    \textit{``Add a teddy bear on the bench"}
    \vspace{1mm}

    \begin{minipage}{0.243\textwidth}
        \centering
        \begin{subfigure}{\linewidth}
            \centering
            \includegraphics[width=1.0\linewidth]{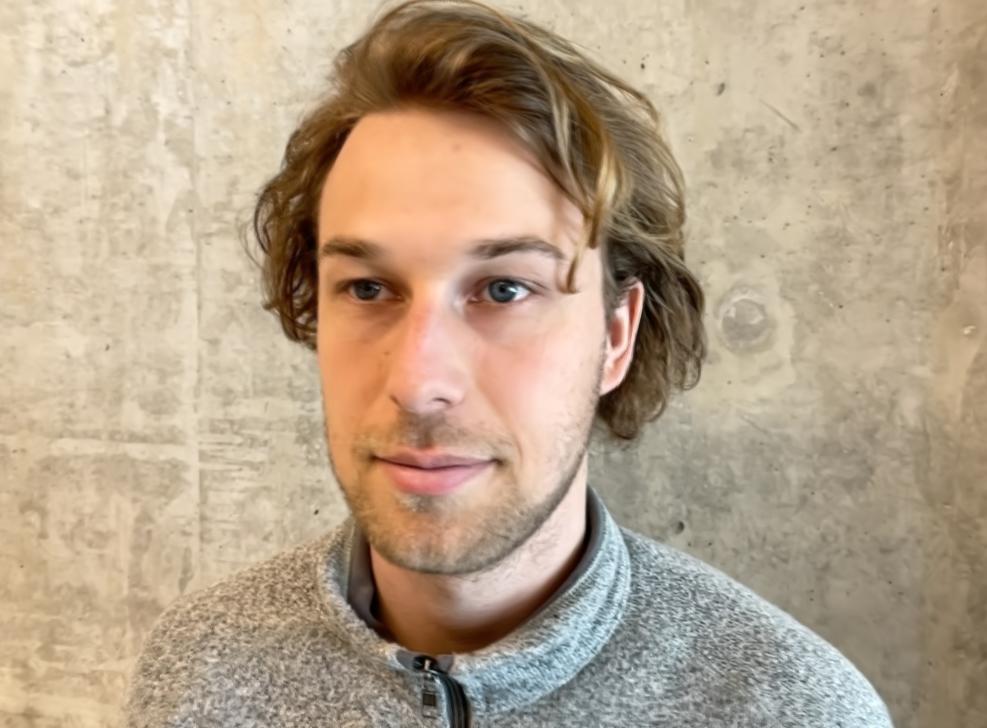}
        \end{subfigure}
    \end{minipage}
    \hfill
    \begin{minipage}{0.243\textwidth}
        \centering
        \begin{subfigure}{\linewidth}
            \centering
            \includegraphics[width=1.0\linewidth]{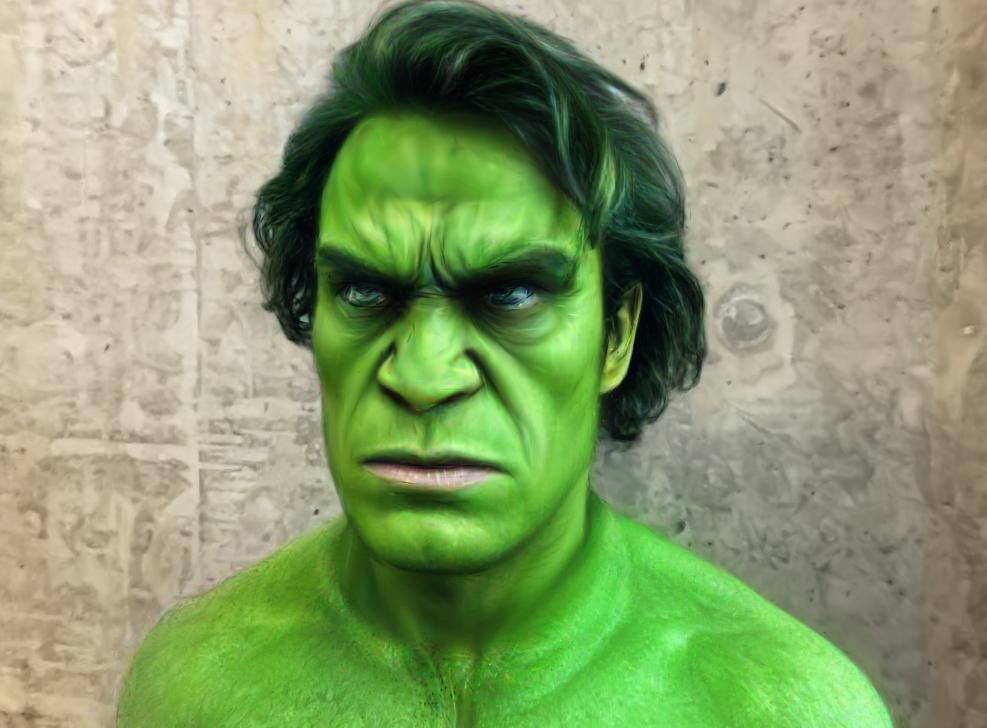}
        \end{subfigure}
    \end{minipage}
    \hfill
    \begin{minipage}{0.243\textwidth}
        \centering
        \begin{subfigure}{\linewidth}
            \centering
            \includegraphics[width=1.0\linewidth]{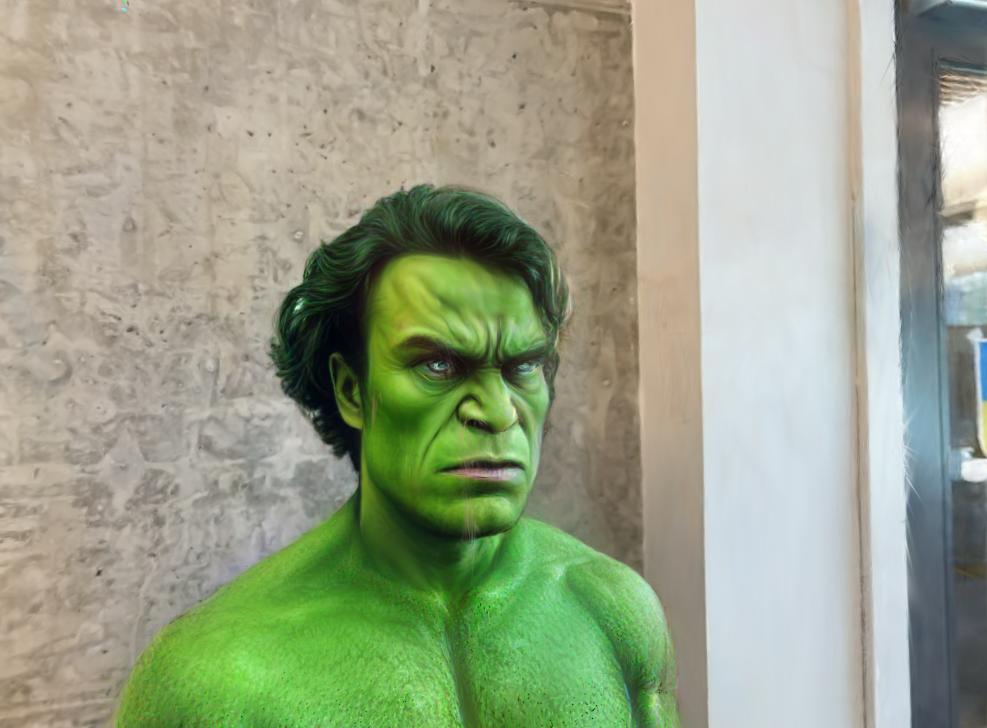}
        \end{subfigure}
    \end{minipage}
    \hfill
    \begin{minipage}{0.243\textwidth}
        \centering
        \begin{subfigure}{\linewidth}
            \centering
            \includegraphics[width=1.0\linewidth]{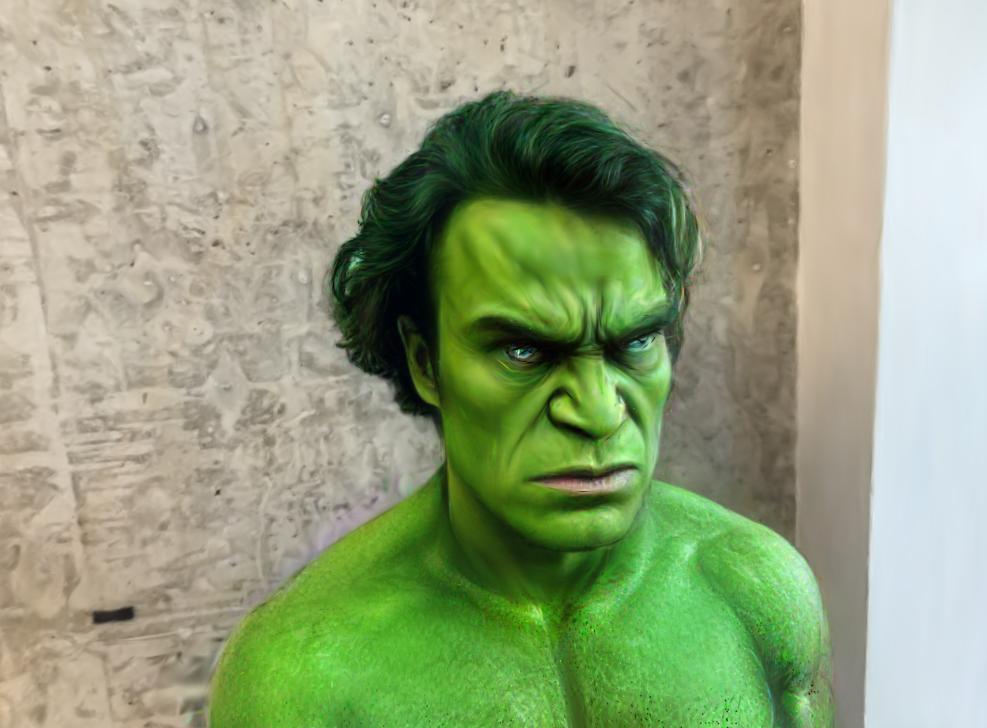}
        \end{subfigure}
    \end{minipage}
    \\ 
    \vspace{1mm}
    \textit{``Turn him into Hulk"}
    \vspace{1mm}

    \begin{minipage}{0.243\textwidth}
        \centering
        \begin{subfigure}{\linewidth}
            \centering
            \includegraphics[width=1.0\linewidth]{imgs/face31.jpg}
        \end{subfigure}
    \end{minipage}
    \hfill
    \begin{minipage}{0.243\textwidth}
        \centering
        \begin{subfigure}{\linewidth}
            \centering
            \includegraphics[width=1.0\linewidth]{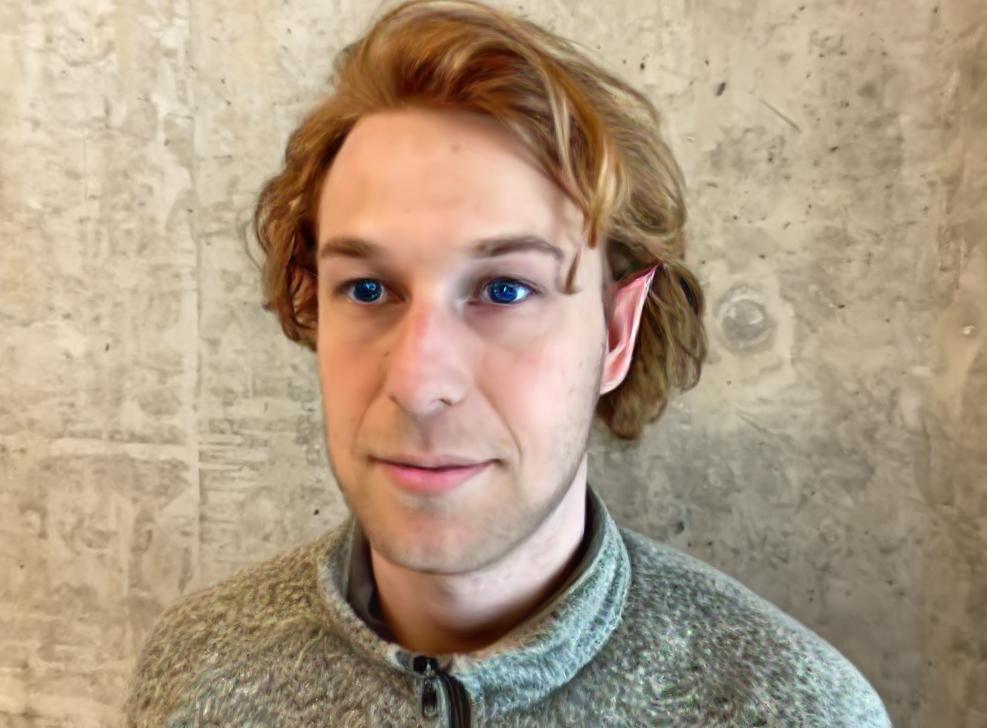}
        \end{subfigure}
    \end{minipage}
    \hfill
    \begin{minipage}{0.243\textwidth}
        \centering
        \begin{subfigure}{\linewidth}
            \centering
            \includegraphics[width=1.0\linewidth]{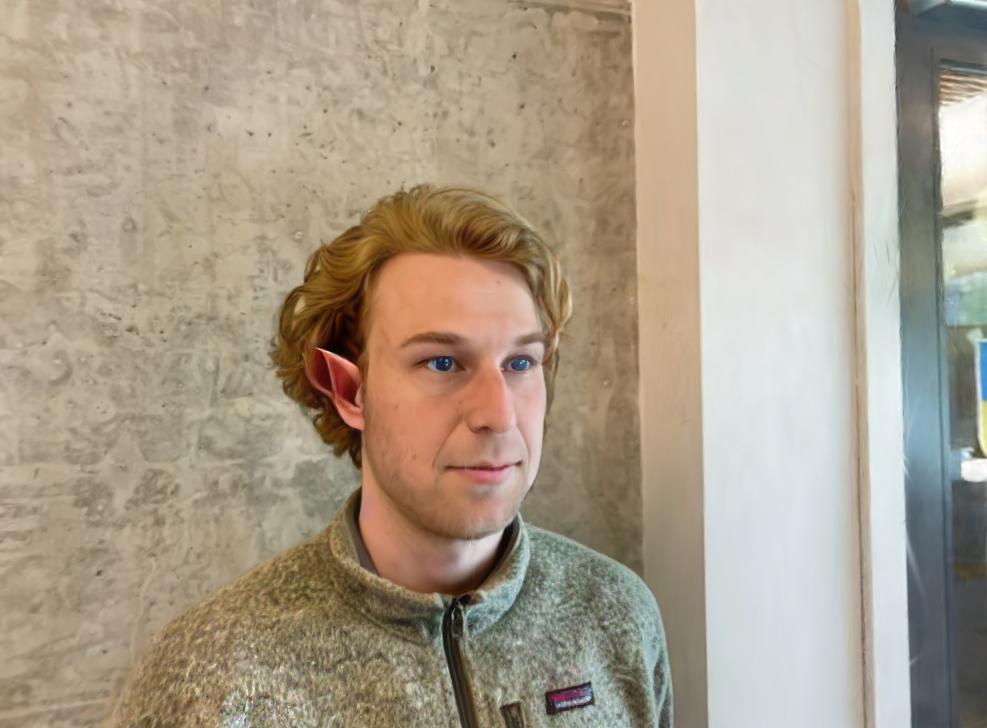}
        \end{subfigure}
    \end{minipage}
    \hfill
    \begin{minipage}{0.243\textwidth}
        \centering
        \begin{subfigure}{\linewidth}
            \centering
            \includegraphics[width=1.0\linewidth]{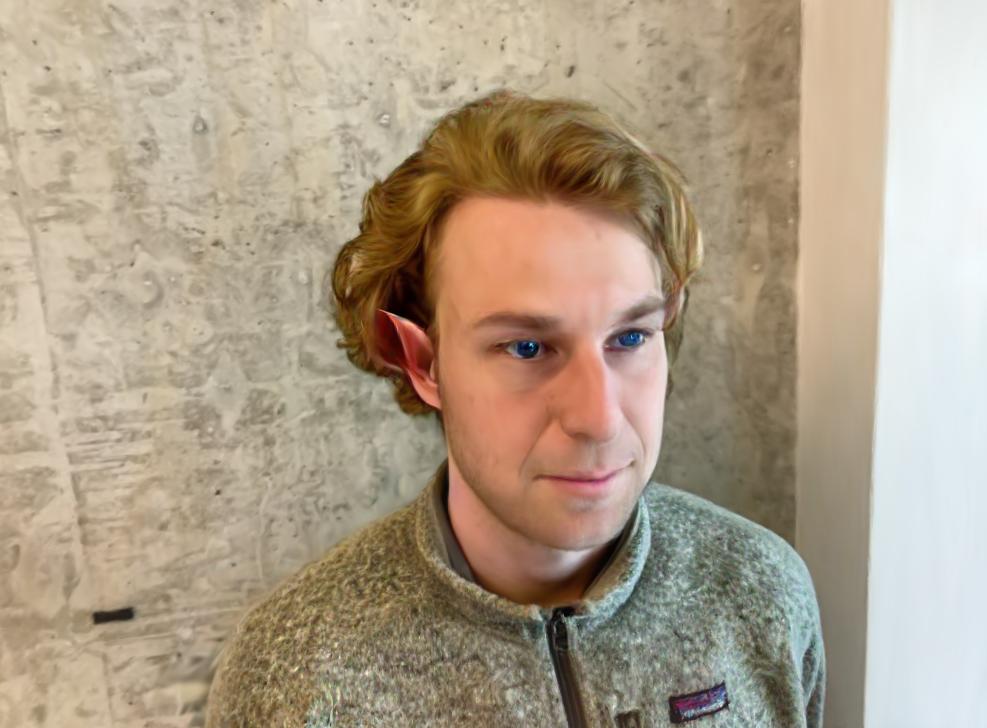}
        \end{subfigure}
    \end{minipage}
    \\ 
    \vspace{1mm}
    \textit{``Turn him into an elf"}
    \vspace{1mm}
    
    \caption{\textbf{More results of GaussianEditor.} GaussianEditor allows for fast, versatile, high-resolution 3D editing, requiring only 2-7 minutes and 10-20GB of GPU memory on a single A6000 GPU. Please note that the background of face editing scenes remains unchanged.}
    \label{fig:more results}
    \vspace{-1mm}
\end{figure*}

In Fig.~\ref{fig:more results}, we demonstrate more results of \name. Our method provides controllable, diverse, high-resolution 3D editing, needing only 2-7 minutes.

\section{WebUI}
\label{sec:web}

Although works in the neural radiance fields (NeRF)~\cite{nerf} domain also incorporate WebUIs, the slow rendering speeds of NeRF mean that users are confined to low resolutions and very low frame rates when using the WebUI, resulting in a subpar user experience. Fortunately, thanks to our adoption of Gaussian Splatting~\cite{3dgs} in \name, a method known for its rapid rendering capabilities, our WebUI can comfortably support usage at 2K resolution and 60fps. Besides, we leverage the interactivity of the webUI to enhance both semantic tracing and object incorporation applications, which will be discussed in the following two subsections.

\subsection{Semantic Tracing with Point-base Prompts}
\label{sec:point}
Interactive WebUI applications with user interfaces are extremely important. In practical scenarios, users often intend to edit only specific areas of a complete scene, a task that can be challenging to specify solely through text prompts. For instance, it becomes difficult to determine through text which object a user wants to edit when multiple objects of the same type are present in a scene, and the user wishes to change just one of them. To address this issue, we propose semantic tracing with point-based prompts.

Semantic tracing with point-based prompts requires users to click on the screen to add 2D points from a specific view. Specifically, when the user clicks a point on the screen, we back-project this point into a spatial point based on the intrinsic and extrinsic parameters of the current viewpoint camera:
\begin{equation}
    [x, y, z]^T=[\mathbf{R}|\mathbf{t}]z(\boldsymbol{p})\mathbf{K}^{-1}[p_x, p_y, 1]^T,
\end{equation}
where $[\mathbf{R}|\mathbf{t}]$ and $\mathbf{K}$ denote the extrinsic and intrinsic of the current camera, $\boldsymbol{p}$, $z(\boldsymbol{p})$ and $[x, y, z]^T$ refer to the user-clicked pixel, its corresponding depth, and the spatial point, respectively.

\begin{figure*}[h]
    \centering
    \vspace{-5mm}
    \begin{minipage}{0.49\textwidth}
        \centering
        \begin{subfigure}{\linewidth}
            \centering
            \includegraphics[width=1.0\linewidth]{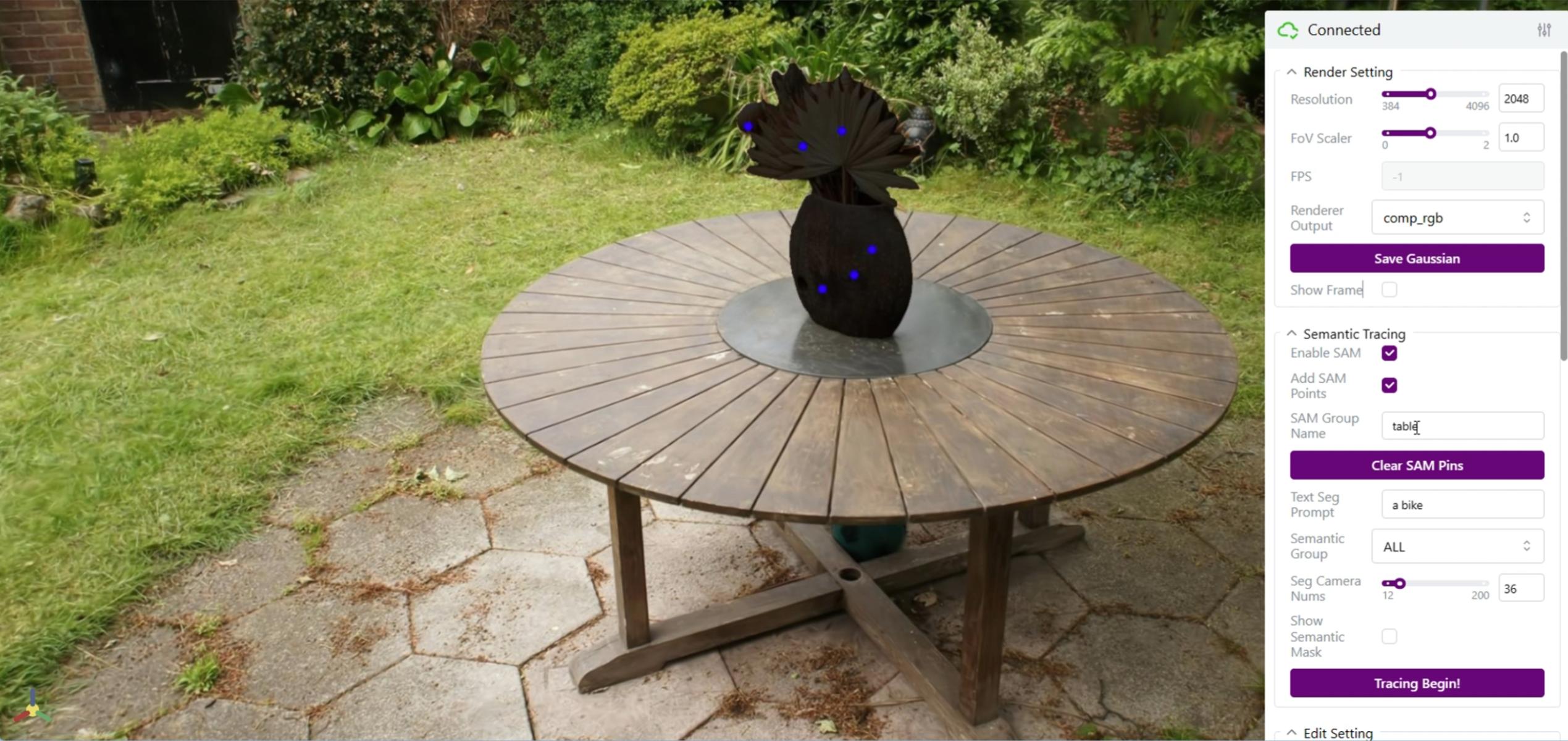}
            \caption*{(a)}
        \end{subfigure}
    \end{minipage}
    \hfill
    \begin{minipage}{0.49\textwidth}
        \centering
        \begin{subfigure}{\linewidth}
            \centering
            \includegraphics[width=1.0\linewidth]{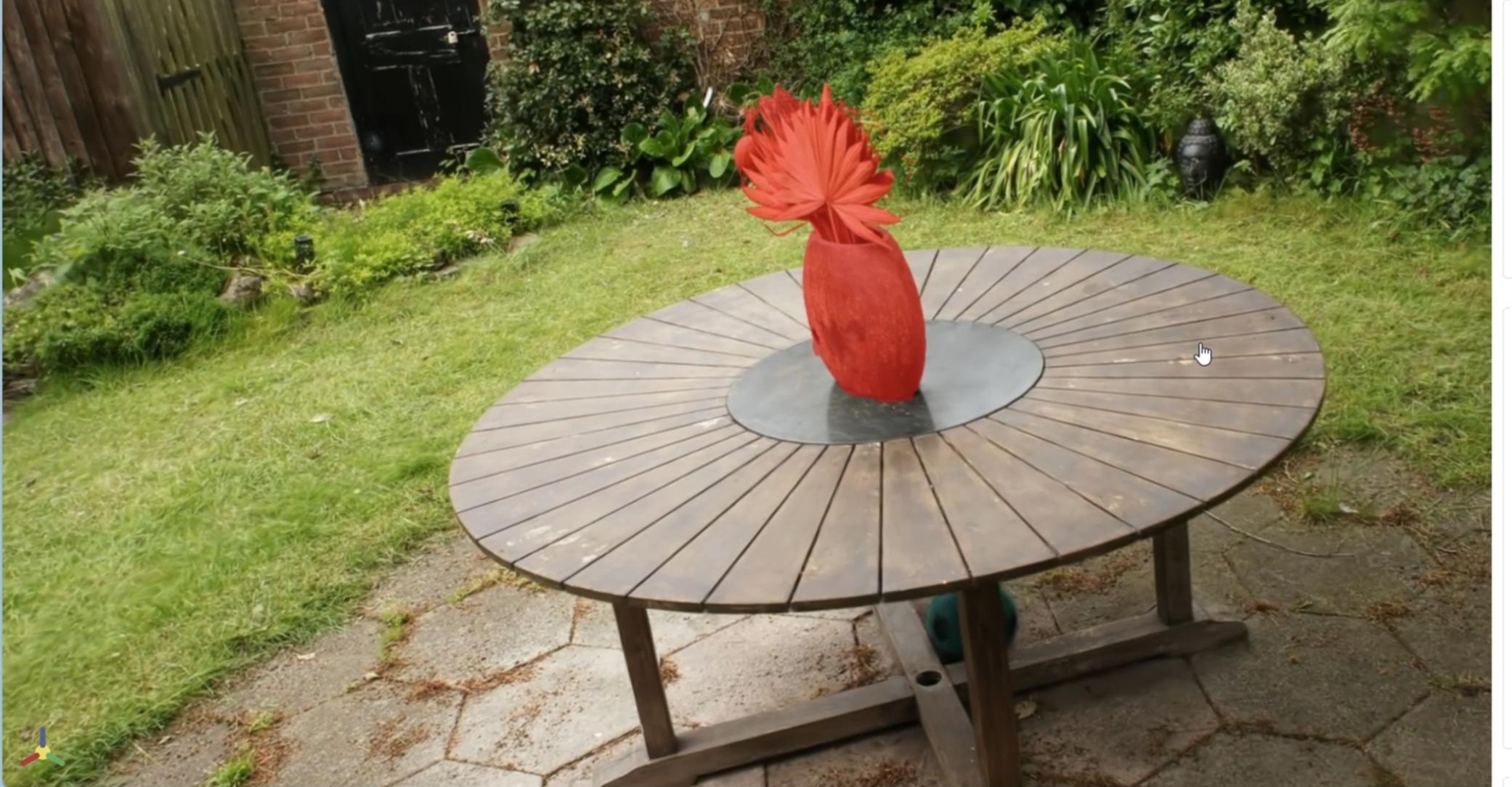}
                        \caption*{(b)}

        \end{subfigure}
    \end{minipage}    
    \\
    \vspace{1mm}
    \vspace{1mm}

    \begin{minipage}{0.49\textwidth}
        \centering
        \begin{subfigure}{\linewidth}
            \centering
            \includegraphics[width=1.0\linewidth]{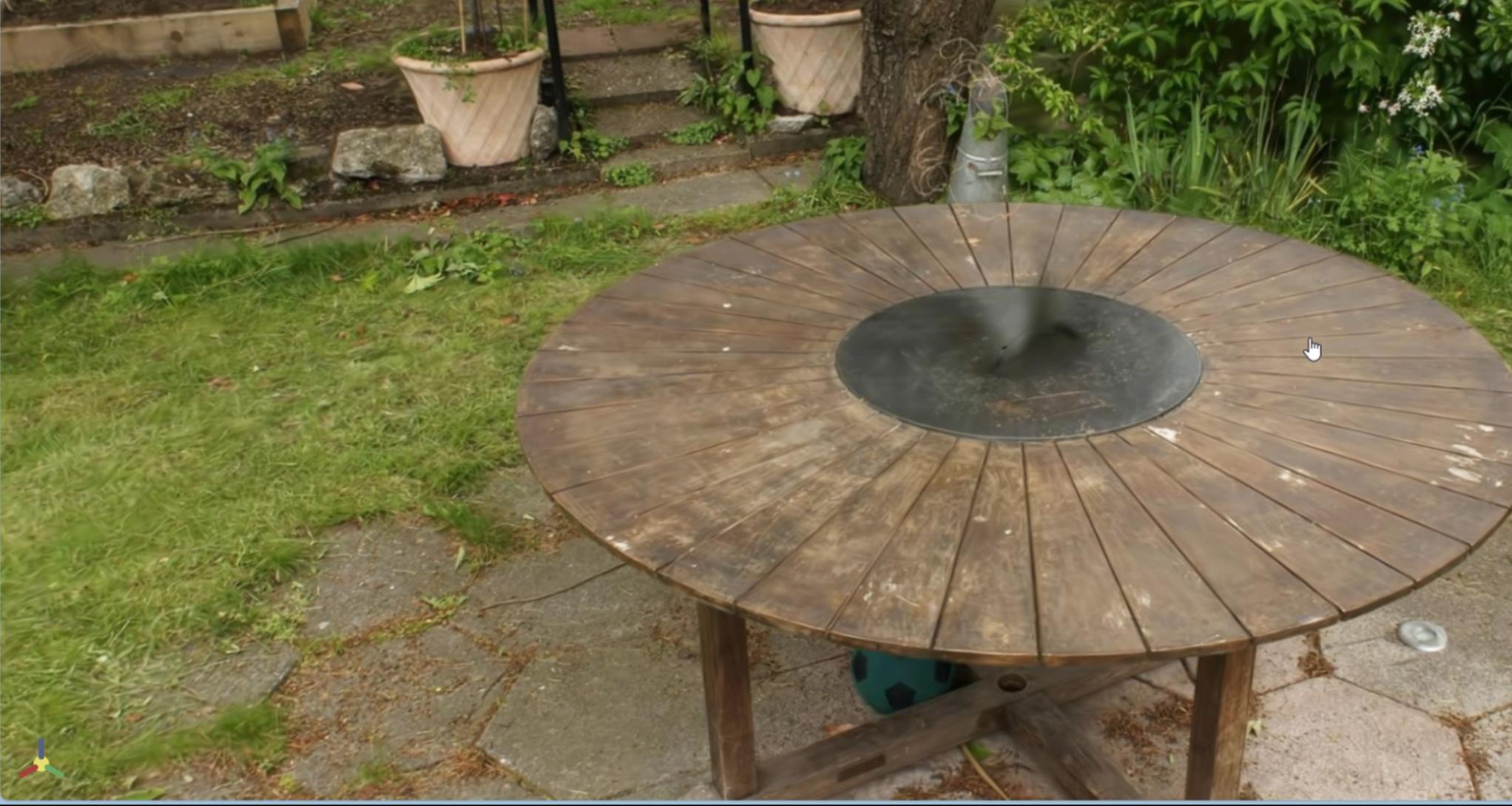}
                        \caption*{(c)}

        \end{subfigure}
    \end{minipage}
    \hfill
    \begin{minipage}{0.49\textwidth}
        \centering
        \begin{subfigure}{\linewidth}
            \centering
            \includegraphics[width=1.0\linewidth]{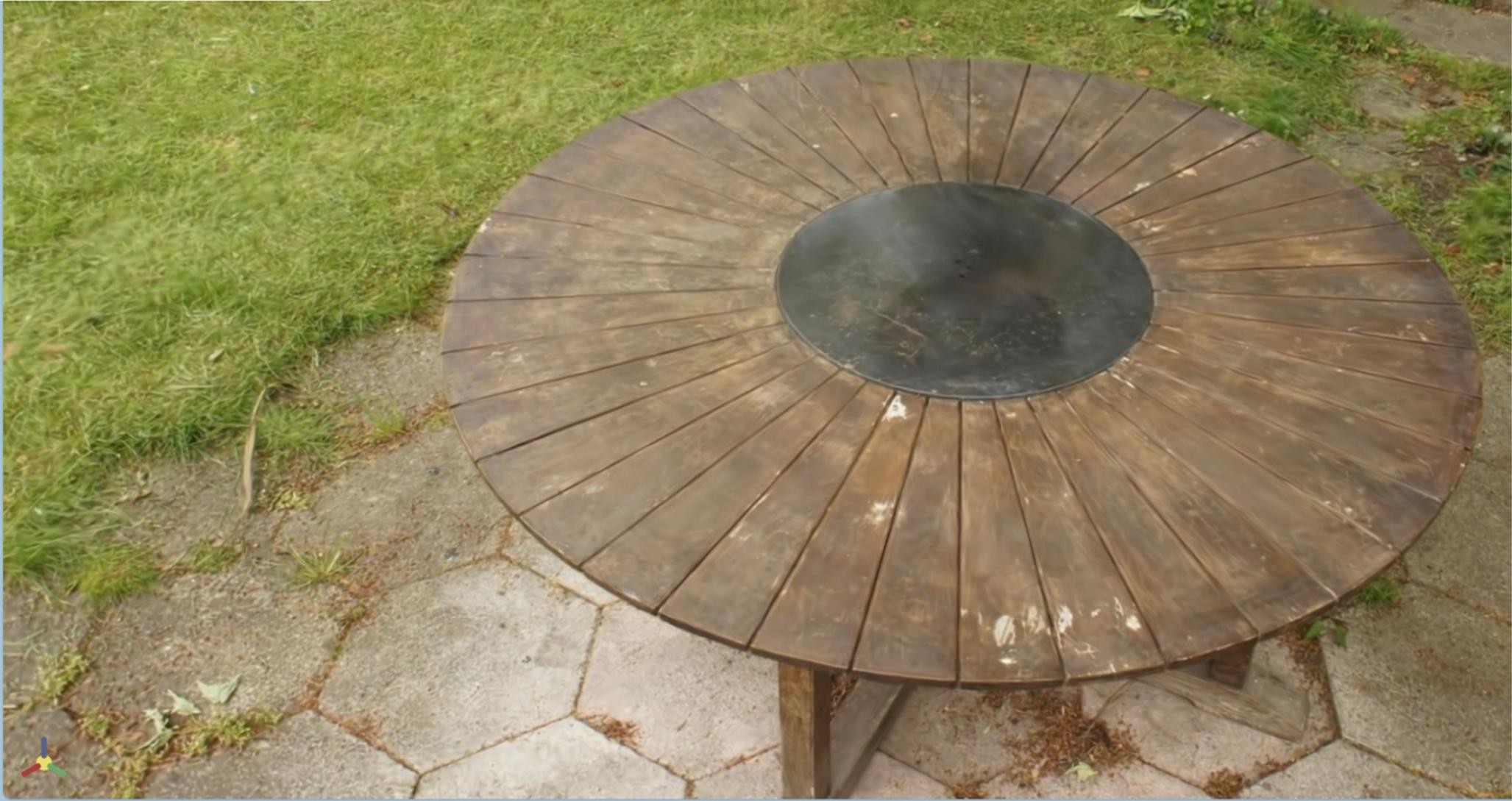}
                        \caption*{(d)}

        \end{subfigure}
    \end{minipage}

    \caption{Semantic Tracing with Point-based Prompts. In (a), users provide key points on a view by clicking the screen with the mouse. In (b), we segment the target object based on these points. (c) and (d) depict the results after removing the segmented objects. It can be seen from the above that our point-based tracing method offers high precision and interactivity.}
    \label{fig:vase}
\end{figure*}
\begin{figure*}[h]
    \centering
    \begin{minipage}{0.49\textwidth}
        \centering
        \begin{subfigure}{\linewidth}
            \centering
            \includegraphics[width=1.0\linewidth]{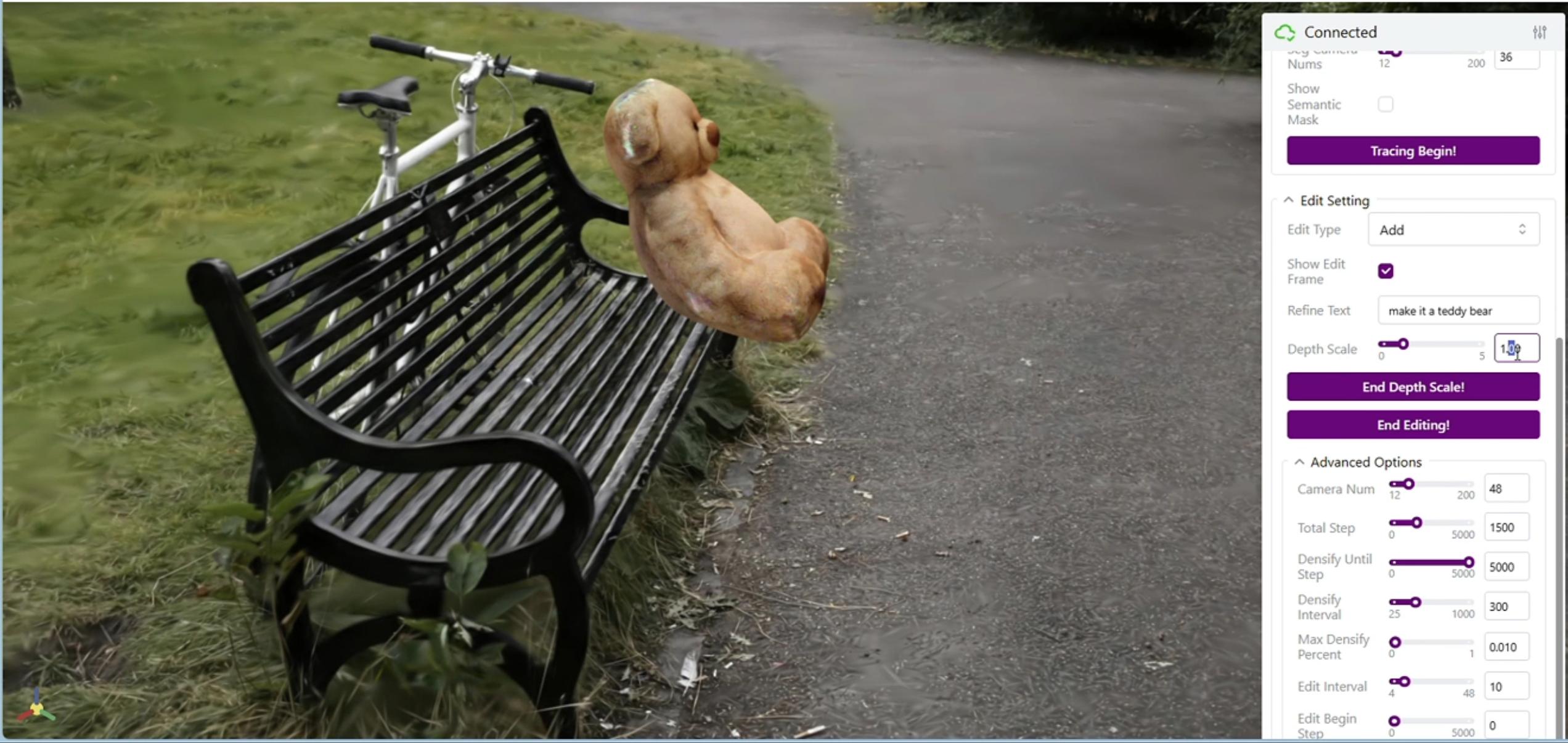}
                                    \caption*{(a)}

        \end{subfigure}
    \end{minipage}
    \hfill
    \begin{minipage}{0.49\textwidth}
        \centering
        \begin{subfigure}{\linewidth}
            \centering
            \includegraphics[width=1.0\linewidth]{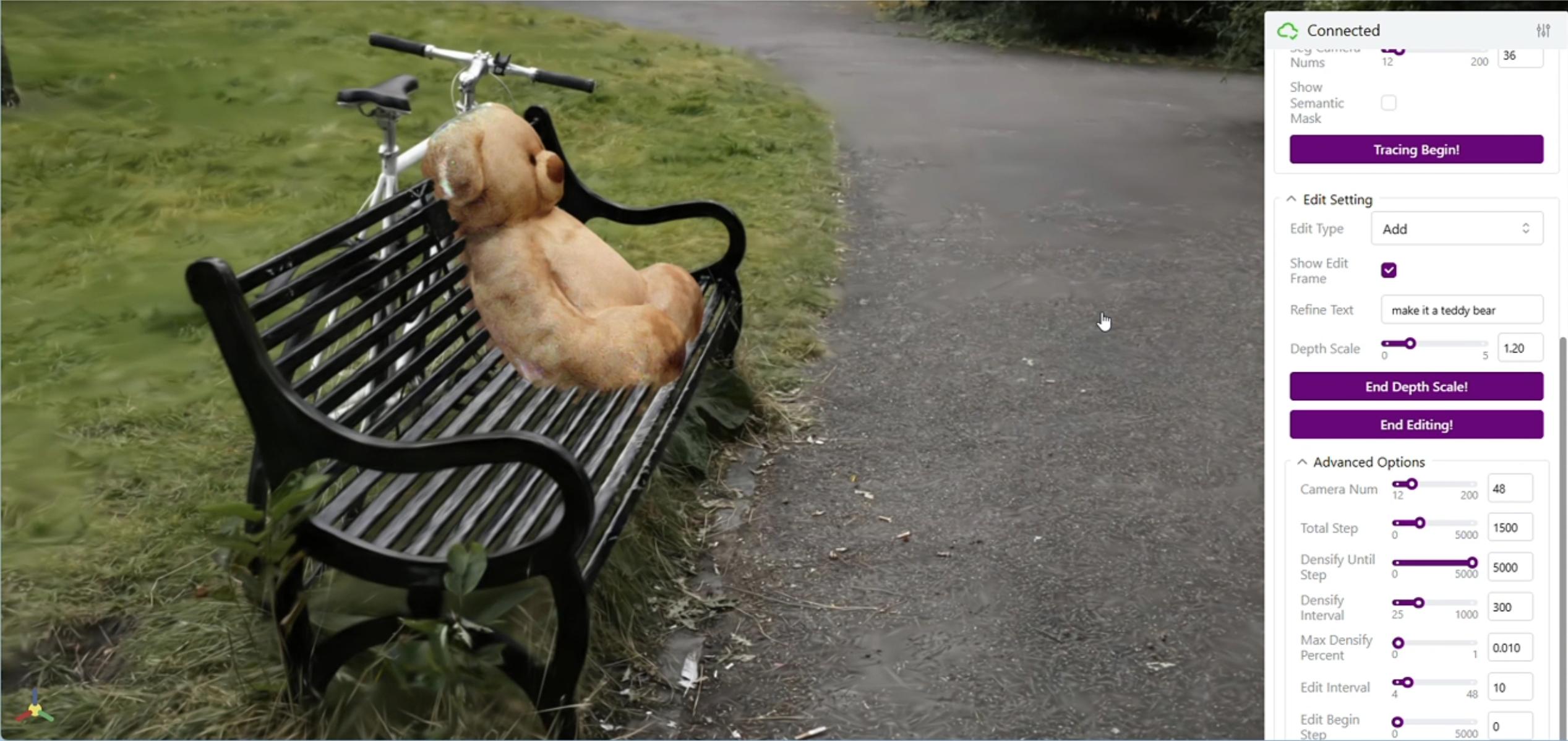}
                                    \caption*{(b)}

        \end{subfigure}
    \end{minipage}    
    \\
    \vspace{1mm}
    \vspace{1mm}

    \begin{minipage}{0.49\textwidth}
        \centering
        \begin{subfigure}{\linewidth}
            \centering
            \includegraphics[width=1.0\linewidth]{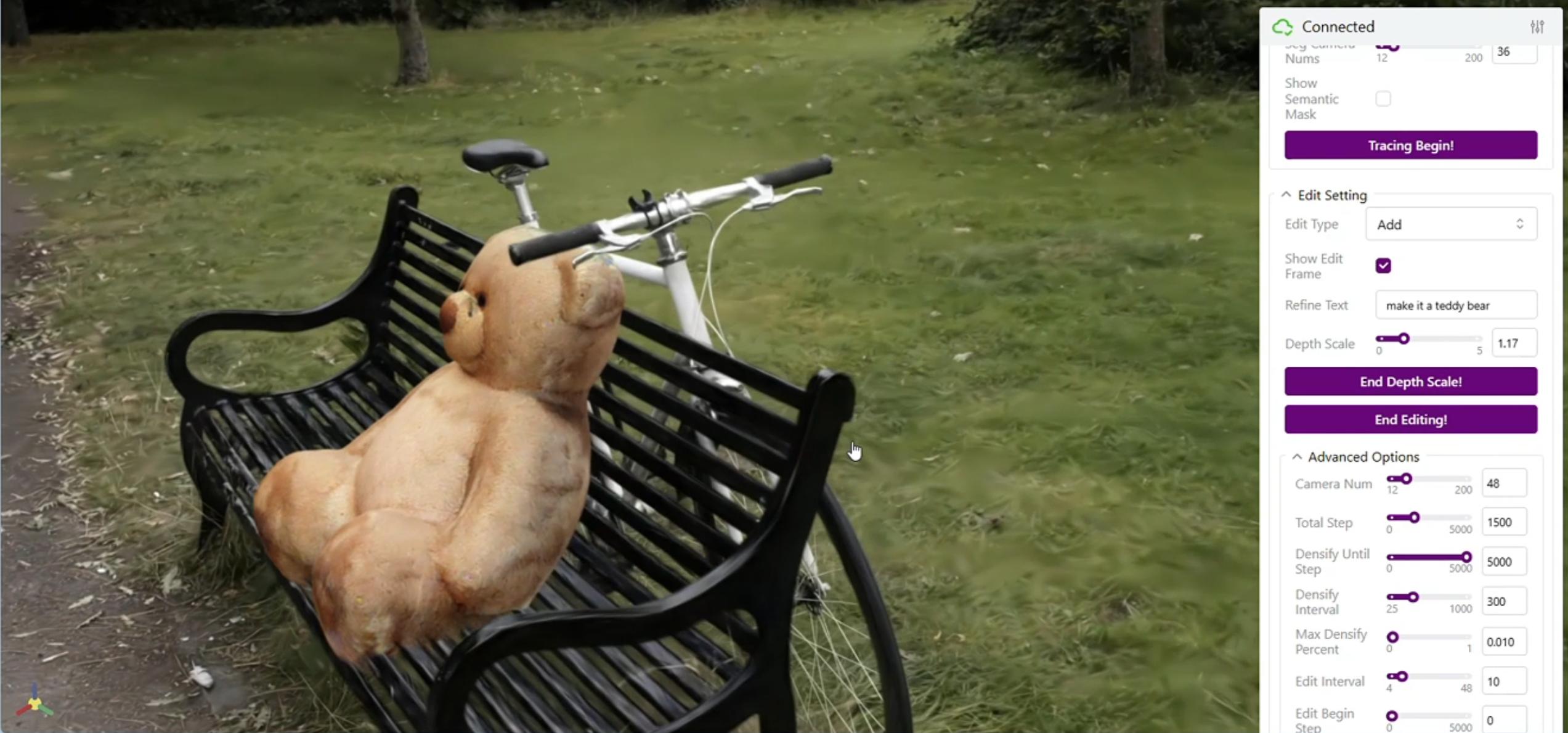}
                                    \caption*{(c)}

        \end{subfigure}
    \end{minipage}
    \hfill
    \begin{minipage}{0.49\textwidth}
        \centering
        \begin{subfigure}{\linewidth}
            \centering
            \includegraphics[width=1.0\linewidth]{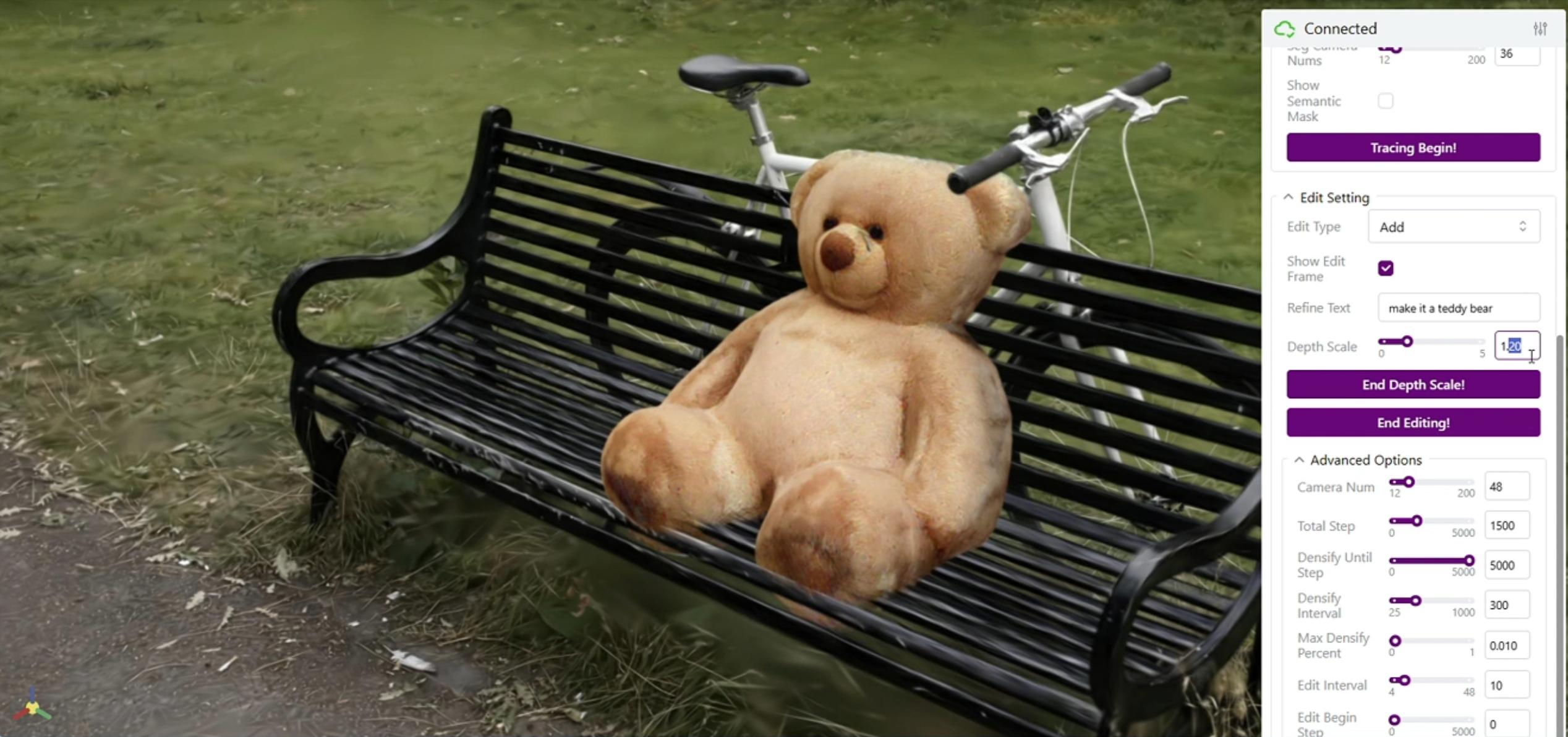}
                                    \caption*{(d)}

        \end{subfigure}
    \end{minipage}

    \caption{Object Incorporation with WebUI. Empowered by our interactive WebUI, the depth scale addresses the limitation of monocular depth estimation, which cannot guarantee precise depth map predictions. As can be seen in (a), inaccurate results can lead to failures when aligning generated objects with Gaussian scenes. We leverage the interactive nature of our WebUI to dynamically adjust the estimated depth in real-time, thus resolving this issue, demonstrated in other images.}
    \label{fig:bear}
\end{figure*}

Subsequently, in other views, we re-project these spatial points onto the camera's imaging plane, identifying the pixels corresponding to these 3D points. We use the projection points of these 3D points in the reference views as the point prompts for semantic segmentation with SAM~\cite{kirillov2023segment}. Then, we unproject these semantic segmentation maps back to Gaussians, as demonstrated in the main text.

As can be seen in Fig.~\ref{fig:vase}, with only about five points indicated by the users, semantic tracing with point-based prompts enables finer granularity control over the areas to be tracked.

\subsection{Object Incorporation with WebUI}
As detailed in the main paper, our proposed method for 3D inpainting with object incorporation allows for the addition of objects specified by text in designated areas. The webUI facilitates users in easily drawing 2D masks to define these areas. Moreover, in this method of 3D inpainting for object incorporation, depth information is crucial for seamlessly integrating new objects into the original Gaussian Scene. Current methods for monocular depth estimation, however, can't always provide completely accurate depth maps, leading to imprecise alignment. Therefore, as depicted in Fig.~\ref{fig:bear}, we utilize the webUI to modify the scale of the estimated depth, enabling users to achieve a more accurate alignment of the objects.

To be more specific, users control the Gaussian scale by sliding a slider. After obtaining a new depth scale, we update the position and size of the added objects according to the new depth scale. Since the entire process involves only minor adjustments to the position and scale parameters of a few Gaussians, real-time scaling can be achieved.

\end{document}